\documentclass[runningheads]{llncs}

\usepackage{eccv}

\usepackage{eccvabbrv}

\usepackage{amsmath}
\usepackage{amssymb}
\usepackage{mathtools}
\usepackage{amsfonts}       %
\usepackage{nicefrac}       %
\usepackage{graphicx}
\usepackage{booktabs}
\usepackage{longtable}
\usepackage{url}
\usepackage{makecell}
\usepackage{array}
\usepackage{subcaption}
\usepackage{diagbox}
\usepackage{multirow}
\usepackage{tcolorbox}
\usepackage[expansion=false]{microtype}
\usepackage{pifont}
\usepackage{marvosym} %
\usepackage[dvipsnames]{xcolor}        %
\usepackage{colortbl}
\usepackage{wrapfig}

\newcommand{\nummodels}{18}

\usepackage[accsupp]{axessibility}  %

\usepackage{hyperref}
\usepackage[capitalize]{cleveref}

\usepackage{orcidlink}

\begin{document}
\raggedbottom
\setlength{\textfloatsep}{8pt plus 2pt minus 2pt}
\setlength{\floatsep}{10pt plus 2pt minus 2pt}
\setlength{\intextsep}{10pt plus 2pt minus 2pt}
\renewcommand{\topfraction}{0.92}
\renewcommand{\bottomfraction}{0.9}
\renewcommand{\textfraction}{0.07}
\renewcommand{\floatpagefraction}{0.85}

\titlerunning{RoboBench}

\author{Yulin Luo\inst{1}$^{\star}$ \and
Chun-Kai Fan\inst{1}$^{\star}$ \and
Menghang Dong\inst{1}$^{\star}$ \and
Jiayu Shi\inst{1}$^{\star}$ \and
Xiangju Mi\inst{1}$^{\star}$ \and
Mengdi Zhao\inst{3}$^{\star\dagger}$ \and
Bo-Wen Zhang\inst{4}$^{\star}$ \and
Cheng Chi\inst{2}$^{\star\dagger}$ \and
Jiaming Liu\inst{1} \and
Gaole Dai\inst{1} \and
Rongyu Zhang\inst{1} \and
Ruichuan An\inst{1} \and
Kun Wu\inst{5} \and
Zhengping Che\inst{5} \and
Shaoxuan Xie\inst{2} \and
Guocai Yao\inst{2} \and
Zhongxia Zhao\inst{1,2} \and
Pengwei Wang\inst{2} \and
Guang Liu\inst{2} \and
Zhongyuan~Wang\inst{2} \and
Tiejun Huang\inst{1,2} \and
Shanghang Zhang\inst{1,2}\textsuperscript{\Letter}}

\authorrunning{Y.~Luo et al.}

\institute{\small
$^{1}$State Key Laboratory of Multimedia Information Processing, School of Computer Science, Peking University \quad
$^{2}$Beijing Academy of Artificial Intelligence \\
$^{3}$Institute for Brain and Intelligence, Fudan University \quad
$^{4}$University of Science and Technology Beijing \quad
$^{5}$Beijing Innovation Center of Humanoid Robotics \\
\email{yulin@stu.pku.edu.cn, shanghang@pku.edu.cn}}
\title{RoboBench: A Comprehensive Evaluation Benchmark for Multimodal Large \texorpdfstring{\\}{ }Language Models as Embodied Brain}
\maketitle
{\renewcommand{\thefootnote}{}\footnotetext{$^{\star}$Equal contribution.\quad $^{\dagger}$Project leader.\quad \textsuperscript{\Letter}\,Corresponding author.}}

\begin{abstract}
Building robots that can perceive, reason, and act in dynamic, unstructured environments remains a core challenge. Recent embodied systems often adopt a dual-system paradigm, where System~2 handles high-level reasoning while System~1 executes low-level control. 
In this work, we refer to System~2 as the \textit{embodied brain}, emphasizing its role as the cognitive core for reasoning and decision-making in manipulation tasks.
Given this role, systematic evaluation of the embodied brain is essential for advancing robotic intelligence. 
Yet existing benchmarks emphasize execution success, or, when targeting high-level reasoning, suffer from incomplete dimensions and limited task realism, offering only a partial picture of cognitive capability.
To bridge this gap, we introduce RoboBench, a benchmark that systematically evaluates multimodal large language models (MLLMs) as embodied brains. Motivated by the distinct cognitive roles required across the full manipulation pipeline, RoboBench defines five dimensions—Instruction Comprehension, Perception Reasoning, Generalized Planning, Affordance Prediction, and Failure Analysis—spanning 14 capabilities, 25 tasks, and 6092 QA pairs. 
To ensure realism, we curate datasets across diverse embodiments, attribute-rich objects, multi-view scenes, and memory-driven navigation, drawing from large-scale real robotic data and in-house collection. 
For planning, RoboBench introduces an evaluation framework that uses an MLLM as a world simulator. It moves beyond symbolic matching to evaluate embodied feasibility by simulating whether predicted plans can achieve critical object-state changes under physical and visual constraints, enabling faithful assessment of long-horizon reasoning. 
Experiments on \nummodels{} state-of-the-art MLLMs reveal fundamental limitations: difficulties with implicit instruction comprehension, spatiotemporal reasoning, cross-scenario planning, fine-grained affordance understanding, and execution failure diagnosis. We further analyze how embodied cognitive abilities relate to downstream robotic control.
RoboBench provides a comprehensive scaffold to quantify high-level cognition, clarify the role of the embodied brain, and guide the development of next-generation MLLMs for more robust robotic intelligence. Project page: \mbox{\url{https://robo-bench.github.io}}.


\end{abstract}

\begin{figure*}[t]
\centering
\includegraphics[width=\textwidth]{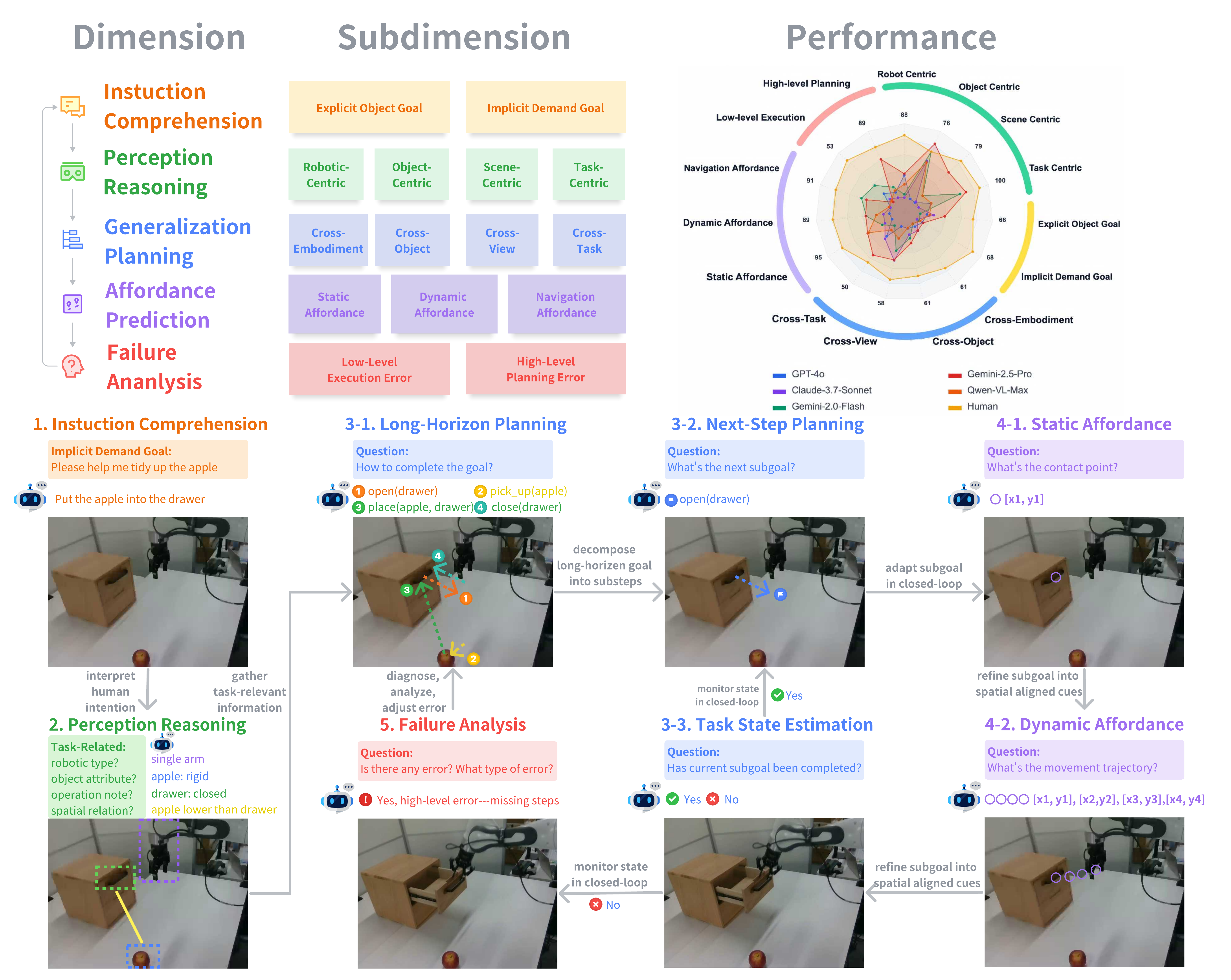}
\captionsetup{hypcap=false}
\caption{{\textbf{Overview of RoboBench}
We evaluate MLLMs as embodied brains with 25 tasks color-coded by 5 dimensions \textbf{(top left)}. These dimensions follow the embodied execution pipeline \textbf{(bottom)}—from intent understanding to failure diagnosing—capturing the core cognitive roles of System 2. Performance comparison \textbf{(top middle)} reveals significant gaps among state-of-the-art MLLMs
} \textbf{(top right)} RoboBench scores strongly correlate with downstream VLA performance on CALVIN.}
\label{fig: 1_teaser}
\end{figure*}

\section{Introduction}
Manipulation in dynamic, unstructured environments remains a core challenge for building generalizable embodied intelligence~\cite{Sun2024,clark1998being,liu2024aligningcyberspacephysical}. 
Such tasks demand not only precise motor control, but also high-level cognition: understanding instructions, perceiving the surroundings, formulating long-horizon plans, inferring affordances, and reflecting on failures~\cite{ chen2025exploringembodiedmultimodallarge, zhang2024vlabenchlargescalebenchmarklanguageconditioned}. 
In this context, multimodal large language models (MLLMs) have shown strong potential for these roles, especially in instruction-following, commonsense reasoning, and general planning~\cite{li2024llavaonevisioneasyvisualtask, wang2024qwen2vlenhancingvisionlanguagemodels}. 
To leverage these capabilities, recent embodied systems integrate MLLMs through a dual-system design~\cite{driess2023palm, black2024pi0visionlanguageactionflowmodel, bjorck2025gr00t}, where System~2 performs high-level reasoning and System~1 handles low-level control~\cite{liu2025hybridvla, chen2025fisvla}. 
In vision-language-action (VLA) models, MLLMs are fine-tuned as backbones~\cite{black2024pi0visionlanguageactionflowmodel}, while in multi-agent frameworks, they serve as high-level planners guiding specialized executors~\cite{driess2023palm, huang2023voxposercomposable3dvalue, huang2024rekepspatiotemporalreasoningrelational, robobrain_2025}. In this work, we call System~2 the \textit{embodied brain}, the cognitive core of reasoning and decision-making.

Given this design, systematic evaluation of the embodied brain is essential, yet current benchmarks remain inadequate.
Most existing efforts focus narrowly on overall task success, offering little insight into underlying reasoning~\cite{james2020rlbench, liu2023libero, chen2025robotwin}.

\begin{figure*}[t]
\centering
\includegraphics[width=\textwidth]{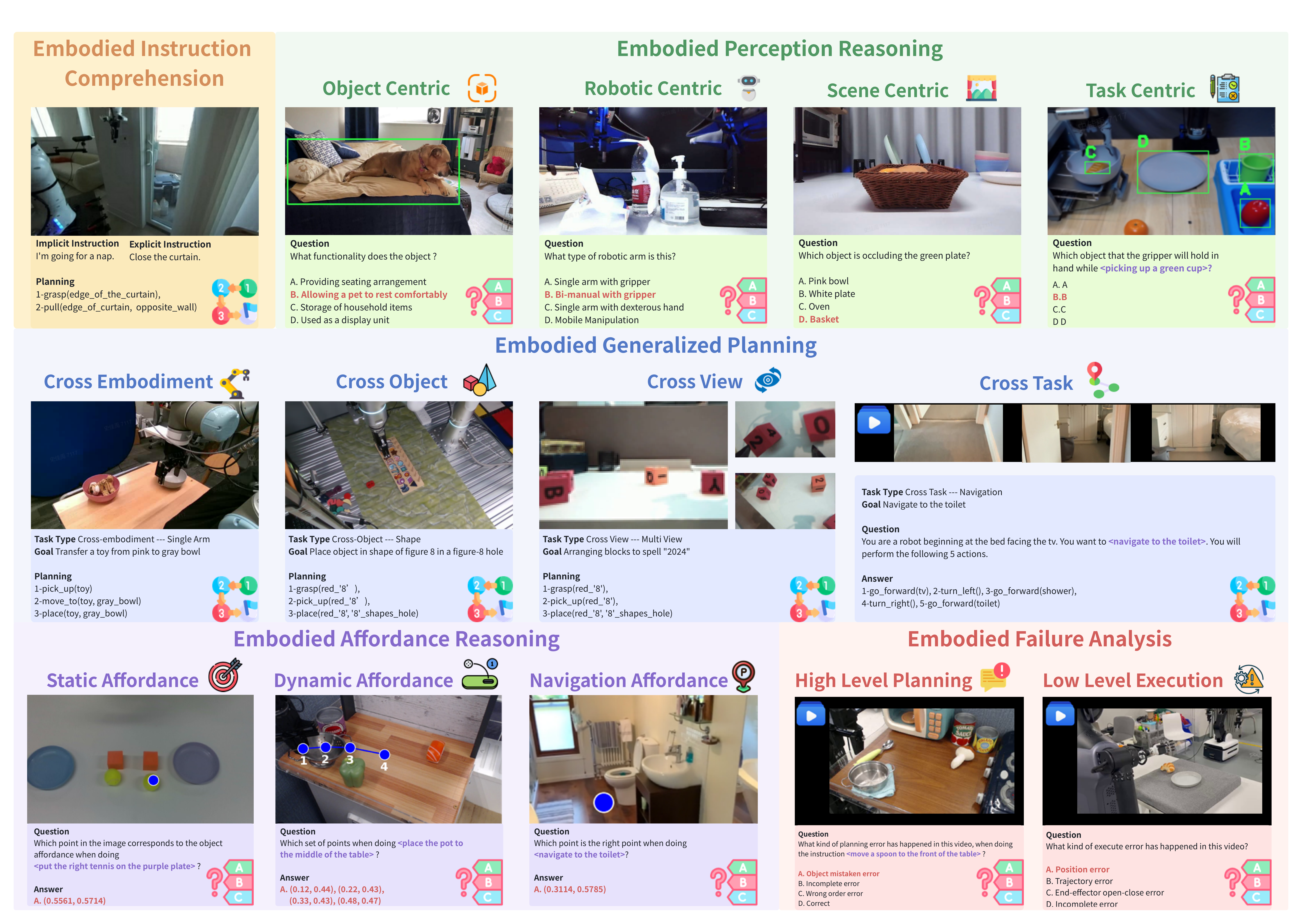}
\captionsetup{hypcap=false}
\caption{{\textbf{Examples of RoboBench}  %
Our benchmark covers 5 dimensions, 14 capabilities, and 25 tasks. We design diverse question formats, including multiple-choice, planning, and point prediction.
}}
\label{fig: 2_demo_case}
\end{figure*}

Even benchmarks that explicitly target embodied cognition exhibit three major shortcomings: 
(1) fragmented coverage of cognitive abilities, typically isolating perception~\cite{Majumdar_2024_CVPR}, planning~\cite{sermanet2024robovqa, chen2024egoplanbenchbenchmarkingmultimodallarge, NEURIPS2024_b631da75}, or error reflection~\cite{liu2023reflect, duan2024aha} rather than assessing them as an integrated whole;
(2) limited task realism and complexity, relying heavily on simulations~\cite{Shridhar_2020_CVPR, shridhar2020alfworld,yang2025embodiedbench, cheng2025embodiedevalevaluatemultimodalllms, zhang2024vlabenchlargescalebenchmarklanguageconditioned}, or ignoring practical challenges such as diverse embodiment, object properties, and occlusion; 
and (3) simplistic planning evaluation, often reduced to multiple-choice~\cite{chen2024egoplanbenchbenchmarkingmultimodallarge, azzolini2025cosmos, team2025gemini} or text similarity metrics such as BLEU~\cite{sermanet2024robovqa} and LLM scoring~\cite{chi2024evaembodiedworldmodel}, failing to capture embodied priors such as skill-object dependencies, execution order flexibility, and feasibility.\looseness=-1

To overcome these gaps, we introduce \textbf{RoboBench}, a benchmark systematically designed to evaluate MLLMs as the cognitive core for robotic manipulation. Specifically, we highlight three key features of RoboBench:
(1) \textbf{Comprehensive evaluation dimensions.} 
RoboBench defines five dimensions—Instruction Comprehension, Perception Reasoning, Generalized Planning, Affordance Prediction, and Failure Analysis—to capture the key capabilities required for embodied cognition. These dimensions follow the manipulation execution pipeline (Fig.~\ref{fig: 1_teaser}): interpreting human intent, perceiving task-relevant context, decomposing long-horizon goals, grounding subgoals into spatial affordances, and diagnosing failures during execution~\cite{zhang2024vlabenchlargescalebenchmarklanguageconditioned, liu2024moka, robobrain_2025, RoboBrain2.0TechnicalReport, duan2024aha}.
(2) \textbf{Realistic and diverse tasks.} 
RoboBench covers diverse embodied scenarios including single-arm, dual-arm, and mobile manipulation, objects with rich physical attributes, and multi-view scenes with occlusion and navigation memory~\cite{yang2024thinkingspacemultimodallarge}. The dataset combines large-scale real-world robotic data~\cite{fang2023rh20t, open_x_embodiment_rt_x_2023, liu2024aligningcyberspacephysical, wu2024robomind, bu2025agibot} with curated in-house collection to better reflect real-world embodied interaction.
(3) \textbf{World-simulation rollout for planning evaluation.} 
We propose an \textit{MLLM-as-world-simulator} framework for long-horizon planning. Given the initial scene, reference plan, and a human-annotated DAG of task dependencies, the evaluator rolls out the plan step-by-step under visual and physical constraints, checking whether it reaches key milestones and stays physically executable.\looseness=-1

We evaluate \nummodels{} state-of-the-art MLLMs and uncover fundamental limitations: poor grounding of implicit instructions, fragile embodiment-specific and spatiotemporal perception, difficulty in long-horizon and cross-scenario planning, shallow fine-grained affordance understanding, and weak execution failure diagnosis. These findings show that while MLLMs hold promise as embodied brains, their reasoning remains superficial. 
We further correlate RoboBench scores with downstream VLA benchmark performance and find that abilities such as perception reasoning and affordance prediction strongly predict manipulation success. This establishes RoboBench as a principled framework for selecting VLM backbones and guiding the development of stronger embodied reasoning.

In summary, our contributions are as below:
\begin{itemize} 
 \item We introduce RoboBench, a comprehensive benchmark for evaluating MLLMs as embodied brains across 5 key dimensions, spanning 14 capabilities, 25 tasks, and 6092 QA pairs.
 \item We design realistic datasets across embodiments, objects, and viewpoints that reflect real-world embodied interaction.
 \item We propose an MLLM-as-world-simulator evaluation framework that faithfully assesses long-horizon planning by simulating whether plans can achieve critical object-state milestones.
 \item We conduct a large-scale evaluation of SOTA MLLMs, revealing key embodied cognitive limitations and demonstrating that RoboBench abilities predict downstream VLA performance and guide VLM selection.
\end{itemize}

\section{Related Work}
\label{sec:related_works}
With the rapid development of embodied AI, the number of benchmarks has also grown~\cite{sermanet2024robovqa, chen2024egoplanbenchbenchmarkingmultimodallarge, li2024mmromultimodalllmseligible, cheng2024videgothinkassessingegocentricvideo, yang2025embodiedbench, qi2025bear, dang2025ecbench}. However, current benchmarks still have significant limitations in evaluating the capabilities of embodied intelligence. We analyze these limitations from four aspects: comprehensiveness, realism, complexity, and evaluation of planning tasks.

\begin{table*}[t]
\centering
\caption{Comparison of various benchmarks for embodied AI, evaluating their features across comprehensiveness, reality, complexity, and planning evaluation methodology.}
\label{tab:benchmark_comparison}
\resizebox{\textwidth}{!}{%
\begin{tabular}{l*{11}{c}}
\toprule
\multirow{2}{*}{\textbf{Benchmark}} & \multicolumn{5}{c}{\textbf{Comprehensiveness}} & \multicolumn{2}{c}{\textbf{Reality}} & \multicolumn{2}{c}{\textbf{Complexity}} & \multirow{2}{*}{\begin{tabular}[c]{@{}c@{}}\textbf{Planning} \\ \textbf{Evaluation}\\\textbf{method}\end{tabular}} & \multirow{2}{*}{\textbf{Size}} \\
\cmidrule(lr){2-6} \cmidrule(lr){7-8} \cmidrule(lr){9-10}
& \begin{tabular}[c]{@{}c@{}}Language \\ Instruction\end{tabular} & \begin{tabular}[c]{@{}c@{}}Perception \\ Reasoning\end{tabular} & \begin{tabular}[c]{@{}c@{}}Generalized \\ Planning\end{tabular} & \begin{tabular}[c]{@{}c@{}}Affordance \\ Prediction\end{tabular} & \begin{tabular}[c]{@{}c@{}}Failure \\ Analysis\end{tabular} & \begin{tabular}[c]{@{}c@{}}Real \\ Scene\end{tabular} & \begin{tabular}[c]{@{}c@{}}Real \\ Robot\end{tabular} & \begin{tabular}[c]{@{}c@{}}Cross \\ Embodiment\end{tabular} & \begin{tabular}[c]{@{}c@{}}Multi \\ View\end{tabular} & & \\
\midrule
RoboVQA~\cite{sermanet2024robovqa}       & \checkmark & $\times$   & \checkmark & $\times$   & $\times$   & \checkmark & \checkmark & \checkmark & $\times$   & BLEU                                                              & 18,248  \\
EgoPlanBench~\cite{chen2024egoplanbenchbenchmarkingmultimodallarge}  & $\times$   & \checkmark & \checkmark & $\times$   & $\times$   & \checkmark & $\times$   & $\times$   & $\times$   & Multi-Choice                                                               & 3,355   \\
MMRo~\cite{li2024mmromultimodalllmseligible}          & $\times$   & \checkmark & \checkmark & $\times$   & $\times$   & \checkmark & \checkmark & $\times$   & $\times$   & LLM Score                                                    & 26,175  \\
EAI~\cite{NEURIPS2024_b631da75}           & $\times$   & \checkmark & \checkmark & $\times$   & $\times$   & $\times$   & $\times$   & $\times$   & $\times$   & Simulator Task SR                                                 & 438     \\
OpenEQA~\cite{Majumdar_2024_CVPR}       & $\times$   & \checkmark & \checkmark & $\times$   & $\times$   & \checkmark & $\times$   & $\times$   & $\times$   & LLM Score                                                         & 1,636   \\
EgoThink~\cite{Cheng_2024_CVPR}      & $\times$   & \checkmark & \checkmark & $\times$   & $\times$   & \checkmark & $\times$   & $\times$   & $\times$   & LLM Score                                                         & 700     \\
VideoEgoThink~\cite{cheng2024videgothinkassessingegocentricvideo} & $\times$   & \checkmark & \checkmark & $\times$   & $\times$   & \checkmark & $\times$   & $\times$   & $\times$   & LLM Score                                                         & 4,993   \\
EmbodiedEval~\cite{cheng2025embodiedevalevaluatemultimodalllms}  & $\times$   & \checkmark & \checkmark & $\times$   & \checkmark & $\times$   & $\times$   & $\times$   & $\times$   & Multi-Choice                                                                & 327     \\
EmbodiedBench~\cite{yang2025embodiedbench} & $\times$   & \checkmark & \checkmark & $\times$   & \checkmark & $\times$   & $\times$   & \checkmark & $\times$   & Simulator Task SR                                                 & 1,128   \\
VLABench~\cite{zhang2024vlabenchlargescalebenchmarklanguageconditioned}      & \checkmark & \checkmark & \checkmark & $\times$   & $\times$   & $\times$ & \checkmark   & \checkmark & \checkmark & Simulator Task SR                                                 & 2,427   \\
BEAR~\cite{qi2025bear} & $\times$ & \checkmark & \checkmark & \checkmark & $\times$ & \checkmark & $\times$ & \checkmark & \checkmark & Multi-Choice & 4,469 \\
ECBench~\cite{dang2025ecbench} & $\times$ & \checkmark & $\times$ & $\times$ & $\times$ & \checkmark & $\times$ & $\times$ & $\times$ & LLM Score & 4,324 \\
RoboBench (Ours)     & \checkmark & \checkmark & \checkmark & \checkmark & \checkmark & \checkmark & \checkmark & \checkmark & \checkmark & \begin{tabular}[c]{@{}c@{}}MLLM World Simulator
\end{tabular} & 6,092   \\
\bottomrule
\end{tabular}
}
\end{table*}

Existing benchmarks related to embodied cognition tend to assess individual abilities separately — for example, perception~\cite{Majumdar_2024_CVPR, Cheng_2024_CVPR}, planning~\cite{sermanet2024robovqa, chen2024egoplanbenchbenchmarkingmultimodallarge, NEURIPS2024_b631da75}, and affordance prediction~\cite{yuan2024robopoint, zhou2025roborefer} or failure analysis~\cite{liu2023reflect, duan2024aha, zhou2025code}. None of them comprehensively evaluate Multimodal Large Language Models (MLLMs) as integrated embodied brains.
In terms of realism, some benchmarks rely on simulators, incurring a sim-to-real gap~\cite{yang2025embodiedbench, cheng2025embodiedevalevaluatemultimodalllms, zhang2024vlabenchlargescalebenchmarklanguageconditioned}, while others build on human daily-activity or egocentric video~\cite{chen2024egoplanbenchbenchmarkingmultimodallarge, cheng2024videgothinkassessingegocentricvideo, dang2025ecbench} that still differs from real robotic manipulation.
Regarding complexity, many benchmarks fail to account for cross-embodiment and cross-viewpoint scenarios~\cite{li2024mmromultimodalllmseligible, chen2024egoplanbenchbenchmarkingmultimodallarge, Majumdar_2024_CVPR}.
For planning evaluation, existing metrics are also limited. Text-similarity metrics such as BLEU~\cite{sermanet2024robovqa} capture only surface-level wording, ignoring the structural and physical feasibility of plans. Multiple-choice evaluations~\cite{chen2024egoplanbenchbenchmarkingmultimodallarge} deviate from real-world tasks whose solutions are open-ended. LLM-based judgments~\cite{chi2024evaembodiedworldmodel} introduce subjectivity, often failing to credit logically valid plans phrased differently from the reference. Close to our work, ECBench~\cite{dang2025ecbench} and BEAR~\cite{qi2025bear} pursue broad-coverage evaluation, yet neither covers the full embodied-brain loop: ECBench omits generalized planning, affordance, and failure analysis, while BEAR still diagnoses atomic skills in isolation, omits execution failure analysis, and adopts multiple-choice scoring that constrains open-ended planning. In contrast, RoboBench unifies all five embodied-cognition dimensions on real-robot data with an MLLM world-simulator judge for open-ended planning (Table~\ref{tab:benchmark_comparison}).

\section{RoboBench}

\subsection{Core Capabilities}
\label{subsec: core capabilities}
Through a detailed analysis, we identify five critical dimensions of the MLLM-based embodied brain, each aligned with the task execution pipeline: understanding human intent, perceiving the environment, formulating and adapting plans, refining actions via affordance prediction, and diagnosing failures. RoboBench evaluates these capabilities to uncover bottlenecks in embodied cognition.

\begin{figure*}[t]
\includegraphics[width=1\textwidth]{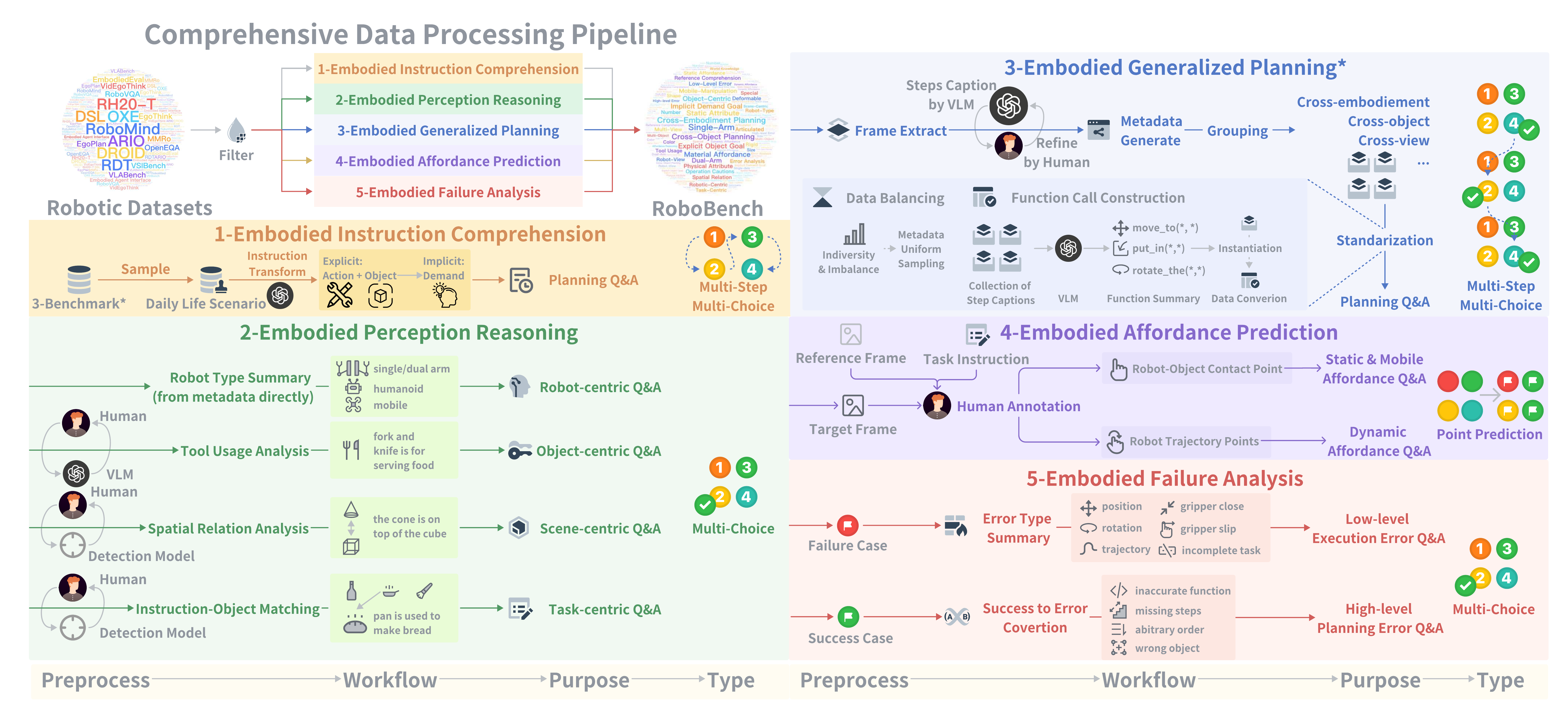}
\centering
\caption{\textbf{Dataset Construction Pipeline.}
RoboBench customizes data workflows for each dimension: \textcolor{orange}{orange} for instruction, \textcolor{OliveGreen}{green} for perception, \textcolor{blue}{blue} for planning, \textcolor{purple}{purple} for affordance, and \textcolor{red}{red} for reflection. Each workflow generally follows three stages—preprocessing; annotation with VLMs, detection models, or human experts; and Q\&A format generation.
}
\label{fig:3_pipeline}
\end{figure*}

\textbf{Embodied Instruction Comprehension}: \textit{Can the embodied brain understand human intention?}  Most embodied tasks rely on explicit instructions where actions and targets are clearly specified~\cite{sermanet2024robovqa, yang2025embodiedbench, li2025embodied}. In contrast, real-world instructions are often implicit (e.g., “I’m thirsty” instead of “Retrieve a drink”)~\cite{zhang2024vlabenchlargescalebenchmarklanguageconditioned}. 
This dimension evaluates whether models can translate both explicit and implicit instructions into actionable plans.

\textbf{Embodied Perception Reasoning}: \textit{Can the embodied brain perceive the environment to gather task-relevant information?} Reliable planning and execution rely on accurate perception~\cite{Majumdar_2024_CVPR, li2024mmromultimodalllmseligible, song2025robospatial, chen2025robo2vlm}, 
RoboBench evaluates this through four aspects: Robotic-centric considers embodiment type and viewpoint. Object-centric examines static and functional attributes. Scene-centric evaluates spatial relations, temporal grounding, and causality analysis. Task-centric assesses the identification of instruction-relevant
 objects.

\textbf{Embodied Generalized Planning}: \textit{Can the embodied brain generalize planning across embodiments, objects, views, and tasks?} Planning begins with decomposing long-horizon goals into subgoals, and continues during execution through predicting the next subtask, monitoring completion, and adapting subsequent steps~\cite{sermanet2024robovqa, yang2025embodiedbench}. We evaluate generalized planning across four aspects: embodiments (single-arm, dual-arm, mobile manipulator, and humanoid robot), objects (material affordance, physical attribute, and world knowledge), views (multi-view integration under occlusion), and tasks (navigation planning using spatial cues from videos~\cite{yang2024thinkingspacemultimodallarge}).

\textbf{Embodied Affordance Prediction}: \textit{Can the embodied brain refine subtask plans through spatial affordances?} Beyond high-level planning, each subgoal should be transformed into spatial cues to guide low-level execution~\cite{liu2024moka, li2024chain_of_affordance, yuan2024robopoint, zhou2025roborefer, team2025gemini, lee2025molmoact}. Affordance prediction links subgoals with the embodiment, objects, and environment, enabling System 2 to instruct System 1 more effectively~\cite{li2024chain_of_affordance}. RoboBench evaluates three types: Static affordance identifies contact points (e.g., grasp an apple); Dynamic affordance predicts motion trajectories (e.g., open a drawer); Navigation affordance determines robot base position (e.g., approach a microwave on a distant table).

\textbf{Embodied Failure Analysis}: \textit{Can the embodied brain detect and correct failures?} Open-world manipulation inevitably introduces errors, requiring the brain not only to identify them but also to diagnose causes and suggest corrections~\cite{liu2023reflect, duan2024aha}. RoboBench evaluates low-level execution errors (e.g., position misalignment, trajectory deviations, gripper failures, and incomplete actions), and high-level planning errors (e.g., wrong object, missing steps, incorrect ordering), providing insights for more robust and generalizable execution.

\subsection{Benchmark Construction}
\label{subsec: benchmark construction}
\paragraph{Dataset Collection and Processing Pipeline}
To evaluate the five cognitive dimensions, we build datasets from open-source and in-house robotic data, enriched with MLLM- and human-provided annotations, aligning each sub-benchmark with the embodied task pipeline for realism. The examples and construction pipeline are shown in Figures \ref{fig: 2_demo_case} and \ref{fig:3_pipeline}.

\textbf{Instruction Comprehension}: Evaluated through planning tasks, this dimension adopts a paired explicit--implicit design. Explicit instructions are drawn from daily-life scenarios with clearly specified actions and targets, whereas implicit instructions are generated by an MLLM conditioned on both the scene image and the explicit goal, then manually checked. A sampled audit confirms the generated implicit instructions are natural and inferable from the scene (audit details in Appendix~D.1), testing a model's ability to interpret human intentions.

\textbf{Perception Reasoning}: Accurate perception is essential for reliable planning and execution. We construct datasets across four aspects: robotic-centric, using real robot data with type and view metadata for template-based QA; object-centric, combining curated static attributes~\cite{gao2024physically} with GPT-generated functional properties and distractors; scene-centric, leveraging Gemini-segmented video steps for temporal grounding, together with manual annotations of relative positions and keypoint changes for spatial and causal reasoning; and task-centric, with human-labeled bounding boxes linking long-horizon instructions to target objects. All data are standardized into multiple-choice QA.

\textbf{Generalized Planning}:
We build a planning pool from high-quality robotic videos~\cite{open_x_embodiment_rt_x_2023, lin2024dsl, liu2024rdt, liu2024aligningcyberspacephysical, fang2023rh20t, wu2024robomind, mu2023embodiedgpt, yang2024thinkingspacemultimodallarge}, as standardized inputs. Gemini~\cite{comanici2025gemini2.5} generates structured annotations—task summaries, step-wise instructions with timestamps, and metadata (objects, actions, scenes, embodiments)—which are refined by human annotators. Each step is then mapped into function templates (e.g., pick\_up(object), move\_to(object, target)) grouped into manipulation or navigation skill lists~\cite{cheng2024videgothinkassessingegocentricvideo} for structured plan generation.
Evaluation covers three types~\cite{sermanet2024robovqa}: (1) Q1: long-horizon planning, predicting the full action sequence from the first frame; (2) Q2: next-step planning, forecasting the (n+1)-th step; and (3) Q3: task state estimation, deciding whether a subtask is completed.

\textbf{Affordance Prediction}:
Affordances refine high-level subgoals into spatial cues for low-level execution~\cite{liu2024moka}. From the planning pool, representative frames are sampled and annotated with three types of affordance: static (contact points), dynamic (motion trajectories), and mobile (base positions). Each annotated affordance is then cast into a multiple-choice question in which the model selects, given the task instruction, the correct contact point, motion trajectory, or base position from scene-grounded candidate options.

\textbf{Failure Analysis}:
It evaluates whether models can detect and reason about errors during execution~\cite{liu2023reflect, duan2024aha}. Execution-level failures (e.g., position misalignment, trajectory deviations, gripper errors, premature release, and trajectory overshoot) are collected from real VLA rollouts in RoboMIND~\cite{wu2024robomind} and labeled by experts, so they reflect actual low-level execution failures rather than perturbations of successful trajectories. Because real high-level planning-failure data is scarce, planning-level failures follow the REFLECT-style protocol~\cite{liu2023reflect}: we perturb correct plans with canonical classes including wrong object, missing step, wrong order, and wrong condition, then manually verify that the injected failure is identifiable and task-relevant.

\paragraph{Quality Control}
We adopt a two-stage quality control process to ensure benchmark quality: data filtering during construction and post-construction validation. In the construction stage, 
we apply both general and task-specific filters. General criteria address image quality and task validity. Sub-benchmarks also have tailored rules, such as excluding single-arm data from dual-arm tasks. After construction, professional researchers validate linguistic clarity, answerability, and correctness~\cite{liu2024mmbenchmultimodalmodelallaround}. Every MLLM-assisted annotation receives at least one human cross-pass, and validation disagreements are tracked per dimension before adjudication; residual disagreement on the validation subset is below 4\%. We further adopt a majority-vote strategy~\cite{liu2024mmbenchmultimodalmodelallaround}: items all models solve are removed, and items all models fail undergo manual review.

\begin{figure}[t]
  \centering

  \begin{minipage}[t]{0.58\textwidth}
    \captionsetup{type=figure}
    \includegraphics[width=\linewidth,keepaspectratio]{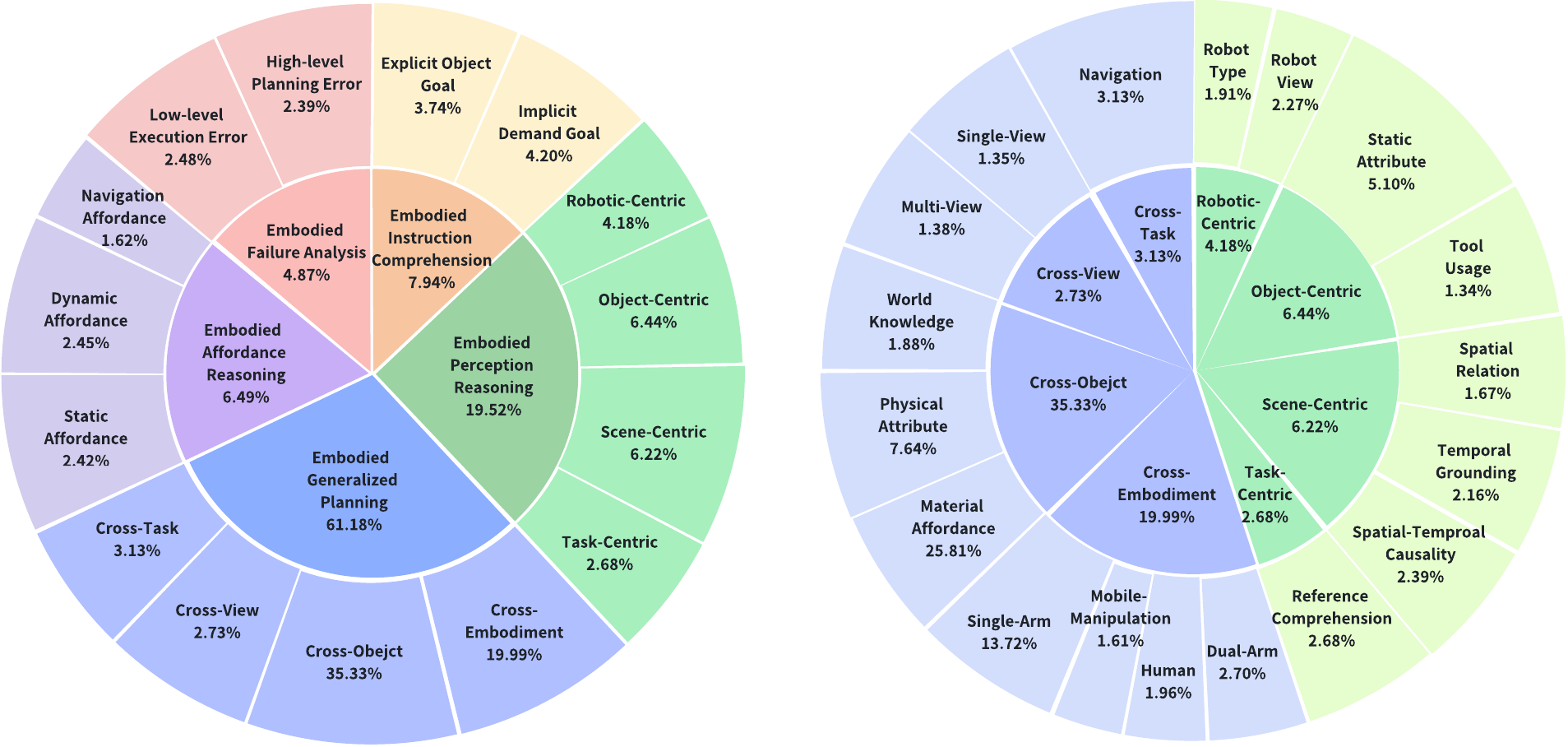}
    \caption{\textbf{Dimension distribution.}}
    \label{fig:4_distribution}
  \end{minipage}
  \hfill
\begin{minipage}[t]{0.40\textwidth}
  \captionsetup{type=table}

  \scriptsize
  \setlength{\tabcolsep}{2pt}
  \renewcommand{\arraystretch}{0.95}
  \resizebox{\linewidth}{!}{%
    \begin{tabular}{@{}l r@{}}
      \toprule
      \textbf{Statistic} & \textbf{Number} \\
      \midrule
      Total items & 4336\\
      Total questions  & 6092\\
      \quad Multiple-choice questions & 1895\\
      \quad\quad Perception / Affordance / Failure & 1205 / 394 / 296\\
      \quad Instruction \& Planning questions \\
      \quad\quad Q1 / Q2 / Q3 & 1973 / 842 / 1192\\
      \quad\quad Navigation & 190\\
      \midrule
      \textit{Planning details (Q1)} \\
      \quad Avg. / Max. steps & 6.75 / 55\\
      \quad Unique instructions / answers & 1403 / 1462\\
      \quad Avg. frames (Q1 / Q2 / Q3) & 1.1 / 10.9 / 8.1\\
      \textit{Perception details} \\
      \quad Avg. / Max. question len. & 93.3 / 317\\
      \quad Avg. / Max. choice len. & 11.8 / 98\\
      \textit{Failure details} \\
      \quad Avg. / Max. question len. & 92.3 / 264\\
      \quad Avg. / Max. choice len. & 17.0 / 29\\
      \bottomrule
    \end{tabular}
  }%
\caption{\textbf{Dataset statistics.}}
\label{tab:dataset-stats}
\end{minipage}

\end{figure}

\paragraph{Dataset Statistics}
\label{subsec: dataset statistics}
RoboBench consists of 6092 samples and 4336 unique items, providing a balanced mix of diversity and complexity to evaluate embodied brain capabilities. It covers 5 dimensions, 14 capabilities, and 25 tasks, ensuring comprehensive and challenging assessments. The detailed distribution and per-dimension question statistics are shown in Fig.~\ref{fig:4_distribution} and Table~\ref{tab:dataset-stats}.

\begin{figure*}[t]
\includegraphics[width=1\textwidth]{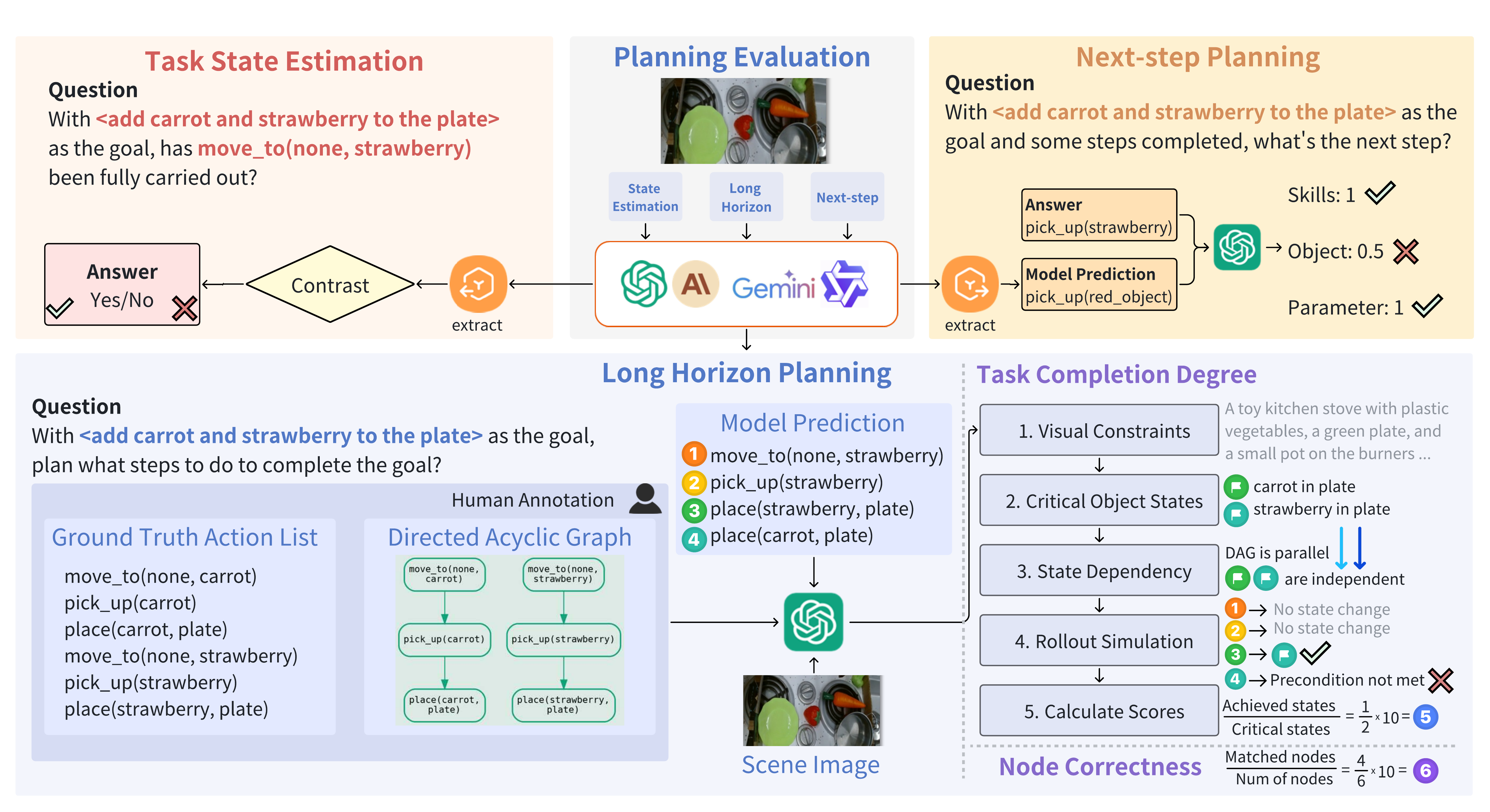}
\centering
\caption{\textbf{Planning Evaluation Pipeline.} The planning benchmark includes three task types: long-horizon planning, next-step planning, and task state estimation. They are evaluated respectively with our proposed MLLM-as-world-simulator framework, LLM scoring, and binary accuracy.}
\label{fig:eval_pipeline}
\end{figure*}

\section{Evaluation Metrics}
\label{sec:metric}

To capture the diverse cognitive demands of embodied manipulation, we tailor metrics to each dimension.  
Perception reasoning, affordance prediction, and failure analysis are evaluated by multiple-choice accuracy~\cite{azzolini2025cosmos, team2025gemini}.  
Planning is assessed across three tasks (Q1--Q3). Q1 serves as the core, leveraging an MLLM world simulator to evaluate both structural correctness and embodied feasibility.

\textbf{Q1 -- Long-horizon planning.}
We model a task as a partially ordered set of \emph{atomic actions} parameterized by
\(\langle \text{skill}, \text{object}, \text{args} \rangle\).
Inter-step constraints are encoded by a Directed Acyclic Graph (DAG) \(G=(V,E)\),
where each \(v\in V\) is an action node and each edge \((u\!\to\!v)\in E\) enforces precedence.
The DAG exposes milestones (e.g., \(\texttt{drawer=open}\)) while allowing valid permutations.
Manually annotated \(G\) serves as the reference for evaluation. The pipeline is in Fig.~\ref{fig:eval_pipeline}.

We calculate two metrics—\emph{NodeCorrectness} and \emph{TaskCompletion}—both computed via an MLLM.
The final score is \(\mathrm{LongHorizon} = (\mathrm{NodeCorrectness} + \mathrm{TaskCompletion})/20\in[0,1]\).

\emph{Node correctness.}
Let \(V^\star\) denote ground-truth nodes and \(\hat V\) predicted nodes.
An MLLM performs one-to-one matching under exact alignment on \{\text{skill}, \text{object}, \text{parameter}\}.
The calculation is:
\(\mathrm{NodeCorrectness}
=\left\lfloor \frac{|V^\star \cap \hat V|}{|V^\star|}\times 10 \right\rfloor .
\)

\emph{Task completion.}
An MLLM conducts a lightweight world-simulation rollout using:
(i) first-frame image \(I_0\) for spatial/physical priors,
(ii) reference action list \(A^\star\), and
(iii) DAG \(G\).
We define \emph{critical states} as object-level state changes
(e.g., \texttt{drawer=open},\ \texttt{apple on plate}).
Let \(S^\star\) be the set of such milestones and \(\hat S\) those achieved during rollout.
The MLLM proceeds as follows:

\begin{enumerate}
\item \textbf{Visual constraint analysis:}
from \(I_0\), identify objects, spatial relations, and physical constraints to form the initial world \(W_0\)
(e.g., \texttt{drawer=closed},\ \texttt{apple on table}).

\item \textbf{Critical object-state detection:}
parse \(A^\star\) and \(G=(V,E)\) to extract state-transition predicates from action effects, then aggregate into \(S^\star\).

\item \textbf{State order \& concurrency validation:}
use \(G\) to enforce precedence and allowable parallelism among states in \(S^\star\), and establish causal links.

\item \textbf{Rollout simulation:}
execute predicted actions step-by-step, checking preconditions against \(W_t\), updating to \(W_{t+1}\),
and marking \(s\in S^\star\) as \emph{achieved \& protected} when it becomes true and remains valid to its last consumer,
accumulating \(\hat S\subseteq S^\star\).
\end{enumerate}

The score is computed as \(\mathrm{TaskCompletion}=\left\lfloor \frac{|\hat S|}{|S^\star|}\times 10 \right\rfloor\).
Across four judge models from three vendors and three prompt variants on 200 stratified Q1 items, this protocol attains high inter-rater agreement (NodeCorrectness ICC(2,k)$=$0.948), confirming the evaluator is not tied to a single grader or prompt; the full robustness analysis is provided in Appendix~D.2.

\textbf{Q2 -- Next-step planning.} 
Given an observation \(I_t\), the model predicts the next action \(\hat a_{t+1}\). 
An MLLM evaluates the prediction against the ground truth \(a^\star_{t+1}\): skill must match exactly (\(s_{\text{skill}}\in\{0,1\}\)), while object and parameters are scored by visual reasonableness (\(s_{\text{obj}}, s_{\text{param}}\in\{0,0.5,1\}\)). 
The per-sample score is \(\mathrm{NextStep}=(s_{\text{skill}}+s_{\text{obj}}+s_{\text{param}})/3\).

\textbf{Q3 -- Task state estimation.}
Given \(I_{t-1}\) or \(I_{t}\), the model predicts whether the subtask $\hat a_t$ has been achieved. Performance is measured using binary accuracy against the ground-truth task state.

\section{Experiment}
\definecolor{lightgray}{gray}{.9}
\definecolor{lightblue}{RGB}{230,240,255}
\definecolor{lightgreen}{RGB}{230,255,230}
\definecolor{lightyellow}{RGB}{255,255,230}
\definecolor{lightred}{RGB}{255,230,230}
\newcommand{\lb}[1]{\bf{\color{lightblue}{#1}}}
\newcommand{\ly}[1]{\bf{\color{lightyellow}{#1}}}
\newcommand{\lr}[1]{\bf{\color{lightred}{#1}}}

\definecolor{lightlightgray}{gray}{.95}
\definecolor{lightlightblue}{RGB}{240,245,255}
\definecolor{lightlightgreen}{RGB}{240,255,240}
\definecolor{lightlightyellow}{RGB}{255,255,240}
\definecolor{lightlightred}{RGB}{255,240,240}
\newcommand{\llb}[1]{\bf{\color{lightlightblue}{#1}}}
\newcommand{\llg}[1]{\bf{\color{lightlightgreen}{#1}}}
\newcommand{\lly}[1]{\bf{\color{lightlightyellow}{#1}}}
\newcommand{\llr}[1]{\bf{\color{lightlightred}{#1}}}

\definecolor{lightlightlightgray}{gray}{.99}
\definecolor{lightlightlightblue}{RGB}{247,250,255}
\definecolor{lightlightlightgreen}{RGB}{247,255,247}
\definecolor{lightlightlightyellow}{RGB}{255,255,247}
\definecolor{lightlightlightred}{RGB}{255,247,247}
\newcommand{\lllb}[1]{\bf{\color{lightlightlightblue}{#1}}}
\newcommand{\lllg}[1]{\bf{\color{lightlightlightgreen}{#1}}}
\newcommand{\llly}[1]{\bf{\color{lightlightlightyellow}{#1}}}
\newcommand{\lllr}[1]{\bf{\color{lightlightlightred}{#1}}}

\newcommand{\best}[1]{\cellcolor{blue!30}\textbf{#1}}
\newcommand{\second}[1]{\cellcolor{blue!15}{#1}}

\begin{table*}[t]
\huge
\caption{\textbf{Results on Perception Reasoning(\%)} Attr. stands for Attribute; Temp. stands for Temporal; Refer. Comprehen. stands for Reference Comprehension.}
\label{tab:Main_table_left}
\centering
\renewcommand{\arraystretch}{1.05}
\setlength{\tabcolsep}{5mm}
\resizebox{\linewidth}{!}{
\begin{tabular}{lccccccccc}
\toprule

\rowcolor{blue!5}
& \multicolumn{9}{c}{\textbf{Perception Reasoning}} \\

\cmidrule(lr){2-10}

\multirow{2}{*}{\textbf{Model}}
& \multicolumn{2}{c}{\textbf{Robotic-centric}}
& \multicolumn{2}{c}{\textbf{Object-centric}}
& \multicolumn{3}{c}{\textbf{Scene-centric}}
& \textbf{Task-centric}
& \multirow{2}{*}{\textbf{Avg}} \\

\cmidrule(lr){2-3} \cmidrule(lr){4-5} \cmidrule(lr){6-8} \cmidrule(lr){9-9}

& Robot-type & Robot-view
& Static Attr. & Functional Attr.
& Spatial Relation & Temp. Grounding & Causality
& Refer. Comprehen.
& \\

\midrule
\rowcolor{blue!5}
\multicolumn{10}{c}{\textbf{Basic Reference}} \\
\midrule

Human Evaluation
& 80.67 & 79.08 & 43.77 & 83.89 & 70.91 & 51.61 & 91.22 & 93.22 & 74.30 \\

GPT-5.4-text-only
& 25.86 & 28.26 & 8.81 & 45.57 & 32.67 & 22.90 & 34.48 & 18.40 & 27.12 \\

\midrule
\rowcolor{blue!5}
\multicolumn{10}{c}{\textbf{Closed-Source MLLMs}} \\
\midrule

GPT-5.4
& 73.28 & \second{50.00} & 42.86 & 73.42 & 54.46 & 38.93 & 45.52 & 71.17 & 56.20 \\

GPT-5.2
& 68.10 & 39.86 & 38.60 & 77.22 & 47.52 & 30.53 & 54.48 & 71.17 & 53.44 \\

GPT-5
& 64.66 & 47.10 & 49.24 & 69.62 & 54.46 & \second{48.09} & 74.48 & 78.53 & 60.77 \\

GPT-4.1
& 66.38 & \second{50.00} & 40.43 & 68.35 & 47.52 & 22.14 & 56.55 & 73.01 & 53.05 \\

GPT-4o
& \second{75.00} & 39.13 & 18.24 & 60.76 & 49.50 & 22.14 & 43.45 & 55.21 & 45.43 \\

Claude-Opus-4.7
& \best{76.72} & \best{53.62} & 57.14 & \second{81.01} & 48.51 & 46.56 & 65.52 & 71.78 & 62.61 \\

Claude-Sonnet-4.6
& 53.45 & 47.83 & 53.80 & 69.62 & 52.48 & 29.01 & 57.24 & 69.33 & 54.09 \\

Claude-Sonnet-4.5
& 46.55 & 33.33 & 37.08 & 72.15 & 48.51 & 33.59 & 51.72 & 36.81 & 44.97 \\

Claude-Haiku-4.5
& 44.83 & 33.33 & 30.70 & 56.96 & 25.74 & 22.14 & 45.52 & 27.61 & 35.85 \\

Gemini-3.1-Pro
& 71.55 & 49.28 & \best{66.26} & 78.48 & \best{61.90} & 31.93 & \best{88.97} & \best{90.18} & \best{67.32} \\

Gemini-2.5-Pro
& 67.24 & 43.48 & 57.14 & \best{82.28} & \second{57.43} & \best{50.38} & 73.10 & \second{80.37} & \second{63.93} \\

Gemini-2.5-Flash
& 66.38 & 34.78 & \second{57.75} & 74.68 & 55.45 & 34.92 & \second{75.17} & 76.69 & 59.48 \\

\midrule
\rowcolor{blue!5}
\multicolumn{10}{c}{\textbf{Open-Source Multi-Image MLLMs}} \\
\midrule

Qwen3-VL-8B
& 52.59 & 36.96 & 27.66 & 65.82 & 36.63 & 25.95 & 31.72 & 54.60 & 41.49 \\

Qwen2.5-VL-7B-Ins
& 37.07 & 23.19 & 24.32 & 56.96 & 26.73 & 22.14 & 33.10 & 34.36 & 32.23 \\

LLaVA-OneVision-7B
& 31.03 & 26.81 & 39.21 & 68.35 & 42.57 & 18.32 & 33.79 & 50.92 & 38.88 \\

\midrule
\rowcolor{blue!5}
\multicolumn{10}{c}{\textbf{Embodied MLLMs}} \\
\midrule

RoboBrain-2.0-7B
& 31.90 & 19.57 & 28.57 & 44.30 & 34.65 & 21.37 & 24.83 & 33.13 & 29.79 \\

RoboBrain-2.5-4B
& 35.34 & 24.64 & 39.82 & 77.22 & 53.47 & 18.32 & 59.31 & 44.17 & 44.04 \\

MiMo-Embodied-7B
& 25.86 & 21.74 & 32.22 & 65.82 & 49.50 & 24.43 & 57.93 & 43.56 & 40.13 \\
\bottomrule
\end{tabular}
}
\end{table*}

We evaluate several MLLMs on RoboBench, including closed-source MLLMs, open-source multi-image MLLMs, and open-source embodied MLLMs. Detailed descriptions of the models are provided in Appendix~C. As additional references, we also report a text-only LLM baseline and a human upper bound; the full human-evaluation protocol is detailed in Appendix~D.1, and we validate that our automated evaluator aligns with these human judgments in the In-Depth Analysis (Section~\ref{subsec:in_depth}).

\subsection{Overall Results}

\noindent \textbf{Large capability gaps, yet the frontier keeps advancing}:
Overall performance varies widely across the \nummodels{} evaluated models, yet the most capable systems keep improving with each new generation. Gemini-3.1-Pro shows the most consistent advantages across the perception, affordance, and failure dimensions, clearly outperforming both other closed-source and open-source MLLMs (e.g., $67.32$ perception average vs.\ the next-best $63.93$ of Gemini-2.5-Pro). The strongest MLLMs already answer many embodied questions with reasonable quality—Gemini-3.1-Pro reaches $85.70$ on affordance prediction and $67.32$ on perception reasoning—indicating a solid foundation of embodied cognition, with only a modest residual gap to the human reference, whereas most other MLLMs remain highly uneven or generally weak.\looseness=-1

\noindent \textbf{Closed-source models outperform open-source models}:
As summarized in Tables~\ref{tab:Main_table_left} and~\ref{tab:Main_table_middle}, closed-source models lead in every dimension by about $20$ points on average (${\sim}50\%$ relative), with the widest margins in instruction comprehension and generalized planning (roughly $28$ and $25$ points) and the narrowest in failure analysis (about $13$ points); even there, the best open-source models still trail the strongest closed-source systems. Within the same family, performance improves consistently with model size and generation.\looseness=-1

\subsection{Fine-grained Results across Tasks}

\begin{table*}[tb]
\huge
\caption{Results on Instruction Comprehension and Generalized Planning Q1.}
\label{tab:Main_table_middle}
\centering
\renewcommand{\arraystretch}{1.15}
\setlength{\tabcolsep}{5mm}
\resizebox{\linewidth}{!}{
\begin{tabular}{lcccccccccccccc} 
\toprule

\rowcolor{blue!5}
\textbf{Model} 
& \multicolumn{3}{c}{\textbf{Instruction Comprehension}} 
& \multicolumn{11}{c}{\textbf{Generalized Planning}} \\

\cmidrule(lr){1-1} \cmidrule(lr){2-4} \cmidrule(lr){5-15}

& \textbf{Explicit} & \textbf{Implicit} & \textbf{Avg} 
& \multicolumn{4}{c}{\textbf{Cross-Embodiment Planning}} 
& \multicolumn{3}{c}{\textbf{Cross-Object Planning}}  
& \multicolumn{2}{c}{\textbf{Cross-View Planning}} 
& \multicolumn{1}{c}{\textbf{Cross-Task Planning}}
& \multicolumn{1}{c}{\textbf{Avg}}\\

\cmidrule(lr){5-8} \cmidrule(lr){9-11} \cmidrule(lr){12-13} \cmidrule(lr){14-15}

& \multicolumn{3}{c}{} 
& \textbf{Single-arm} & \textbf{Dual-arm} & \textbf{Mobile-manip.} & \textbf{Human} 
& \textbf{Material Afford.} & \textbf{Physical Attr.} & \textbf{World Knowl.} 
& \textbf{Multi} & \textbf{Single}
& \textbf{Navigation Plan.} & \textbf{} \\

\midrule
\rowcolor{blue!5}
\multicolumn{15}{c}{\textbf{Basic Reference}} \\
\midrule

Human Evaluation
& 59.94 & 61.13 & 60.54
& 72.50 & 41.93 & 41.55 & 62.28
& 56.70 & 58.98 & 49.36
& 52.82 & 51.59
& 45.23 & 54.50 \\

GPT-5.4-text-only
& 74.88 & 38.54 & 56.71 & 83.53 & 66.47 & 74.71 & 62.65 & 76.03 & 80.36 & 66.95 & 71.33 & 72.41 & 47.66 & 73.74 \\

\midrule
\rowcolor{blue!5}
\multicolumn{15}{c}{\textbf{Closed-Source MLLMs}} \\
\midrule

GPT-5.4
& 74.58 & 50.80 & 62.69 & 85.56 & 70.28 & 75.59 & 56.26 & 78.64 & 64.83 & 72.44 & 73.33 & 72.50 & 53.92 & 70.91 \\

GPT-5.2
& 75.90 & 48.85 & 62.38 & 86.92 & 70.00 & 80.38 & 60.08 & 81.60 & 64.62 & 75.49 & 75.84 & 71.22 & 55.03 & 72.31 \\

GPT-5
& \second{77.63} & 54.71 & 66.17 & 84.25 & 69.48 & 81.27 & \best{70.13} & 81.58 & 59.80 & 71.95 & 70.95 & 72.76 & 58.29 & 71.84 \\

GPT-4.1
& 76.23 & 57.30 & 66.77 & 88.08 & 68.56 & 78.17 & 55.38 & 81.37 & 65.76 & 64.88 & 73.70 & 71.83 & 52.94 & 71.79 \\

GPT-4o
& 74.22 & 54.90 & 64.56 & 86.02 & 66.40 & 76.44 & 62.31 & 80.86 & \second{72.63} & 73.78 & 64.50 & 65.63 & 57.62 & 73.46 \\

Claude-Opus-4.7
& 73.92 & \best{61.52} & 67.72 & \best{89.43} & 71.98 & 84.30 & \second{68.53} & 84.56 & 67.70 & 70.00 & 75.06 & 72.06 & \second{59.53} & 75.54 \\

Claude-Sonnet-4.6
& \best{79.94} & \second{61.38} & \best{70.66} & 88.89 & 73.93 & \best{84.42} & 66.19 & \second{84.76} & \best{79.10} & 73.54 & \second{80.55} & \best{77.90} & \best{67.06} & \best{79.38} \\

Claude-Sonnet-4.5
& 76.65 & 53.62 & 65.13 & \second{89.11} & \best{75.06} & 81.70 & 64.27 & \best{85.10} & 69.70 & \second{76.22} & \best{82.17} & 75.06 & 59.52 & \second{76.62} \\

Claude-Haiku-4.5
& 73.78 & 42.88 & 58.33 & 86.13 & \second{74.07} & 76.63 & 60.38 & 81.37 & 58.93 & 71.75 & 78.48 & 71.73 & 50.87 & 71.01 \\

Gemini-3.1-Pro
& 73.25 & 59.90 & 66.58 & 80.64 & 69.85 & 79.90 & 50.13 & 73.11 & 71.66 & 69.63 & 74.64 & 74.39 & 57.14 & 70.71 \\

Gemini-2.5-Pro
& 76.20 & 60.96 & \second{68.58} & 83.53 & 69.08 & \second{84.31} & 59.16 & 76.72 & 66.93 & \best{77.68} & 73.57 & \second{75.24} & 55.34 & 71.50 \\

Gemini-2.5-Flash
& 71.45 & 49.90 & 60.67 & 83.58 & 69.41 & 81.06 & 58.43 & 75.72 & 70.76 & 74.88 & 72.14 & 72.65 & 55.08 & 70.98 \\

\midrule
\rowcolor{blue!5}
\multicolumn{15}{c}{\textbf{Open-Source Multi-Image MLLMs}} \\
\midrule

Qwen3-VL-8B
& 59.46 & 30.80 & 45.13 & 74.49 & 44.54 & 57.98 & 52.25 & 63.88 & 54.66 & 55.49 & 49.15 & 58.75 & 37.22 & 56.71 \\

Qwen2.5-VL-7B-Ins
& 56.04 & 23.90 & 39.97 & 73.89 & 30.19 & 56.06 & 53.85 & 58.90 & 57.78 & 53.90 & 25.83 & 37.50 & 11.95 & 49.92 \\

LLaVA-OneVision-7B
& 38.25 & 10.61 & 24.43 & 54.87 & 31.05 & 35.88 & 43.99 & 37.59 & 51.37 & 30.00 & 31.43 & 36.60 & 25.11 & 41.02 \\

\midrule
\rowcolor{blue!5}
\multicolumn{15}{c}{\textbf{Embodied MLLMs}} \\
\midrule

RoboBrain-2.0-7B
& 43.54 & 21.10 & 32.32 & 62.49 & 30.16 & 44.42 & 42.90 & 46.62 & 52.87 & 45.24 & 31.25 & 32.69 & 25.98 & 45.12 \\

RoboBrain-2.5-4B
& 36.30 & 16.65 & 26.47 & 39.32 & 23.99 & 45.87 & 54.53 & 31.69 & 29.16 & 24.39 & 28.39 & 25.75 & 23.97 & 31.85 \\

MiMo-Embodied-7B
& 66.87 & 37.30 & 52.09 & 82.20 & 37.11 & 61.76 & 63.03 & 73.05 & 66.85 & 70.88 & 58.95 & 43.88 & 28.71 & 62.72 \\

\bottomrule
\end{tabular}
}
\end{table*}

We further break down results across tasks, highlighting distinct weaknesses in instruction comprehension, perception, planning, and failure analysis.

\noindent \textbf{Commonsense vs. Embodied Grounding}:
Before analyzing individual dimensions, we verify that RoboBench cannot be solved by language priors alone: a text-only baseline (GPT-5.4 without images) stays close to random on perception and affordance multiple-choice tasks---a $27.12$ perception and $25.61$ affordance average, far below the $67.32$ and $85.70$ of the best vision-conditioned MLLM (Gemini-3.1-Pro)---and trails on all questions that require scene evidence. This confirms that RoboBench questions are constructed to demand grounding in the observed scene---visual state, embodiment, and physical feasibility---rather than commonsense recall, so the scores and gaps dissected below genuinely measure embodied capability.

\noindent \textbf{Implicit vs. Explicit Instruction Comprehension}:
Models perform significantly worse on implicit instructions compared to explicit ones, even when images and ground truth remain the same: the strongest model on explicit goals (Claude-Sonnet-4.6) drops from $79.94$ to $61.38$ on implicit ones, and the gap widens further for weaker MLLMs. This reveals a clear weakness in grounding indirect human demands into actionable goals, likely because current MLLMs fail to fully integrate scene context when inferring task intent. Notably, a paired diagnostic ablation in Appendix~D.3 shows that prompting models to first rewrite the implicit instruction into an explicit goal via chain-of-thought does not reliably close this gap, indicating a genuine intent-grounding limitation rather than an artifact of prompting. Future models must jointly reason over language, perception, and context to infer and execute human intent.

\noindent \textbf{Perception Challenges}:
While models perform reasonably on object property analysis---Gemini-2.5-Pro reaches $82.28$ on functional-attribute reasoning---they struggle with basic robotic perception and spatiotemporal reasoning: the best per-task scores are only $53.62$ on robot-view understanding (Claude-Opus-4.7) and $50.38$ on temporal grounding (Gemini-2.5-Pro), the two weakest perception tasks. Common failures include misidentifying robot type and viewpoint, or failing to reason under different reference frames. These gaps suggest that future models should incorporate stronger embodiment-aware perception modules and explicit spatiotemporal reasoning.  

\noindent \textbf{Planning Limitations}:
As shown in Table~\ref{tab:Main_table_middle}, models exhibit limited capacity in complex planning scenarios.
Cross-Embodiment: Current models, largely trained on single-arm settings, often fail to coordinate dual-arm actions or perform mobile manipulation, leading to poor spatial search and movement decisions.  
Cross-Object: Performance drops sharply when tasks involve uncommon objects, symbolic reasoning, or world knowledge, suggesting weak integration of heterogeneous information, even though results on common objects remain acceptable.  
Cross-View: When the front-view camera is occluded, multi-view inputs effectively improve model performance. This underscores the need to strengthen multi-view reasoning capabilities, thereby enhancing planning robustness in real-world settings.\looseness=-1

\begin{table*}[t]
\centering

\begin{minipage}[t]{0.49\linewidth}
\captionsetup{type=table}
\captionof{table}{Results on Planning Q2, Q3. Instr. Compre. stands for Instruction Comprehension.}
\label{tab:q2q3_left}
\centering
\small
\renewcommand{\arraystretch}{1.08}
\setlength{\tabcolsep}{3.5pt}
\resizebox{\linewidth}{!}{%
\begin{tabular}{lcccccccc}
\toprule
\rowcolor{blue!5}
\textbf{Model}
& \multicolumn{2}{c}{\textbf{Instr. Compre.}}
& \multicolumn{6}{c}{\textbf{Generalized Planning}} \\
\cmidrule(lr){2-3}\cmidrule(lr){4-9}
& \multicolumn{2}{c}{Explicit Goal}
& \multicolumn{2}{c}{Single Arm}
& \multicolumn{2}{c}{Mater. Afford.}
& \multicolumn{2}{c}{World Knowl.} \\
\cmidrule(lr){2-3}\cmidrule(lr){4-5}\cmidrule(lr){6-7}\cmidrule(lr){8-9}
& Q2 & Q3
& Q2 & Q3
& Q2 & Q3
& Q2 & Q3 \\
\midrule
\rowcolor{blue!5}
\multicolumn{9}{c}{\textbf{Basic Reference}} \\
\midrule

Human Evaluation
& 45.28 & 74.32 & 27.52 & 71.35 & 43.62 & 71.20 & 43.89 & 69.83 \\

GPT-5.4-text-only
& 36.98 & 46.25 & 40.98 & 52.86 & 40.43 & 52.33 & 43.01 & 41.46 \\

\midrule
\rowcolor{blue!5}
\multicolumn{9}{c}{\textbf{Closed-Source MLLMs}} \\
\midrule

GPT-5.4
& \second{49.48} & 62.50 & 48.20 & 67.85 & 44.38 & 64.67 & 42.19 & 51.22 \\

GPT-5.2
& 39.32 & \second{75.00} & 42.86 & \second{73.02} & 41.07 & 66.67 & 37.50 & 56.10 \\

GPT-5
& 44.09 & 72.97 & 47.26 & \best{75.75} & 44.38 & 62.83 & 39.58 & 63.41 \\

GPT-4.1
& 45.31 & 70.00 & 48.62 & 63.76 & 44.32 & 63.67 & 39.58 & 58.54 \\

GPT-4o
& 42.86 & 65.00 & 43.94 & 59.95 & 41.23 & 55.33 & 42.19 & 51.22 \\

Claude-Opus-4.7
& 44.53 & 67.50 & \second{51.08} & 65.12 & 45.97 & 63.00 & \second{45.31} & 63.41 \\

Claude-Sonnet-4.6
& 43.75 & 70.00 & 43.22 & 66.21 & 43.90 & 64.33 & 39.58 & 60.98 \\

Claude-Sonnet-4.5
& 41.67 & 61.25 & 44.73 & 56.68 & 41.99 & 54.00 & 41.15 & 48.78 \\

Claude-Haiku-4.5
& 34.13 & 62.50 & 43.07 & 60.38 & 38.14 & 63.18 & 31.77 & \best{70.73} \\

Gemini-3.1-Pro
& 44.79 & 71.25 & 50.65 & 64.58 & \second{46.97} & 67.67 & 26.56 & 65.85 \\

Gemini-2.5-Pro
& 37.76 & 72.50 & \best{52.02} & 71.66 & \best{49.20} & \best{70.00} & 35.42 & \second{68.29} \\

Gemini-2.5-Flash
& 47.66 & \best{77.50} & 49.85 & 54.22 & 44.81 & \second{68.83} & \second{45.31} & \second{68.29} \\

\midrule
\rowcolor{blue!5}
\multicolumn{9}{c}{\textbf{Open-Source Multi-Image MLLMs}} \\
\midrule

Qwen3-VL-8B
& \best{49.74} & 62.50 & 50.94 & 61.58 & 44.80 & 55.50 & 36.98 & 56.10 \\

Qwen2.5-VL-7B-Ins
& 29.43 & 55.00 & 31.88 & 52.59 & 30.17 & 49.33 & 20.31 & 51.22 \\

LLaVA-OneVision-7B
& 33.59 & 41.25 & 35.06 & 46.05 & 35.97 & 40.50 & 34.90 & 43.90 \\

\midrule
\rowcolor{blue!5}
\multicolumn{9}{c}{\textbf{Embodied MLLMs}} \\
\midrule

RoboBrain-2.0-7B
& 34.90 & 52.50 & 33.84 & 53.95 & 34.29 & 49.00 & 27.60 & 48.78 \\

RoboBrain-2.5-4B
& 32.81 & 58.75 & 37.37 & 54.50 & 35.25 & 55.17 & 32.29 & 56.10 \\

MiMo-Embodied-7B
& 38.17 & 55.00 & 43.87 & 55.31 & 42.90 & 51.50 & \best{46.88} & 63.41 \\

\bottomrule
\end{tabular}
}%
\end{minipage}%
\hfill
\begin{minipage}[t]{0.495\linewidth}
\captionsetup{type=table}
\captionof{table}{Results on Affordance Prediction and Failure Analysis. Naviga. stands for Navigation.}
\label{tab:afford_fail_right}
\centering
\small
\renewcommand{\arraystretch}{1.08}
\setlength{\tabcolsep}{3.5pt}
\resizebox{\linewidth}{!}{%
\begin{tabular}{lccccccc}
\toprule
\rowcolor{blue!5}
\textbf{Model}
& \multicolumn{4}{c}{\textbf{Affordance Prediction}}
& \multicolumn{3}{c}{\textbf{Failure Analysis}} \\
\cmidrule(lr){1-1}\cmidrule(lr){2-5}\cmidrule(lr){6-8}
& Static & Dynamic & Naviga. & Avg
& Execution & Planning & Avg \\
\multicolumn{8}{c}{} \\
\midrule
\rowcolor{blue!5}
\multicolumn{8}{c}{\textbf{Basic Reference}} \\
\midrule

Human Evaluation
& 86.08 & 80.02 & 81.85 & 82.63 & 47.30 & 80.67 & 63.99 \\

GPT-5.4-text-only
& 23.81 & 27.52 & 25.51 & 25.61 & 11.92 & 32.64 & 22.28 \\

\midrule
\rowcolor{blue!5}
\multicolumn{8}{c}{\textbf{Closed-Source MLLMs}} \\
\midrule

GPT-5.4
& 44.22 & 36.91 & 58.16 & 46.43 & 26.49 & 65.97 & 46.23 \\

GPT-5.2
& 43.54 & 39.60 & 47.96 & 43.70 & 26.49 & 68.06 & 47.27 \\

GPT-5
& 62.59 & 49.66 & 62.24 & 58.16 & 19.87 & \second{80.56} & \second{50.21} \\

GPT-4.1
& 29.93 & 42.95 & 68.37 & 47.08 & 20.53 & 70.83 & 45.68 \\

GPT-4o
& 40.82 & 42.28 & 50.00 & 44.37 & \second{31.79} & 57.64 & 44.71 \\

Claude-Opus-4.7
& 53.74 & 62.42 & 79.59 & 65.25 & 14.57 & 72.22 & 43.40 \\

Claude-Sonnet-4.6
& 37.41 & 52.35 & 41.84 & 43.87 & 17.88 & 77.78 & 47.83 \\

Claude-Sonnet-4.5
& 34.69 & 38.93 & 53.06 & 42.23 & 14.57 & 63.19 & 38.88 \\

Claude-Haiku-4.5
& 27.89 & 26.17 & 21.43 & 25.16 & 17.22 & 45.83 & 31.53 \\

Gemini-3.1-Pro
& \best{82.31} & \best{77.85} & \best{96.94} & \best{85.70} & 25.17 & \best{80.74} & \best{52.95} \\

Gemini-2.5-Pro
& \second{65.99} & 61.07 & \second{93.88} & \second{73.65} & 18.54 & 72.22 & 45.38 \\

Gemini-2.5-Flash
& 61.22 & \second{69.80} & 36.73 & 55.92 & 25.83 & 65.49 & 45.66 \\

\midrule
\rowcolor{blue!5}
\multicolumn{8}{c}{\textbf{Open-Source Multi-Image MLLMs}} \\
\midrule

Qwen3-VL-8B
& 23.81 & 17.45 & 22.45 & 21.24 & 22.52 & 55.56 & 39.04 \\

Qwen2.5-VL-7B-Ins
& 18.37 & 31.54 & 26.53 & 25.48 & 13.91 & 35.42 & 24.66 \\

LLaVA-OneVision-7B
& 38.78 & 33.56 & 66.33 & 46.22 & 20.53 & 31.25 & 25.89 \\

\midrule
\rowcolor{blue!5}
\multicolumn{8}{c}{\textbf{Embodied MLLMs}} \\
\midrule

RoboBrain-2.0-7B
& 31.97 & 27.52 & 31.63 & 30.37 & 15.23 & 40.28 & 27.75 \\

RoboBrain-2.5-4B
& 50.34 & 21.48 & 72.45 & 48.09 & \best{43.71} & 46.53 & 45.12 \\

MiMo-Embodied-7B
& 51.70 & 36.91 & 70.41 & 53.01 & 19.21 & 42.36 & 30.78 \\
\bottomrule
\end{tabular}
}%
\end{minipage}

\end{table*}

\noindent \textbf{Failure Analysis}:  
As reported in Table~\ref{tab:afford_fail_right}, analyzing execution errors proves harder than diagnosing planning errors, with scores consistently lower on the former: even the best score reaches only $43.71$ on execution-error diagnosis (RoboBrain-2.5-4B, an embodied-finetuned model) while most models fall between $15$ and $27$, whereas planning-error diagnosis reaches $80.74$ (Gemini-3.1-Pro); the human upper bound exhibits the same asymmetry ($47.30$ versus $80.67$). Execution-level failures often require fine-grained distinctions—such as separating position errors from rotation errors (correct position but wrong gripper angle)—which demand expert-level embodied knowledge. 
This suggests that fine-grained perception is crucial for improving error diagnosis.

\subsection{In-Depth Analysis}
\label{subsec:in_depth}
\noindent \textbf{Planning Error Analysis}:
To understand the limitations of current multimodal models as embodied planners, we analyze representative failure cases across RoboBench tasks, and find that most failures stem not from misunderstanding instructions but from difficulty grounding semantic reasoning into executable actions. We categorize common failures into four types (Fig.~\ref{fig:error_analysis}; definitions in Appendix~D.6).
\textbf{Execution Errors} account for the majority of failures (45\%), arising from incomplete or incorrect action sequences such as missing steps or wrong functions. 
\textbf{Identification Errors} (24\%) occur when models confuse objects or assign incorrect parameters. 
\textbf{Common Sense Errors} (25\%) involve violations of physical or spatial constraints. 
\textbf{Mode-Specific Errors} (6\%) result from failing to follow structured output conventions.
These results highlight a persistent perception–action gap: models can generate plausible reasoning but struggle to consistently translate it into executable embodied actions.

\begin{figure}[t]
\centering
\begin{minipage}{0.40\textwidth}
    \centering
    \includegraphics[width=\linewidth]{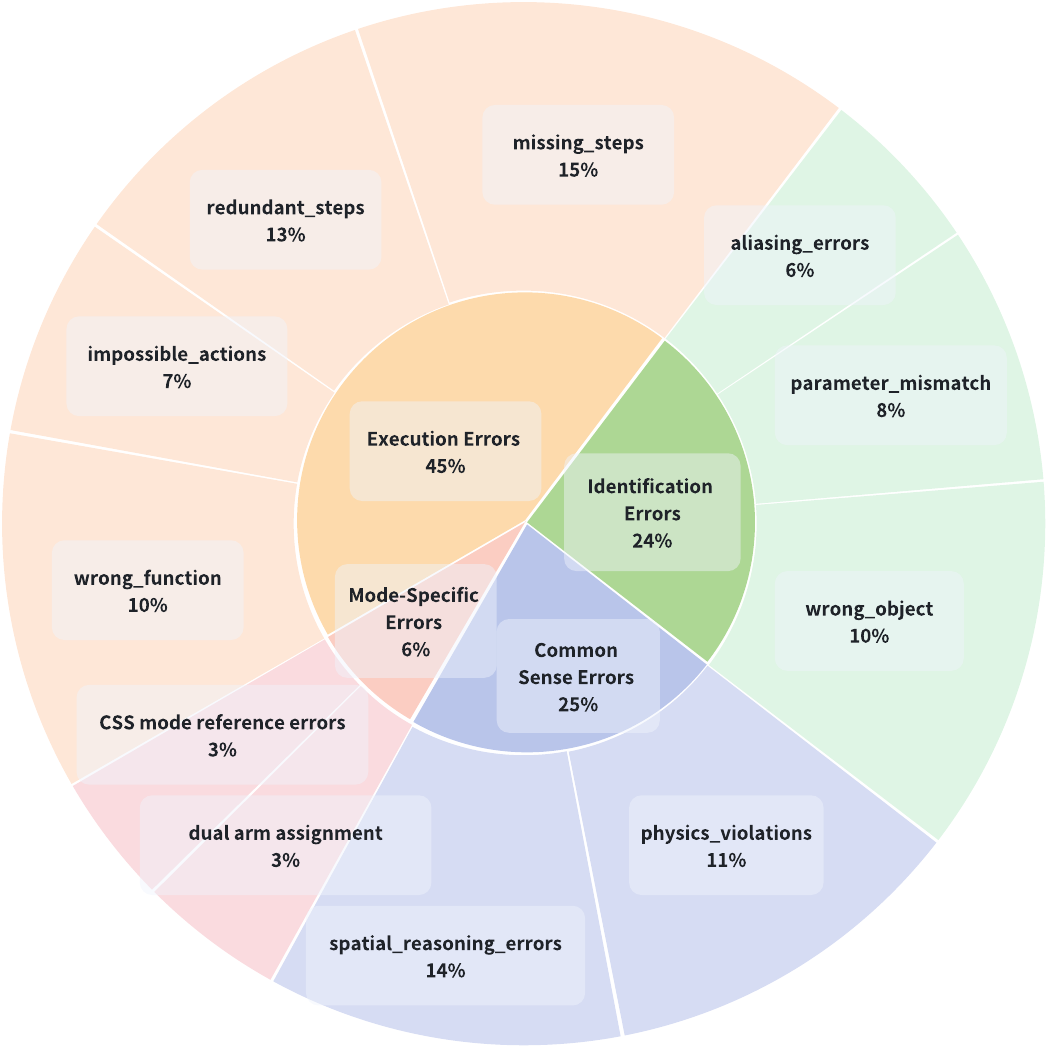}
\end{minipage}
\hfill
\begin{minipage}{0.56\textwidth}
    \centering
    \includegraphics[width=\linewidth, trim=6cm 0cm 6cm 0cm, clip]{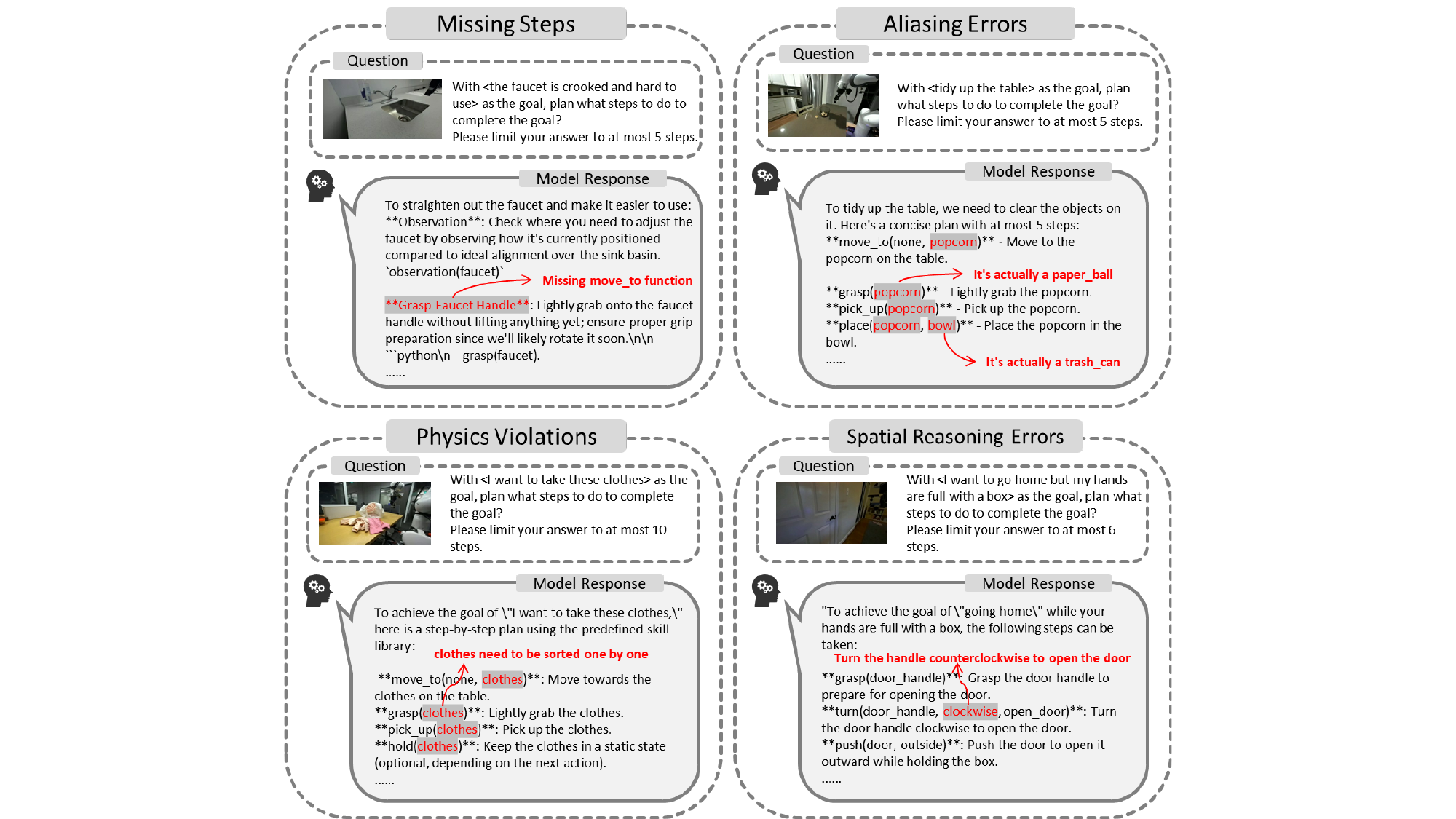}
\end{minipage}
\caption{\textbf{Planning error analysis.}
Left: distribution of error types in RoboBench planning tasks.
Right: representative failure cases illustrating common planning errors.}
\label{fig:error_analysis}
\end{figure}

\begin{figure}[t]
    \centering
    \begin{subfigure}{0.48\textwidth}
        \centering
        \includegraphics[width=\linewidth]{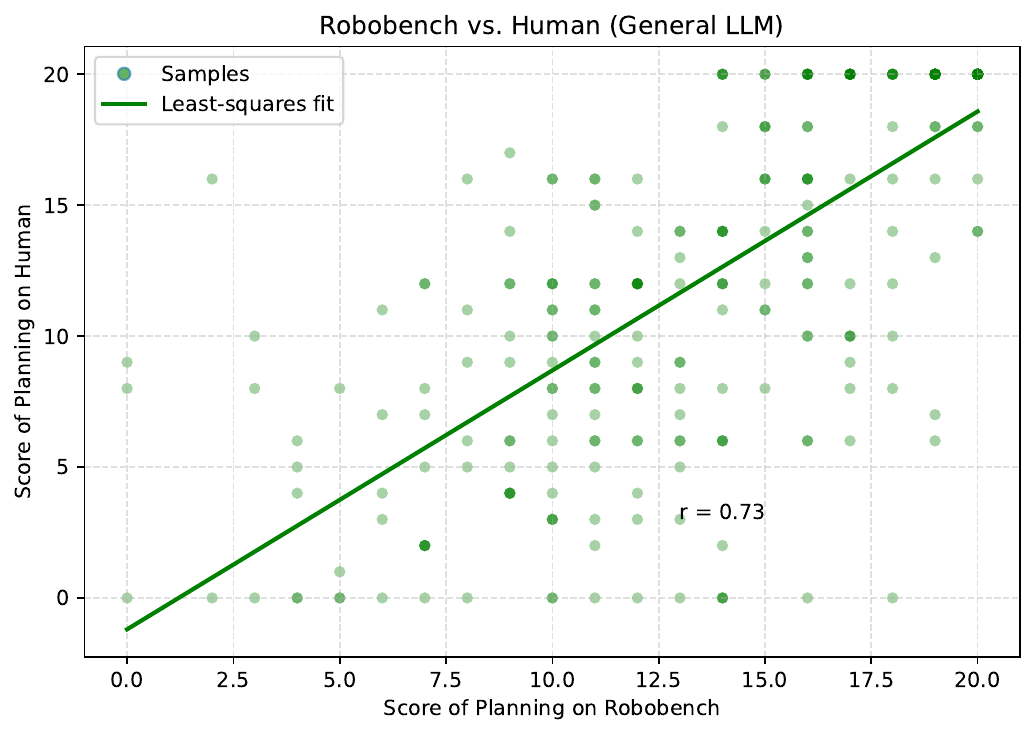}
        \caption{Baseline evaluation framework.}
        \label{fig:sub2}
    \end{subfigure}
    \hfill
    \begin{subfigure}{0.48\textwidth}
        \centering
        \includegraphics[width=\linewidth]{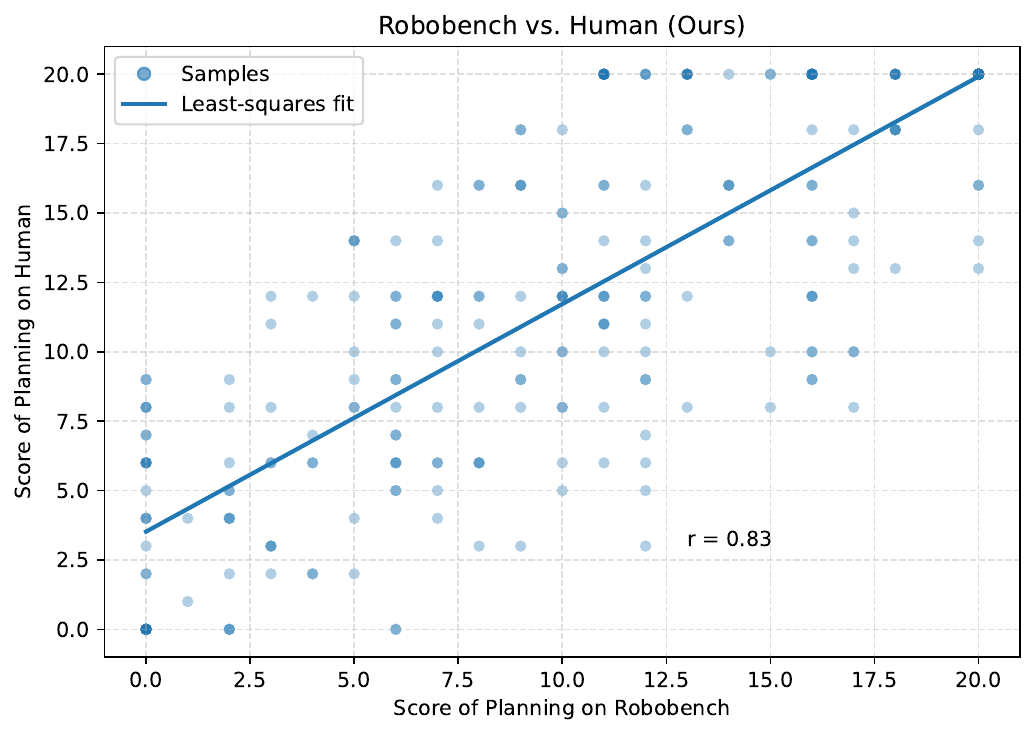}
        \caption{MLLM-as-world-simulator evaluation.}
        \label{fig:sub3}
    \end{subfigure}
    \caption{Alignment between automated evaluation scores and human judgments. Pearson correlation coefficient ($r$) measures the consistency between each evaluation method and human evaluation. Our MLLM-as-world-simulator evaluation shows stronger alignment with human judgments than the baseline LLM-based scoring method.}
    \label{fig:three_subfigs}

    \includegraphics[width=0.86\textwidth]{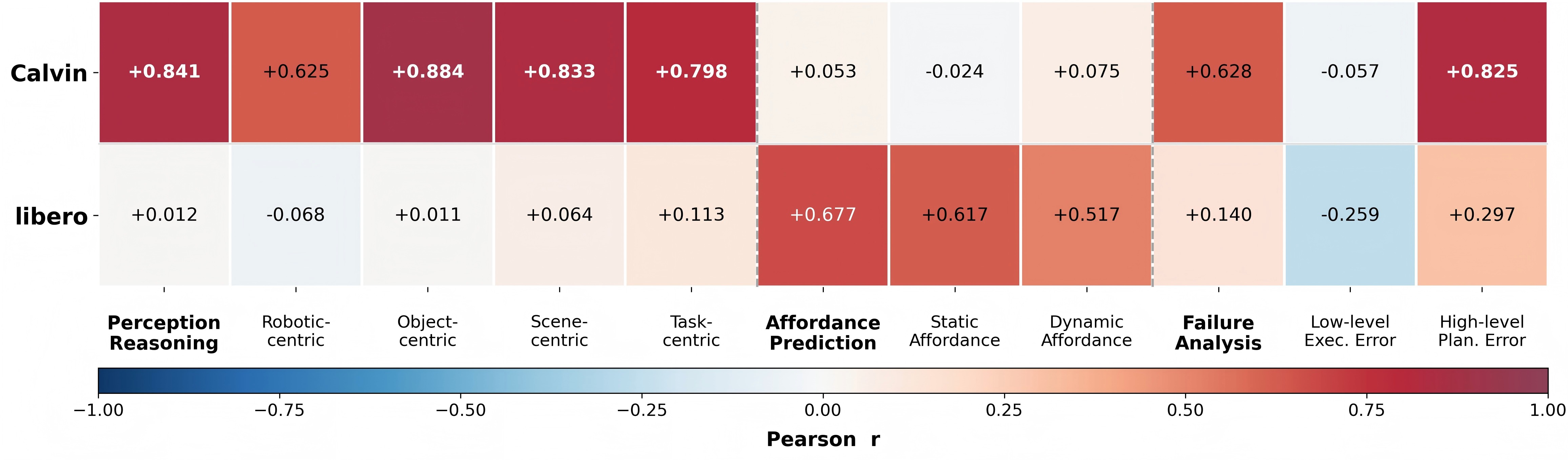}
    \caption{Correlation between RoboBench cognitive abilities and downstream VLA performance. Results show that perception reasoning strongly correlates with long-horizon manipulation in CALVIN, while affordance prediction is more related to fine-grained manipulation tasks in LIBERO-10.}
    \label{fig:robobench_vla_correlation}
\end{figure}

\noindent \textbf{Human Alignment of the Evaluation Framework}: To validate reliability, we compare our framework with a baseline LLM-based scoring method~\cite{chi2024evaembodiedworldmodel} that judges criteria such as planning rationality and object correctness, whereas we judge whether the plan, when executed, completes the task. As shown in Fig.~\ref{fig:three_subfigs}, our MLLM-as-world-simulator achieves stronger alignment with human judgments ($r=0.83$) than the baseline ($r=0.73$). By scoring plans at structured skill–object nodes guided by the underlying DAG, it improves consistency with humans while reducing the bias of free-form language scoring.
The protocol is also robust to grader and prompt: across four judge models and three prompt variants, NodeCorrectness remains stable (ICC(2,k)$=0.948$), confirming that the protocol—not a particular model or phrasing—drives the evaluation (Appendix~D.2); it also tolerates ground-truth annotation noise (Appendix~D.4).\looseness=-1

\noindent \textbf{Correlation between RoboBench and downstream VLA performance}:
Following VLM4VLA~\cite{zhang2026vlm4vla}, we convert several open-source VLMs into VLA policies via minimal fine-tuning and evaluate them on two manipulation benchmarks, CALVIN~\cite{mees2022calvin} and LIBERO-10~\cite{liu2023libero}. This ensures that downstream VLA performance largely reflects the intrinsic capabilities of the VLM backbone, with implementation details in Appendix~D.5.

We correlate RoboBench scores with downstream VLA success rates across three dimensions: \textit{Perception Reasoning}, \textit{Affordance Prediction}, and \textit{Failure Analysis}. As shown in Fig.~\ref{fig:robobench_vla_correlation}, perception reasoning strongly correlates with CALVIN (object-centric $r=0.884$, scene-centric $r=0.833$), reflecting the role of object and scene understanding in long-horizon manipulation, whereas LIBERO correlates more with affordance prediction (static+dynamic $r=0.677$), indicating greater reliance on fine-grained interaction dynamics. Thus different benchmarks emphasize different abilities, and RoboBench helps identify VLM capabilities predictive of VLA performance.

\section{Conclusion}

We presented RoboBench, a benchmark to systematically evaluate MLLMs as the embodied brain for robotic manipulation. It traces the full execution pipeline across five dimensions—Instruction Comprehension, Perception Reasoning, Generalized Planning, Affordance Prediction, and Failure Analysis—covering 25 tasks with 6092 QA pairs, and contributes a world-simulator framework that evaluates not only symbolic plan fidelity but also embodied feasibility. Experiments expose major gaps across all five dimensions and reveal the correlation between embodied cognitive abilities and downstream VLA performance. RoboBench thus provides a unified scaffold for measuring embodied cognition and guiding the development of more robust, generalizable embodied intelligence.

\clearpage
\begin{center}
    {\Large\bfseries Supplementary Material\par}
\end{center}
\bigskip
\appendix

\section{Appendix Overview}
This appendix provides additional material that complements the main paper, spanning extended related work, the full model roster, deeper analyses, dataset statistics, qualitative case studies, and the complete prompt set:
\begin{itemize}
    \item \textbf{More Related Work (Appendix~\ref{sec:related_work_supp})}: An expanded survey of embodied MLLMs---end-to-end and hierarchical VLAs and planner-based agent systems---together with a detailed comparison to the closest recent benchmarks, BEAR~\cite{qi2025bear} and ECBench~\cite{dang2025ecbench}.
    \item \textbf{Baseline Models (Appendix~\ref{sec:baseline})}: The full roster of the \nummodels{} evaluated closed- and open-source models with their exact API endpoints or Hugging Face identifiers for reproducibility.
    \item \textbf{More Experiment Analysis (Appendix~\ref{sec:analysis})}: Our human evaluation protocol; a cross-evaluator robustness study of the MLLM-as-world-simulator (inter-judge ICC(2,$k$) and Cronbach's $\alpha$ across four judge models from three vendors and three scoring-prompt variants); an implicit-instruction chain-of-thought ablation; robustness of the completion score to ground-truth annotation noise; the relationship between grouped embodied cognitive abilities and downstream VLA performance on \textit{Calvin} and \textit{Libero}; and the error taxonomy with per-type failure analysis.
    \item \textbf{More Statistics of Planning Tasks (Appendix~\ref{sec:statistics})}: Annotation quality-control checks, detailed task statistics, and skill, action-length, action-frequency, and data-source distributions.
    \item \textbf{Case Study (Appendix~\ref{sec:case})}: Qualitative examples that illustrate the robustness of the world-simulator evaluation.
    \item \textbf{Prompts (Appendix~\ref{sec:prompts})}: The complete set of prompts used for benchmark data construction, model inference, and evaluation---including the world-simulator scoring prompts---enabling full transparency and reproducibility.
\end{itemize}

\section{More Related Work}
\label{sec:related_work_supp}
Recent advances in multimodal large language models (MLLMs) have delivered strong perception–reasoning–planning capabilities, leading to growing interest in their applications for embodied intelligence~\cite{hurst2024gpt,li2024llavaonevisioneasyvisualtask,Chen_2024_CVPR,lin2023vila,zhang2024videoinstructiontuningsynthetic,wang2024qwen2vlenhancingvisionlanguagemodels,bai2025qwen2.5vl, chen2025exploringembodiedmultimodallarge, luo2024ssdllm, an2024mc, an2025unictokens, lin2024draw, lin2025perceive, chi2025wow, tan2025roboos}. One line of work develops vision-language-action (VLA) systems for language-conditioned manipulation~\cite{octomodelteam2024octoopensourcegeneralistrobot,kim2024openvlaopensourcevisionlanguageactionmodel,black2024pi0visionlanguageactionflowmodel,open_x_embodiment_rt_x_2023}. Early end-to-end VLAs extend MLLMs with action heads (autoregressive~\cite{kim2024openvlaopensourcevisionlanguageactionmodel}, diffusion~\cite{li2024cogact}, flow matching~\cite{black2024pi0visionlanguageactionflowmodel}, or hybrid~\cite{liu2025hybridvla}). While promising, these models often struggle with long-horizon tasks and out-of-distribution generalization. To mitigate these issues, hierarchical VLAs let the MLLM emit explicit, semantically rich guidance for the controller—e.g., next-step predictions~\cite{black2025pi_0.5}, static affordances~\cite{li2024chain_of_affordance}, or spatial trajectories~\cite{huang2025thinkact, lee2025molmoact}—thereby externalizing task knowledge and strengthening transfer. This language-mediated structuring of state, goals, and affordances connects MLLM priors to embodied cognition, improving robustness~\cite{black2025pi_0.5}.
A complementary direction treats the MLLM as a high-level planner that issues explicit constraints and sketches for low-level executors—such as spatial waypoints and temporal anchors~\cite{driess2023palm,huang2023voxposercomposable3dvalue, huang2024rekepspatiotemporalreasoningrelational, han2025tiger} or manipulability regions and trajectories~\cite{huang2024rekepspatiotemporalreasoningrelational,robobrain_2025, RoboBrain2.0TechnicalReport, zhou2025roborefer}. These systems demonstrate strong zero- and few-shot generalization to novel objects and tasks while maintaining robotic feasibility.
Despite this progress, MLLMs' embodied cognition—the ability to plan, reason, and adapt under physical and task constraints—has not been systematically assessed. In this work, we articulate the core capabilities required of embodied brains and provide a systematic evaluation across them.

\paragraph{Comparison with BEAR and ECBench.}
BEAR~\cite{qi2025bear} and ECBench~\cite{dang2025ecbench} are the closest recent benchmarks to RoboBench in scope, but they target different parts of the embodied-cognition space. BEAR evaluates atomic embodied capabilities through 4,469 interleaved image-video-text entries across 14 domains and 6 categories, making it valuable for diagnosing fine-grained skills such as pointing, trajectory understanding, spatial reasoning, and high-level planning. ECBench evaluates 4,324 egocentric RGB-D video QA pairs across 30 embodied-cognition dimensions, with emphasis on robot-centric perception, dynamic scenes, and hallucination. RoboBench differs in three ways. First, its taxonomy follows the manipulation execution pipeline from implicit instruction understanding to perception, generalized planning, affordance grounding, and failure analysis. Second, it centers on real robot manipulation data, including single-arm, dual-arm, mobile manipulation, and human-demonstration settings. Third, its long-horizon planning metric is not a surface text score: the MLLM-as-world-simulator evaluates skill--object nodes and task completion under a human-annotated DAG of state dependencies.

\section{Baseline Models}
\label{sec:baseline}

To ensure fair and comprehensive evaluation, we benchmarked against a diverse set of baseline models spanning both open- and closed-source systems. Table~\ref{tab:model_info} provides a systematic overview of these models, including their accessibility (open vs.\ closed source), creators, and corresponding API endpoints or model identifiers.

\begin{table*}[ht]
\caption{Specific information of the \nummodels{} models evaluated in RoboBench, grouped into closed-source and open-source systems, plus a text-only ablation of GPT-5.4 (listed for completeness but not counted among the \nummodels{} models). GPT-5.4-text-only shares the GPT-5.4 backbone but is queried without visual input as a text-only ablation.}
\footnotesize
\centering
\resizebox{\textwidth}{!}{%
\begin{tabular}{c|c|c|c}
\hline
\textbf{Model Name} & \textbf{Type} & \textbf{Creator} & \textbf{API Name/Huggingface Model ID} \\
\hline
\multicolumn{4}{c}{\textbf{Closed-Source MLLMs}} \\
\hline
GPT-5.4~\cite{openai2026gpt54} & Closed-Source & OpenAI & gpt-5.4 \\
GPT-5.4-text-only~\cite{openai2026gpt54} & Closed-Source & OpenAI & gpt-5.4 (text-only input) \\
GPT-5.2~\cite{openai2025gpt52} & Closed-Source & OpenAI & gpt-5.2 \\
GPT-5~\cite{openai2025gpt5} & Closed-Source & OpenAI & gpt-5 \\
GPT-4.1~\cite{hurst2024gpt} & Closed-Source & OpenAI & gpt-4.1 \\
GPT-4o~\cite{hurst2024gpt} & Closed-Source & OpenAI & gpt-4o-2024-11-20 \\
Claude-Opus-4.7~\cite{anthropic2026claudeopus47} & Closed-Source & Anthropic & claude-opus-4-7 \\
Claude-Sonnet-4.6~\cite{anthropic2026claudesonnet46} & Closed-Source & Anthropic & claude-sonnet-4-6 \\
Claude-Sonnet-4.5~\cite{anthropic2025claudesonnet45} & Closed-Source & Anthropic & claude-sonnet-4-5-20250929 \\
Claude-Haiku-4.5~\cite{anthropic2025claudehaiku45} & Closed-Source & Anthropic & claude-haiku-4-5-20251001 \\
Gemini-3.1-Pro~\cite{gemini3_2026} & Closed-Source & Google & gemini-3.1-pro-preview \\
Gemini-2.5-Pro~\cite{comanici2025gemini2.5} & Closed-Source & Google & gemini-2.5-pro \\
Gemini-2.5-Flash~\cite{comanici2025gemini2.5} & Closed-Source & Google & gemini-2.5-flash \\
\hline
\multicolumn{4}{c}{\textbf{Open-Source MLLMs}} \\
\hline
Qwen3-VL-8B~\cite{qwen3vl2025} & Open-Source & Alibaba & Qwen/Qwen3-VL-8B-Instruct \\
Qwen2.5-VL-7B-Ins~\cite{bai2025qwen2.5vl} & Open-Source & Alibaba & Qwen/Qwen2.5-VL-7B-Instruct \\
LLaVA-OneVision-7B~\cite{li2024llavaonevisioneasyvisualtask} & Open-Source & NTU & llava-hf/llava-onevision-qwen2-7b-ov-hf \\
RoboBrain-2.0-7B~\cite{RoboBrain2.0TechnicalReport} & Open-Source & BAAI & BAAI/RoboBrain2.0-7B \\
RoboBrain-2.5-4B~\cite{robobrain2_5_2026} & Open-Source & BAAI & BAAI/RoboBrain2.5-4B \\
MiMo-Embodied-7B~\cite{xiaomi2025mimo} & Open-Source & Xiaomi & XiaomiMiMo/MiMo-Embodied-7B \\
\hline
\end{tabular}%
}
\label{tab:model_info}
\end{table*}

\section{More Experiment Analysis}
\label{sec:analysis}

\subsection{Human Evaluation Protocol}
\label{subsec:human_eval_protocol}
The human upper bound reported in the main paper is produced on sampled RoboBench items that cover the evaluated dimensions and planning subcategories. Annotators familiar with robotic manipulation receive the same visual context and task instructions as the models, together with the candidate answer or plan to be judged. For multiple-choice and affordance questions, they select or verify the correct answer under the original question constraints. For planning questions, they judge whether the predicted skill--object sequence is executable and whether it can complete the task under the scene constraints and the human-annotated dependency DAG.

\noindent\textbf{Annotator pool and per-dimension workflow.} All annotators hold a background in robotics or embodied AI and are trained on a shared rubric before labeling. Each dimension follows a two-stage label-then-validate flow: an initial answer or plan judgment is produced (model-assisted where ground truth is drafted by an MLLM), after which every annotation receives at least one independent human cross-pass that re-checks it against the visual context and the dimension rubric. Disagreements are tracked per dimension and adjudicated before reporting; on the held-out validation subset the residual disagreement after cross-pass is below 4\%.

\noindent\textbf{Difficulty calibration and item filtering.} To keep the benchmark discriminative, items are filtered by a majority-vote signal over the evaluated models: questions that all models answer correctly are removed as non-informative, and questions that all models answer incorrectly are routed to manual review to confirm they are solvable rather than mislabeled. This calibration removes both trivially easy and unsolvable items while preserving items with genuine spread.

\noindent\textbf{Implicit-instruction naturalness audit.} For the implicit instruction-comprehension subset, we additionally audit a 50-sample subset for naturalness (whether the implicit instruction reads as a plausible user request) and inferability (whether the intended goal is recoverable from the scene). A text-only judge rates the subset at $4.76\pm0.65$ for naturalness and $4.92\pm0.27$ for inferability, while a vision-grounded judge rates it at $4.80\pm0.49$ and $5.00\pm0.00$ on a 5-point scale, confirming that the implicit items are well-posed.

These human references provide an upper bound for each table and serve as the reference for the evaluator-alignment study in the main paper.

\subsection{Cross-Evaluator Robustness of the World-Simulator}
\label{subsec:supp_icc_robustness}

A central concern for the MLLM-as-world-simulator planning metric is whether its scores depend on the specific judge model or scoring prompt. We therefore stress-test the evaluator along two axes. First, we re-score the same 200 stratified Q1 samples with four heterogeneous judge models spanning three vendors (including \texttt{gpt-4o-2024-11-20}) and measure inter-judge agreement. Second, we vary the scoring prompt across the three variants described in Section~\ref{subsec:eval_variants} and detailed in the evaluation-prompt appendix: \textbf{v1} (basic judge, no rollout trace), \textbf{v2} (structural scoring rules, no rollout), and \textbf{v3} (the production DAG-grounded rollout prompt used in RoboBench). For each variant we report the inter-judge ICC(2,$k$) and Cronbach's $\alpha$ at both the per-node level (skill--object--parameter matches) and the composite (Comp) plan-completion level.

\begin{table}[t]
\centering
\caption{Cross-evaluator robustness of the world-simulator on 200 stratified Q1 samples scored by four judge models from three vendors. We report inter-judge ICC(2,$k$) and Cronbach's $\alpha$ at the per-node and composite (Comp) levels for three scoring-prompt variants. The DAG-grounded rollout prompt (v3) used in RoboBench attains the strongest consistency.}
\label{tab:supp_icc_robustness}
\setlength{\tabcolsep}{6pt}
\footnotesize
\begin{tabular}{lcccc}
\toprule
\multirow{2}{*}{\textbf{Scoring prompt variant}} & \multicolumn{2}{c}{\textbf{ICC(2,$k$)}} & \multicolumn{2}{c}{\textbf{Cronbach's $\alpha$}} \\
\cmidrule(lr){2-3} \cmidrule(lr){4-5}
 & \textbf{Node} & \textbf{Comp} & \textbf{Node} & \textbf{Comp} \\
\midrule
v1: base judge, no rollout      & 0.935 & 0.859 & 0.764 & 0.567 \\
v2: scoring rules, no rollout   & 0.939 & \textbf{0.900} & 0.780 & \textbf{0.687} \\
\textbf{v3: DAG rollout (paper)} & \textbf{0.948} & 0.854 & \textbf{0.790} & 0.585 \\
\bottomrule
\end{tabular}
\end{table}

As shown in Table~\ref{tab:supp_icc_robustness}, the production evaluator reaches a cross-judge node-level ICC(2,$k$) of $0.948$ and $\alpha=0.790$, indicating that node scores are nearly invariant to the choice of judge model. Robustness to the scoring prompt is equally strong: the panel-mean node score differs by only $0.22$ points on the $0$--$10$ scale across the three variants, and node-level ICC stays within $1.3$ percentage points ($0.935$--$0.948$). The DAG-grounded rollout prompt (v3) attains the best node-level ICC and $\alpha$—the node score is the quantity we report in the main results—while all three variants keep composite-level agreement high (ICC $\geq 0.85$). Together these results show the world-simulator metric is robust to both evaluator-model and prompt perturbations.

\subsection{Implicit-Instruction Chain-of-Thought Ablation}
\label{subsec:supp_implicit_cot}

Instruction comprehension is evaluated under two settings: an explicit setting, where the target object and goal are stated, and an implicit setting, where the user's intent must be inferred from context. To probe whether the implicit-setting gap can be closed by explicitly surfacing latent intent at inference time, we run a paired diagnostic ablation on 100 episodes from the implicit split: 61 plan-type items (36 full-plan, 25 next-step), scored by the world-simulator judge, and 39 step-verification (yes/no) items, scored by answer accuracy. The chain-of-thought (CoT) condition augments the implicit prompt with a scaffold that asks the model to first infer the latent intent, then identify the scene objects relevant to that intent, and finally infer the desired goal state, before emitting its final answer in the format the question requires. Each episode is evaluated under both conditions with identical images and ground truth. Table~\ref{tab:supp_implicit_cot} reports the paired comparison for three frontier models (GPT-5, Claude-Opus-4.7, Gemini-3.1-Pro).

\begin{table}[t]
\centering
\caption{Implicit-instruction chain-of-thought ablation on a paired 100-episode diagnostic subset of the implicit instruction-comprehension split (absolute values are not directly comparable to the full-split numbers in the main results). Plan-type items ($n{=}61$) are scored by the world-simulator composite ($0$--$100$); step-verification items ($n{=}39$) by yes/no accuracy (\%). ``Imp.'' is the implicit baseline; ``+CoT'' adds an intent$\rightarrow$objects$\rightarrow$goal reasoning scaffold to the same prompt. $\Delta$ is the paired change from adding the scaffold.}
\label{tab:supp_implicit_cot}
\setlength{\tabcolsep}{5pt}
\footnotesize
\begin{tabular}{lcccccc}
\toprule
\multirow{2}{*}{\textbf{Model}} & \multicolumn{3}{c}{\textbf{Planning ($n{=}61$)}} & \multicolumn{3}{c}{\textbf{Verification ($n{=}39$)}} \\
\cmidrule(lr){2-4} \cmidrule(lr){5-7}
 & \textbf{Imp.} & \textbf{+CoT} & \textbf{$\Delta$} & \textbf{Imp.} & \textbf{+CoT} & \textbf{$\Delta$} \\
\midrule
GPT-5  & 43.6 & 45.2 & $+1.6$ & 53.8 & 59.0 & $+5.1$ \\
Claude & 34.2 & 34.2 & $0.0$  & 64.1 & 59.0 & $-5.1$ \\
Gemini & 39.5 & 36.4 & $-3.1$ & 56.4 & 61.5 & $+5.1$ \\
\bottomrule
\end{tabular}
\end{table}

The scaffold yields no consistent gain. On plan-type items the paired deltas are small ($+1.6$, $0.0$, $-3.1$) and the vast majority of episodes are exact ties (41, 44, and 39 of 61, respectively); on step-verification items the deltas are likewise small and mixed in sign. We also find the outcome is highly sensitive to the scaffold's output-format instruction: an earlier variant that constrained the final line to a single action systematically truncated full-plan answers and corrupted yes/no answers, producing large spurious deltas in both directions. With a format-neutral scaffold, the implicit-setting deficit persists essentially unchanged, indicating that for frontier models it reflects a genuine intent-grounding limitation rather than a shortfall that inference-time prompting can repair—motivating training-time interventions instead. We note this ablation covers only frontier models; weaker MLLMs, whose implicit-setting gap is larger in the main results, may still benefit from explicit intent scaffolding, which we leave to future study. The scaffold is used only for this diagnostic analysis; the benchmark's default instruction-comprehension setting evaluates direct model behavior without it.

\subsection{Robustness to Ground-Truth Annotation Noise}
\label{subsec:supp_gt_noise}

Because the task-completion metric is scored against human-annotated reference plans and dependency DAGs, we test how sensitive model scores are to imperfect ground truth. On a 30-item Q1 subset, we inject two kinds of controlled perturbation into the reference annotations: \textbf{deletion}, which drops a fraction of ground-truth nodes, and \textbf{modification}, which corrupts a fraction of nodes, each at $5\%$ and $10\%$ rates. We then re-score three frontier models under each perturbed reference and compare against the clean ($0\%$) composite score.

\begin{table}[t]
\centering
\caption{Robustness of the composite ($0$--$100$) task-completion score to ground-truth annotation noise on a 30-item Q1 subset. ``Del'' deletes and ``Mod'' modifies the indicated fraction of ground-truth nodes. Under modification—the realistic corruption—scores shift only modestly and the ranking is preserved; deletion removes required milestones and mechanically inflates completion, yet the strongest model (Gemini) stays best throughout.}
\label{tab:supp_gt_noise}
\setlength{\tabcolsep}{6pt}
\footnotesize
\begin{tabular}{lccccc}
\toprule
\multirow{2}{*}{\textbf{Model}} & \textbf{Clean} & \multicolumn{2}{c}{\textbf{Deletion}} & \multicolumn{2}{c}{\textbf{Modification}} \\
\cmidrule(lr){3-4} \cmidrule(lr){5-6}
 & \textbf{0\%} & \textbf{5\%} & \textbf{10\%} & \textbf{5\%} & \textbf{10\%} \\
\midrule
GPT-5  & 66.6 & 75.0 & 74.8 & 61.7 & 60.0 \\
Claude & 70.3 & 65.5 & 71.0 & 68.7 & 61.7 \\
Gemini & 79.7 & 83.5 & 80.3 & 75.0 & 78.6 \\
\bottomrule
\end{tabular}
\end{table}

Table~\ref{tab:supp_gt_noise} separates the two perturbation types. Under \emph{modification}—which keeps the plan length but corrupts node content, the more realistic form of annotation error—composite scores drop by at most a few points even at the $10\%$ rate (well above the residual annotation-disagreement rate of under $4\%$ measured in Section~\ref{subsec:human_eval_protocol}), and the ordering Gemini~$>$~Claude~$>$~GPT-5 is preserved. \emph{Deletion} removes required milestones, which mechanically makes plans easier to ``complete'' and therefore inflates the score; under this artifact the two closely matched mid-tier models (GPT-5 and Claude, within $\sim\!4$ points when clean) can swap, but Gemini remains clearly best. Overall, the metric's conclusions are stable under the corruption that models real annotation disagreement, and the top of the ranking is unaffected by either perturbation.

\subsection{Embodied Cognitive Ability and Downstream VLA Performance}

In this section, we investigate whether the embodied cognitive abilities of VLM backbones are predictive of their downstream Vision-Language-Action (VLA) performance. To this end, we combine the embodied cognitive scores of 8 VLMs with their downstream fine-tuning results under the VLM4VLA protocol~\cite{zhang2026vlm4vla}, and perform correlation analysis on two representative manipulation benchmarks, \textit{Calvin}~\cite{mees2022calvin} and \textit{Libero}~\cite{liu2023libero}. Rather than focusing only on top-level dimensions, we analyze grouped dimensions at the next level of granularity, including robotic-centric, object-centric, and scene-centric perception reasoning, together with affordance- and failure-related groups. Our results reveal a clear benchmark-dependent pattern: downstream performance on \textit{Calvin} is more strongly associated with perception reasoning, whereas performance on \textit{Libero} exhibits its strongest dependence on affordance-related capability.

\subsubsection{Experimental Setup}

We follow the downstream fine-tuning protocol of VLM4VLA for all evaluations. Specifically, each VLM is adapted into a VLA policy by appending a learnable \texttt{ActionQuery} token to the multimodal input sequence, and decoding its final hidden state into an action chunk through a lightweight MLP policy head. All model parameters are fine-tuned during downstream training, including the vision encoder, token embeddings, and the language model. Following VLM4VLA, we only use a single-view RGB observation from the current frame and do not provide proprioceptive state input, so that the comparison isolates how intrinsic visual-language capability transfers to robotic control.

For downstream evaluation, we adopt the same settings as VLM4VLA on \textit{Calvin} and \textit{Libero}. On \textit{Calvin}, we use the ABC-D split, where models are trained on scenes A/B/C and evaluated on the unseen scene D. Policies are trained for 30k steps and evaluated on 1000 task sequences, each containing 5 sequential subtasks; the reported metric is the average number of successfully completed tasks per sequence. On \textit{Libero}, we evaluate on the most challenging suite, \textit{Libero-Long} (Libero-10), which consists of 10 long-horizon tabletop manipulation tasks involving diverse objects, scenes, and manipulation types. All models are trained for 50k steps and evaluated with 50 trials under random initializations for each task. We also keep the same benchmark-specific hyperparameters as VLM4VLA for fairness: batch size 128, learning rate $2\times10^{-5}$, and action chunk size 10 for \textit{Calvin}; batch size 512, learning rate $5\times10^{-5}$, and action chunk size 4 for \textit{Libero}.

\begin{table*}[t]
\centering
\caption{Grouped embodied cognitive scores of the 8 evaluated VLMs and their downstream VLA performance on \textit{Calvin} and \textit{Libero}. 
\textbf{Robotic-centric} is averaged over \textit{robot\_type} and \textit{robot\_view}; 
\textbf{Object-centric} is averaged over \textit{static\_attribute} and \textit{tool\_usage}; 
\textbf{Scene-centric} is averaged over \textit{spatial\_relation}, \textit{timestamp\_analysis}, and \textit{spatial\_temporal\_causality}; 
\textbf{Static + Dynamic Affordance} is averaged over \textit{static\_affordance} and \textit{dynamic\_affordance}.}
\label{tab:grouped_brain_vla_scores}
\resizebox{\textwidth}{!}{%
\begin{tabular}{lcccccccc}
\toprule
\textbf{Model} & \textbf{Robotic-centric} & \textbf{Object-centric} & \textbf{Scene-centric} & \textbf{Static+Dynamic} & \textbf{Navigation} & \textbf{Execution Error} & \textbf{Calvin} & \textbf{Libero} \\
\midrule
Paligemma-1      & 16.37 & 17.19 & 22.28 & 22.64 & 22.45 & 13.91 & 3.51 & 44.2 \\
Paligemma-2      & 19.42 & 18.10 & 19.02 & 27.04 & 21.43 & 19.87 & 3.41 & 46.2 \\
Qwen2.5VL-3B     & 19.18 & 45.84 & 30.01 & 17.27 & 25.51 & 11.92 & 3.86 & 43.0 \\
Qwen2.5VL-7B     & 29.15 & 42.49 & 26.73 & 26.32 & 26.53 & 7.95  & 4.06 & 45.0 \\
Qwen3VL-2B       & 23.20 & 41.10 & 29.01 & 28.38 & 12.24 & 9.27  & 4.14 & 55.8 \\
Qwen3VL-4B       & 38.98 & 49.94 & 30.28 & 20.31 & 39.80 & 23.18 & 3.94 & 44.4 \\
Qwen3VL-8B       & 44.14 & 51.18 & 35.23 & 21.29 & 32.65 & 22.52 & 4.04 & 46.2 \\
Qwen3VL-30B-A3B  & 41.94 & 52.08 & 32.17 & 25.34 & 22.45 & 25.83 & 4.08 & 46.8 \\
\bottomrule
\end{tabular}%
}
\end{table*}

\begin{table}[t]
\centering
\caption{Pearson correlation between grouped embodied cognitive dimensions and downstream VLA performance.}
\label{tab:grouped_brain_vla_corr}
\setlength{\tabcolsep}{4pt}
\footnotesize
\begin{tabular}{lcc}
\toprule
\textbf{Dimension} & \textbf{Calvin} & \textbf{Libero} \\
& \textbf{($r$, $p$)} & \textbf{($r$, $p$)} \\
\midrule
Robotic-centric      & 0.625, 0.0972  & -0.068, 0.8732 \\
Object-centric       & 0.884, 0.0035  & 0.011, 0.9801 \\
Scene-centric        & 0.833, 0.0103  & 0.064, 0.8803 \\
Static + Dynamic Aff. & 0.053, 0.9000 & 0.677, 0.0652 \\
Navigation Aff.      & 0.075, 0.8593  & -0.680, 0.0637 \\
Low-level Exec. Err. & -0.057, 0.8940 & -0.259, 0.5353 \\
\bottomrule
\end{tabular}
\end{table}

\begin{figure*}[h]
    \centering
    \includegraphics[width=\textwidth]{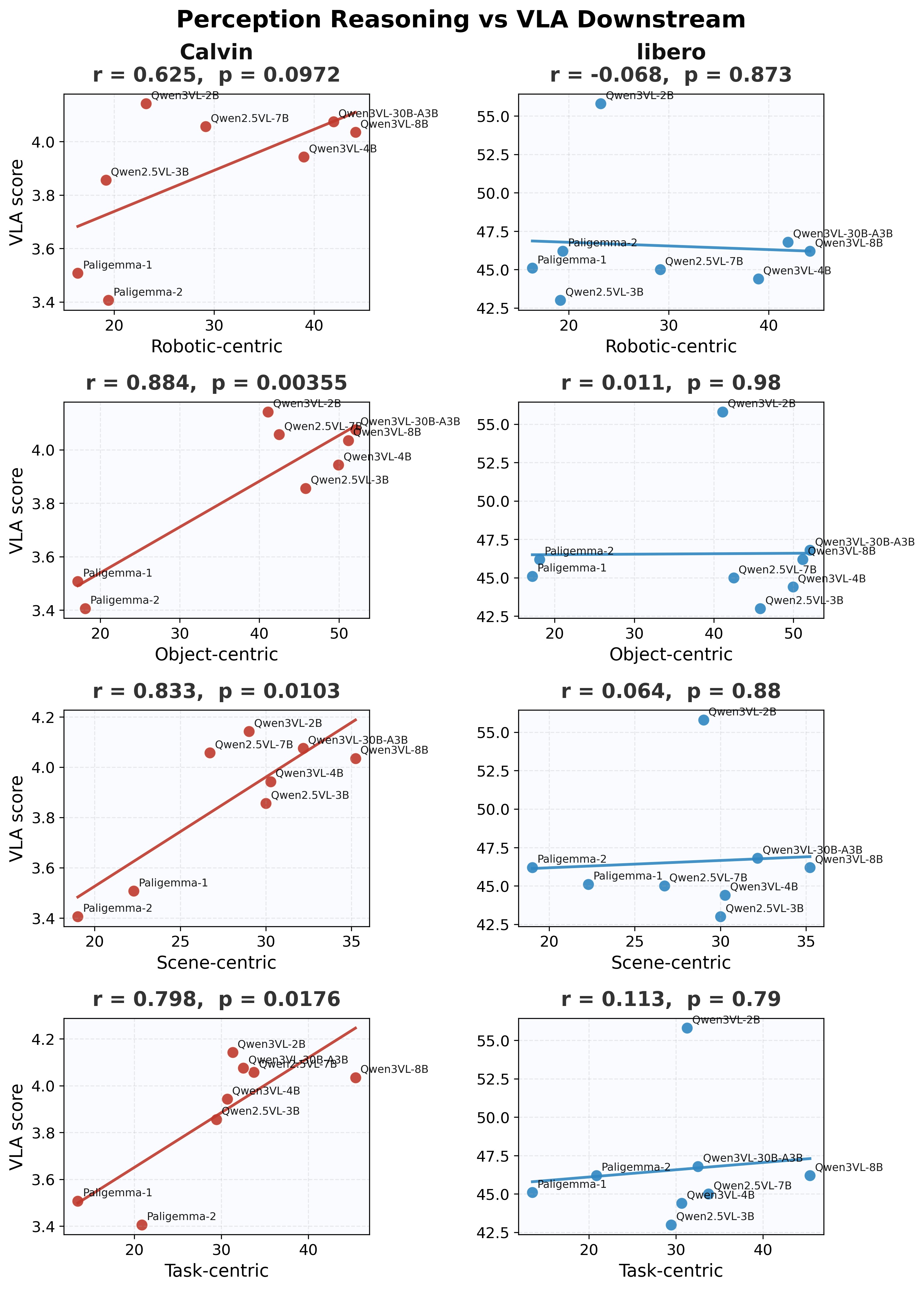}
    \caption{Correlation between grouped perception-reasoning dimensions and downstream VLA performance on \textit{Calvin} and \textit{Libero}. Each point corresponds to one VLM backbone, and the fitted line indicates the linear trend. In the main discussion, we focus on \textbf{Robotic-centric}, \textbf{Object-centric}, and \textbf{Scene-centric} perception reasoning, which show the clearest benchmark-dependent differences.}
    \label{fig:perception_group_corr}
\end{figure*}

\begin{figure*}[h]
    \centering
    \includegraphics[width=\textwidth]{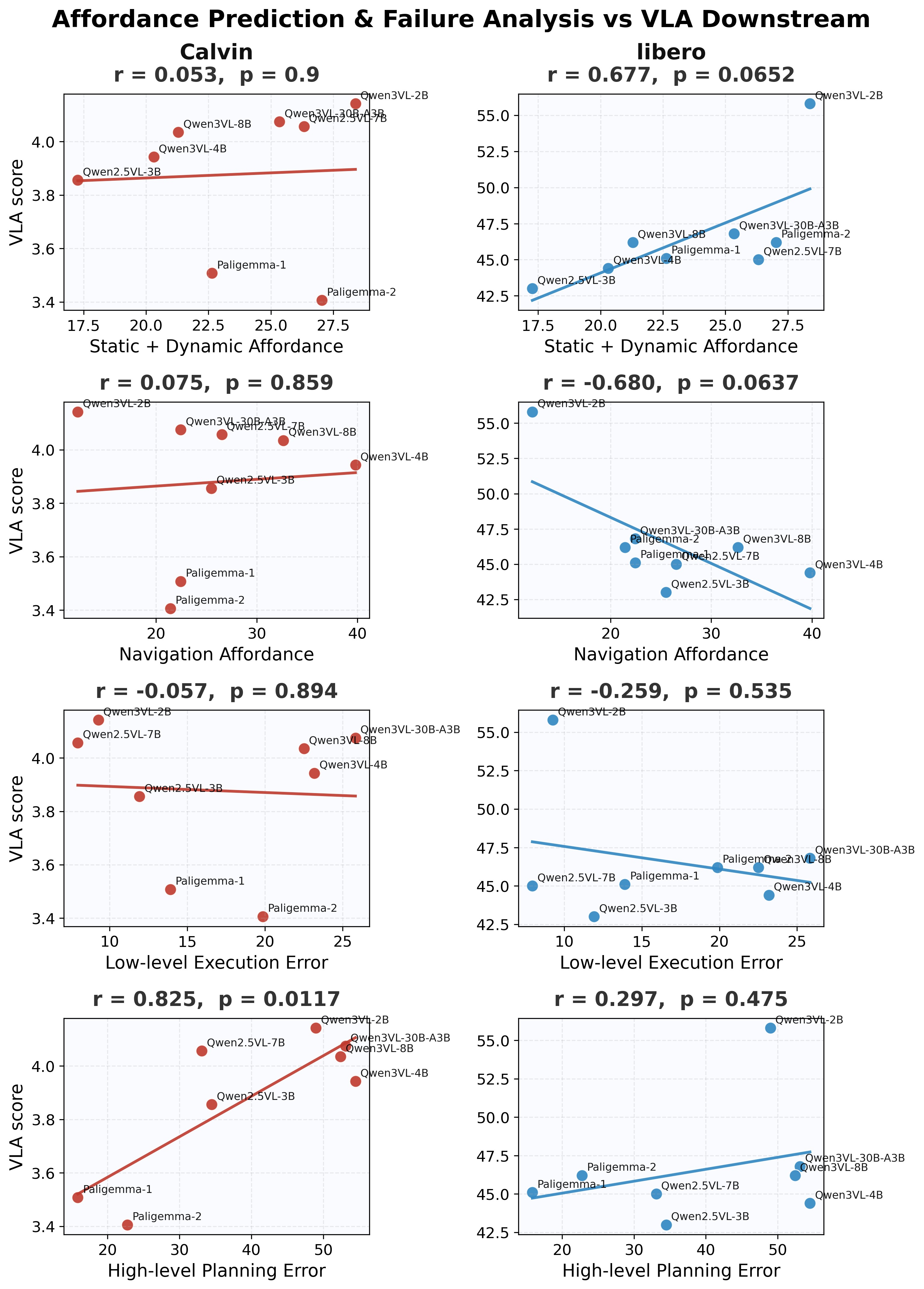}
    \caption{Correlation between grouped affordance/failure-related dimensions and downstream VLA performance on \textit{Calvin} and \textit{Libero}. We primarily discuss \textbf{Static + Dynamic Affordance}, \textbf{Navigation Affordance}, and \textbf{Low-level Execution Error}. Among these dimensions, \textbf{Static + Dynamic Affordance} shows the strongest positive trend with downstream performance on \textit{Libero}.}
    \label{fig:affordance_group_corr}
\end{figure*}

\subsubsection{Results and Discussion}

The grouped embodied cognitive scores of the evaluated VLMs and their downstream VLA performance are reported in Table~\ref{tab:grouped_brain_vla_scores}. The Pearson correlation results are summarized in Table~\ref{tab:grouped_brain_vla_corr}, and the corresponding visualizations are shown in Fig.~\ref{fig:perception_group_corr} and Fig.~\ref{fig:affordance_group_corr}. Overall, the results suggest that embodied cognitive ability provides informative signals for predicting downstream VLA performance, while the most predictive dimensions are clearly benchmark-dependent. Rather than revealing a single universal cognitive predictor, the analysis shows that different downstream environments emphasize different aspects of embodied intelligence.

\paragraph{Perception reasoning is more predictive on \textit{Calvin}.}
On \textit{Calvin}, perception-related grouped dimensions exhibit clear positive correlations with downstream performance. In particular, \textbf{Object-centric} perception shows the strongest association with \textit{Calvin} performance ($r=0.884$, $p=0.0035$), followed by \textbf{Scene-centric} perception ($r=0.833$, $p=0.0103$). \textbf{Robotic-centric} perception also shows a positive trend ($r=0.625$, $p=0.0972$), although it does not reach conventional significance. Taken together, these results indicate that \textit{Calvin} is especially sensitive to the perception-reasoning ability of the backbone VLM.

A plausible explanation lies in the structure of the benchmark itself. \textit{Calvin} evaluates policies on sequential 5-step long-horizon tasks and, under the ABC-D protocol, explicitly requires generalization from seen scenes to an unseen scene. This evaluation places substantial pressure on robust perceptual grounding across sequential decision-making: the policy must correctly identify task-relevant objects, understand scene layout, and maintain consistent interpretation of the evolving visual context over multiple subgoals. Under such conditions, errors in object recognition or scene understanding can accumulate across steps and compromise the entire execution chain. This is consistent with the strong predictive value of object- and scene-centric perception reasoning on \textit{Calvin}.

More broadly, these results suggest that \textit{Calvin} depends less on isolated local manipulation cues and more on holistic visual reasoning over long-horizon interaction. In other words, success on \textit{Calvin} is not solely determined by whether a model understands how an object can be manipulated, but also by whether it can stably ground \emph{what} the relevant entities are, \emph{where} they are situated, and \emph{how} the scene evolves as the task unfolds.

\paragraph{Affordance prediction is more predictive on \textit{Libero}.}
In contrast, \textit{Libero} exhibits its strongest positive trend with affordance-related ability. Specifically, \textbf{Static + Dynamic Affordance} shows the largest positive correlation with downstream \textit{Libero} performance ($r=0.677$, $p=0.0652$). Although this result does not pass the conventional $p<0.05$ threshold, the trend is substantial and notably stronger than the correlations observed for the perception-reasoning groups on the same benchmark. At the same time, \textbf{Navigation Affordance} is negatively correlated with \textit{Libero} performance ($r=-0.680$, $p=0.0637$), indicating that not all affordance-related sub-skills transfer equally well across benchmarks.

We conjecture that this pattern is closely related to the structure of \textit{Libero-Long}. While \textit{Libero} also contains long-horizon tasks, its evaluation is centered on fixed-tabletop, multi-category manipulation scenarios involving diverse objects and fine-grained interactions. Relative to \textit{Calvin}, the challenge in \textit{Libero} appears to lie less in scene-level generalization and more in whether the policy can correctly infer manipulation-relevant object properties, contact opportunities, and action constraints. Put differently, \textit{Libero} places stronger emphasis on understanding \emph{how} an object can be acted upon, which naturally makes affordance prediction more relevant to downstream success.

This benchmark-specific pattern is intuitively consistent with the distinction between the two downstream environments. \textit{Calvin} stresses perceptual robustness over long-horizon scene changes, whereas \textit{Libero} more directly tests fine-grained manipulation competence on diverse tabletop objects. Our correlation results align well with this distinction.

\paragraph{Low-level execution error is less predictive under the current setting.}
Compared with perception reasoning and affordance prediction, \textbf{Low-level Execution Error} shows weak correlations with both \textit{Calvin} and \textit{Libero}. This suggests that, under the current VLM4VLA setting, differences in downstream policy performance are not primarily explained by low-level execution-error awareness. Instead, the dominant predictive signals appear to come from higher-level embodied cognitive capabilities, especially benchmark-relevant perception reasoning in \textit{Calvin} and affordance understanding in \textit{Libero}.

\paragraph{Takeaway.}
Overall, the results support the hypothesis that embodied cognitive ability is closely related to downstream VLA performance, but the relevant ability dimensions are environment-dependent. \textit{Calvin} mainly rewards strong perception reasoning, especially object- and scene-centric understanding, whereas \textit{Libero} is more sensitive to affordance-related capability for fine-grained manipulation. This benchmark-specific structure suggests that evaluating VLM backbones through embodied cognitive dimensions can offer more diagnostic signals for downstream VLA transfer than relying on generic VLM capability alone. 
\clearpage

\subsection{Error Definitions and Additional Analysis}
\subsubsection{Error Taxonomy} 
In this section, we define the types of errors and sub-errors encountered in our benchmark. 
We categorize errors into four main types: \textit{Execution Errors}, \textit{Identification Errors}, \textit{Common Sense Errors}, and \textit{Mode-Specific Errors}. 
Each error type corresponds to a distinct failure mode of the embodied brain. 
For instance, execution errors arise when generating or executing action sequences, identification errors occur when associating objects or parameters, common sense errors reflect violations of physical or spatial priors, and mode-specific errors are tied to task-specific formatting requirements. 
A detailed taxonomy of error types and definitions is provided in Table~\ref{tab:error_taxonomy}.

\begin{table}[h!]
\footnotesize
\centering
\resizebox{\columnwidth}{!}{%
\begin{tabular}{p{3cm} p{3.5cm} p{6cm}}
\toprule
\textbf{Error Type} & \textbf{Sub-error} & \textbf{Definition} \\
\midrule
\multirow{4}{*}{\makecell{Execution\\Errors}} 
 & Missing Steps & The predicted action list omits necessary functions for task completion. \\
 & Impossible Actions & The model outputs action functions that are not defined in the action space. \\
 & Redundant Steps & The model outputs an excessive number of action functions beyond requirements. \\
 & Wrong Function & The model selects an incorrect action function for the intended operation. \\
\midrule
\multirow{3}{*}{\makecell{Identification\\Errors}} 
 & Aliasing Errors & Failing to distinguish between visually similar objects (e.g., recognizing a crumpled paper ball as popcorn). \\
 & Parameter Mismatch & Producing incorrect parameter values for otherwise valid functions. \\
 & Wrong Object & Referring to objects completely unrelated to the intended target. \\
\midrule
\multirow{2}{*}{\makecell{Common Sense\\Errors}} 
 & Physics Violations & Generating physically impossible action sequences (e.g., folding multiple clothes simultaneously). \\
 & Spatial Reasoning Errors & Failing to infer correct spatial operations (e.g., twisting a faucet in the wrong direction). \\
\midrule
\multirow{2}{*}{\makecell{Mode-Specific\\Errors}}
 & CSS Mode Reference Errors & Referring to objects directly with natural language descriptions instead of the required object indices. \\
 & Dual Arm Assignment & In dual arm tasks, the robot fails to use the arm that is closer to the target object, resulting in an unreasonable operation. \\
\bottomrule
\end{tabular}%
}
\caption{Error taxonomy with definitions in RoboBench.}
\label{tab:error_taxonomy}
\end{table}

\subsubsection{Error Analysis}

\textbf{Execution Errors.}  
This category reflects failures in generating valid action sequences. We identify four common subtypes:  
(1) \textit{Missing Steps}: the predicted sequence omits essential actions, such as failing to include a “pick up object” step, which prevents successful task completion.  
(2) \textit{Impossible Actions}: the model produces functions not defined in the available action space, e.g., outputting an undefined “teleport” action.  
(3) \textit{Redundant Steps}: the model generates more actions than required, often inserting unnecessary “move” operations that deviate from the ground truth sequence.  
(4) \textit{Wrong Function}: the model selects the incorrect action for the intended goal, such as using “push” instead of “pull.”  

\textbf{Identification Errors.}  
These errors occur when the model fails to correctly associate actions with the appropriate objects or parameters.  
(1) \textit{Aliasing Errors}: the model confuses visually similar objects, such as misidentifying a crumpled paper ball as popcorn.  
(2) \textit{Parameter Mismatch}: although the correct function is chosen, its parameters are incorrect, e.g., attempting to grasp an object with an inappropriate force setting.  
(3) \textit{Wrong Object}: the output involves manipulating an irrelevant object, such as operating on a cup when the instruction specifies a book.  

\textbf{Common Sense Errors.}  
These errors reveal violations of physical feasibility or basic spatial reasoning.  
(1) \textit{Physics Violations}: the predicted plan contains physically impossible steps, such as folding multiple pieces of clothing simultaneously.  
(2) \textit{Spatial Reasoning Errors}: the model fails to infer correct spatial or directional relations, e.g., twisting a faucet in the wrong direction.  

\textbf{Mode-Specific Errors.}  
Certain failures are tied to formatting or mode-specific requirements.  
(1) \textit{CSS Mode Reference Errors}: instead of adhering to symbolic references, the model outputs natural language descriptions of objects, such as predicting “the red cup” rather than the required identifier “Object\_3.”  

Overall, these errors highlight distinct limitations across execution, perception, reasoning, and adherence to task-specific conventions. Addressing them requires better grounding of action functions, stronger disambiguation of objects, improved spatial reasoning, and stricter alignment with mode constraints.

\section{More Statistics of Planning Tasks}
\label{sec:statistics}

This section provides additional statistics for planning tasks in \textbf{RoboBench}, covering detailed distributions of skills, actions, objects, instructions, and task dimensions, as well as dataset source distributions.

\subsection{Annotation Quality Control}

RoboBench uses dimension-specific validation to control the quality of MLLM-assisted and human-authored annotations. Each item receives automated format checks and at least one human cross-pass. Items with ambiguous language, visually unsupported answers, inconsistent state transitions, or invalid action functions are either corrected or removed. Table~\ref{tab:annotation_qc} summarizes the main validation focus by dimension. On the validation subset, residual disagreement after adjudication is below 4\%, and the disagreement log is retained for release-time auditing.

\begin{table}[h]
\centering
\small
\caption{Annotation quality-control checks by dimension.}
\label{tab:annotation_qc}
\resizebox{\linewidth}{!}{%
\begin{tabular}{p{3.1cm} p{5.2cm} p{4.8cm}}
\toprule
\textbf{Dimension} & \textbf{Primary validation focus} & \textbf{Adjudication trigger} \\
\midrule
Instruction comprehension & Naturalness and inferability of implicit demands; consistency with the scene and explicit goal & Ambiguous demand, missing scene grounding, or multiple plausible target goals \\
Perception reasoning & Correct object, robot, scene, temporal, and task-centric labels & Visual mismatch, distractor ambiguity, or unsupported causal relation \\
Generalized planning & Skill--object--parameter validity and DAG dependency consistency & Invalid function, missing prerequisite, wrong order, or infeasible state transition \\
Affordance prediction & Contact point, trajectory, and mobile-base feasibility under the visible scene & Unreachable point, inconsistent trajectory, or task-irrelevant affordance \\
Failure analysis & Separation of real execution failures from synthetic high-level perturbations & Non-identifiable failure, wrong failure class, or correction not grounded in the scene \\
\bottomrule
\end{tabular}%
}
\end{table}

\subsection{Detailed Task Statistics}

\begin{figure*}[t]
    \centering
    \includegraphics[width=0.9\textwidth]{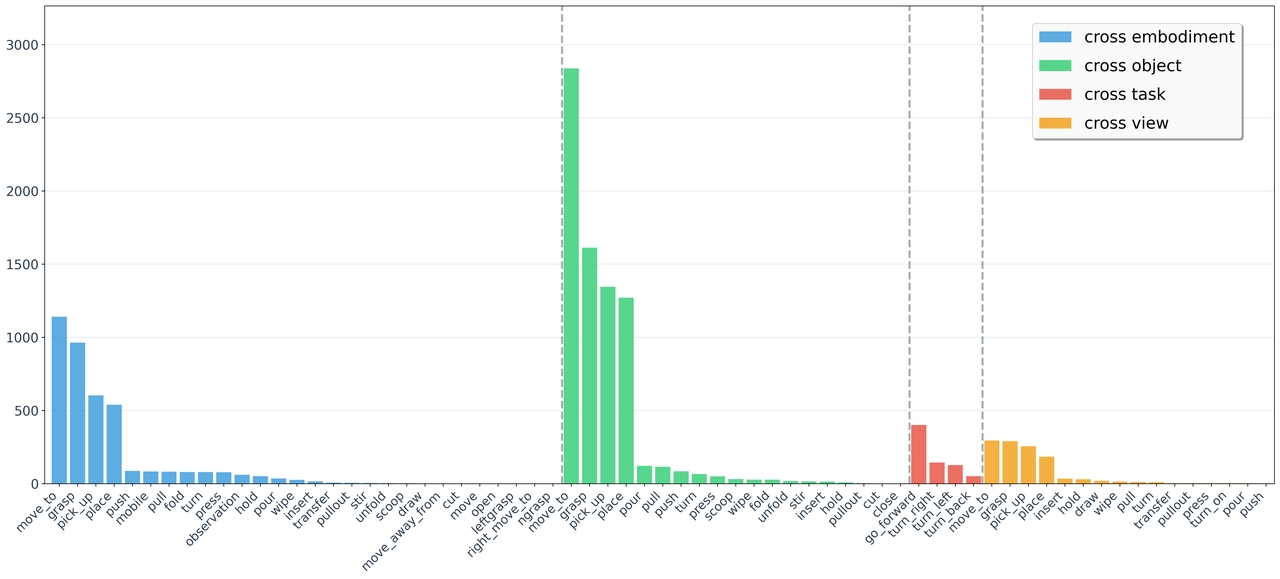}
    \caption{Skill distributions in RoboBench planning tasks. unique skill counts per data.}
    \label{fig:skill_statistics}
\end{figure*}

\paragraph{Skill Statistics}  
Figure~\ref{fig:skill_statistics} presents the distribution of skills. The horizontal axis shows skill names, and the vertical axis indicates counts. %

\begin{figure*}[t]
    \centering
    \includegraphics[width=0.9\textwidth]{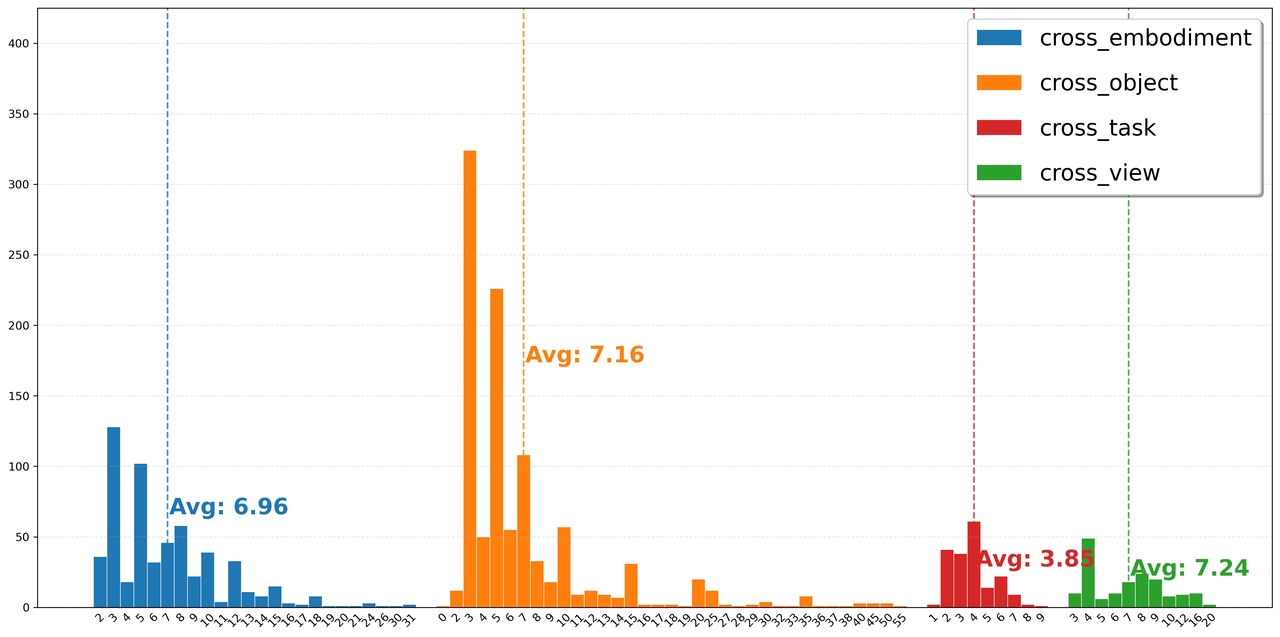}
    \caption{Distribution of action sequence lengths in RoboBench planning tasks.}
    \label{fig:action_statistics_length}
\end{figure*}

\begin{figure*}[h]
    \centering
    \includegraphics[width=\textwidth]{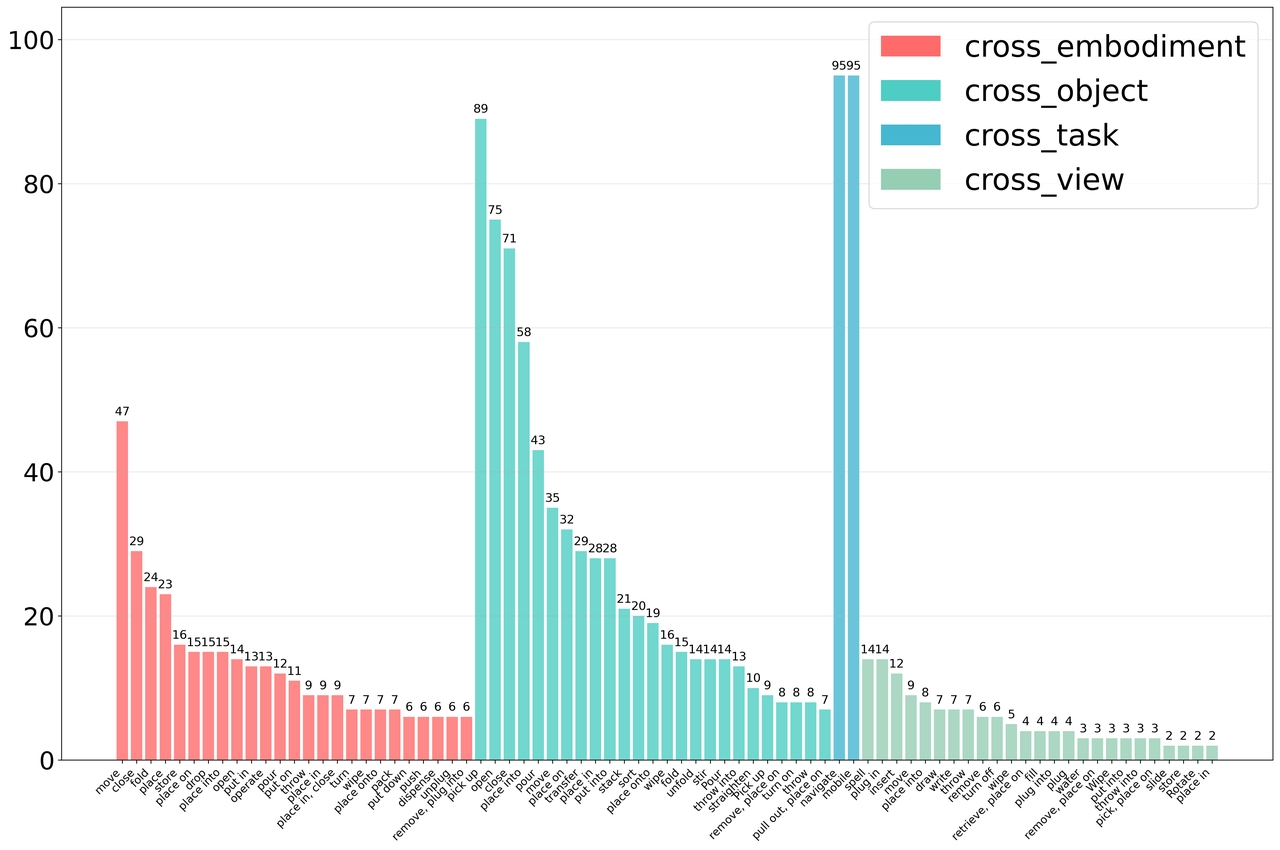}
    \caption{Frequency of each action name in RoboBench planning tasks.}
    \label{fig:action_statistics_name}
\end{figure*}

\paragraph{Action Statistics}  
Figure~\ref{fig:action_statistics_length} shows the distribution of action sequence lengths across tasks, while Figure~\ref{fig:action_statistics_name} shows action name vs. frequency counts.

\subsection{Dataset Source Distribution}

\begin{figure*}[h]
    \centering
    \includegraphics[width=0.8\textwidth]{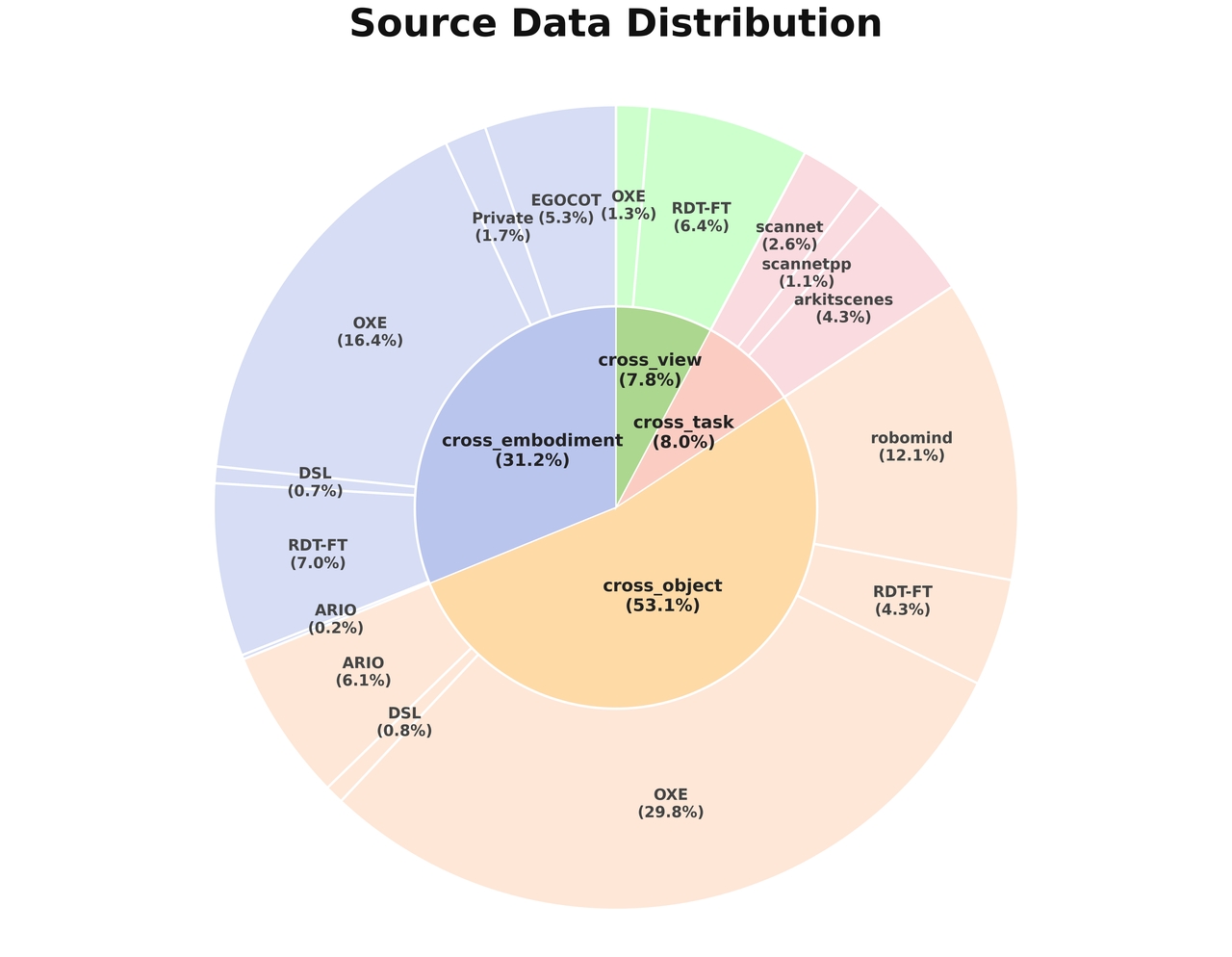}
    \caption{Source distribution of planning tasks in RoboBench dataset.}
    \label{fig:dataset_source}
\end{figure*}

Figure~\ref{fig:dataset_source} presents the distribution of source datasets for planning tasks in RoboBench. The figure shows the proportion of tasks contributed by each source.

\section{Case Study}
\label{sec:case}
\begin{figure}[h]
    \centering
    \includegraphics[width=\linewidth]{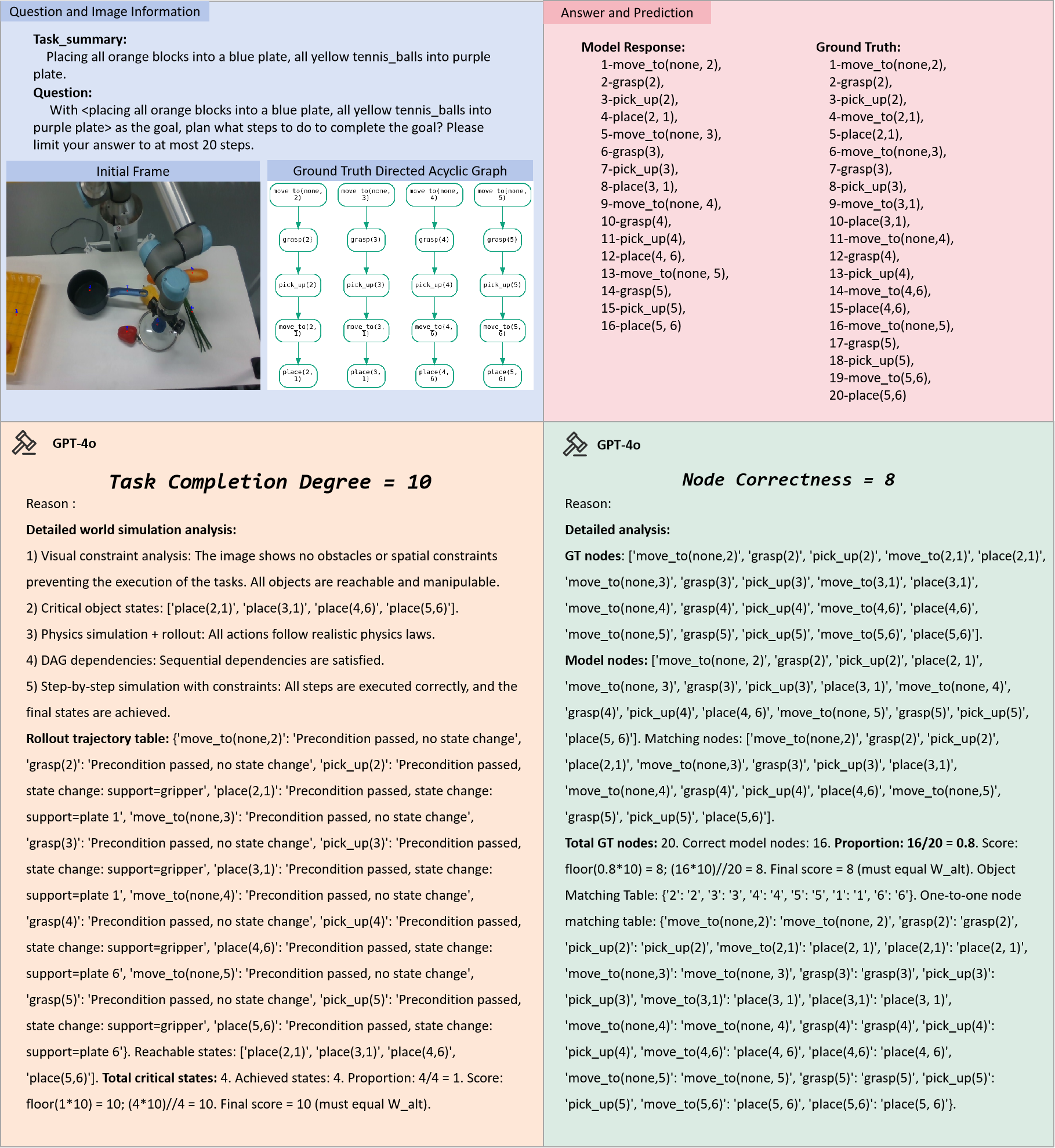}
    \caption{Successful evaluation examples illustrating robustness.}
    \label{fig:evaluation_success}
\end{figure}

To provide further insights, we present illustrative case studies of our evaluation. 
Figure~\ref{fig:evaluation_success} highlights successful examples that demonstrate the model's robustness.

\section{Prompts}
\label{sec:prompts}

This section presents the prompts employed for both \textbf{benchmark data construction} and \textbf{model evaluation} in RoboBench. We organize them by task type and processing stage, emphasizing reproducibility and precision.

\subsection{Benchmark Data Construction}

\subsubsection{Planning Tasks}

\paragraph{1. Video to Natural Language Task Description} 
Robot videos are converted into natural language task descriptions, including sequential step descriptions. Three prompts are used to cover different robot types: single-arm, dual-arm, and mobile manipulator robots, shown in Fig.~\ref{prompt:video2nl_single},~\ref{prompt:video2nl_dual} and~\ref{prompt:video2nl_mobile}.

\begin{figure*}
\begin{tcolorbox}[colback=gray!5!white, colframe=gray!75!black, 
title=Prompt: Video $\rightarrow$ NL Task \& Steps (Single-Arm), boxrule=0.3mm, width=\textwidth, arc=3mm, auto outer arc=true]
\tiny	
You will analyze a video (represented by image frames) of a robotic arm performing a specific task, where the task is described as: \texttt{\{desc\}}. Note that the referenced task summary might not be accurate or complete. Your task is to identify the primary task during the video with the help of the referenced description, summarize the task and rewrite the description, extract the necessary steps to complete it, and specify the frame range for each step. Follow these instructions:

\begin{enumerate}
    \item \textbf{Task Identification}: First, identify the main task the robotic arm is performing. This task could be a clear goal or a series of related activities (e.g., assembling furniture, repairing equipment, preparing food, etc.). Briefly describe the primary task in one sentence.

    \item \textbf{Step Extraction}: Once the task is identified, extract the key steps required to complete it, ensuring that each step is clearly described and logically ordered. Each step may include:
    \begin{itemize}
        \item Specific actions (e.g., tightening screws, stirring mixtures, pressing buttons, etc.)
        \item Frame window: Specify the start and end frame for each step (from \(0\) to \texttt{\{maxframeid\}}, since the video has \texttt{\{cnt\}} frames).
    \end{itemize}

    \item \textbf{Frame Range Constraints}:
    \begin{itemize}
        \item \textbf{No Overlapping Frames}: Ensure that the frame ranges of each step do not overlap with each other. Each frame should be assigned to exactly one step.
        \item \textbf{Full Frame Coverage}: Ensure that all \texttt{\{cnt\}} frames (from \(0\) to \texttt{\{maxframeid\}}) are included in the steps. No frames should be missed or duplicated.
        \item \textbf{Must start from frame 0}
    \end{itemize}

    \item \textbf{Notes}:
    \begin{itemize}
        \item Please annotate as finely as possible and try not to have more than ten frames of the same thing being done (unless it is difficult to distinguish).
        \item A step has only one verb (unless two or more actions are strictly performed simultaneously).
        \item Please ensure the accuracy of labeling and image matching.
        \item For example, if the gripper changes from an open state to a closed state in a sequence of frames, the action is called \texttt{pick}; On the contrary, if it changes from a closed state to an open state, the action is called \texttt{release}.
    \end{itemize}

    \item \textbf{Failure Identification}: If the robotic arm attempts an action but does not succeed, clearly indicate this in the step description. For example, if the robotic arm tries to pick up a block but fails, the step description should be something like \texttt{Attempt to pick up a block but fails}.

    \item \textbf{Output Format}: Provide the task description and steps in two parts, formatted as JSON:
    \begin{itemize}
        \item \textbf{Task Summary}: A string summarizing the primary task in the video without mentioning the subjects - the robotic arm.
        \item \textbf{Steps}: An array where each element represents a step, containing:
        \begin{itemize}
            \item \texttt{step\_description}: A concise description of the step with the action being performed in the format of verb phrases without mentioning the subjects - the robotic arm (e.g., "Add syrup in the glass").
            \item \texttt{start\_frame}: The start frame of the step (from \(0\) to \texttt{\{cnt-1\}}).
            \item \texttt{end\_frame}: The end frame of the step (from \(0\) to \texttt{\{cnt-1\}}).
        \end{itemize}
    \end{itemize}
\end{enumerate}

\textbf{Task Description}: \texttt{\{desc\}}
\newline
\newline
**Example Output Format 1: (This is an example with a total frame length of 30, and the specific situation depends on the actual frame length.)**
\begin{verbatim}
```json
{{
  "task_summary": "Assembling an office desk.",
  "steps": [
    {{
      "step_description": "Remove all components and screws from the package.",
      "start_frame": 0,
      "end_frame": 4
    }},
    {{
      "step_description": "Use a screwdriver to attach the legs to the tabletop.",
      "start_frame": 5,
      "end_frame": 14
    }},
    ...
    {{
      "step_description": "Ensure all screws are tight and the desk is stable.",
      "start_frame": 25,
      "end_frame": 29
    }}
  ]
}}
```
\end{verbatim}
**Example Output Format 2: 
...
\end{tcolorbox}
\caption{Prompt for converting single-arm robot videos into task and step descriptions in natural language.}
\label{prompt:video2nl_single}
\end{figure*}

\begin{figure*}[ht]
    \begin{tcolorbox}[colback=gray!5!white, colframe=gray!75!black,
    title=Prompt: Video $\rightarrow$ NL Task \& Steps (Dual-Arm), boxrule=0.3mm, width=\textwidth, arc=3mm, auto outer arc=true]
\tiny	
You will analyze a video (represented by image frames) of a dual-arm robotic system performing a specific task, where the task is described as: \texttt{\{desc\}}. Note that the referenced task summary might not be accurate or complete. Your task is to identify the primary task during the video with the help of the referenced description, summarize the task and rewrite the description, extract the necessary steps to complete it, and specify the frame range for each step. Follow these instructions:

\begin{enumerate}
    \item \textbf{Left-Right Hand Identification}: 
    The video is recorded from a first-person view of the dual-arm robotic system. Your first task is to accurately identify which hand is the left arm (\texttt{[left]}) and which hand is the right arm (\texttt{[right]}). This is crucial as you proceed with the task analysis. Use visual cues such as the relative position of the hands, their orientation, and any distinguishing features to determine which hand is on the left and which is on the right.
    
    \item \textbf{Task Identification}: 
    Once the left and right arms are correctly identified, determine the main task the robotic system is performing. This task could be a clear goal or a series of related activities (e.g., assembling furniture, repairing equipment, preparing food, etc.). Briefly describe the primary task in one sentence.
    
    \item \textbf{Step Extraction}: 
    After identifying the task and distinguishing the left and right arms, extract the key steps required to complete the task, ensuring that each step is clearly described and logically ordered. \textbf{Make sure the following criteria are met}:
    \begin{itemize}
        \item The \textbf{first step} must always start with \texttt{start\_frame} equal to \texttt{0}, and the \textbf{last step} must end with \texttt{end\_frame} equal to \texttt{\{cnt-1\}}.
        \item Every step must explicitly describe the actions of both the left arm (\texttt{[left]}) and right arm (\texttt{[right]}). If one arm is inactive during a step, mention its inactive state (e.g., \texttt{[left] holds the object steady while [right] tightens the screw}). If both arms are involved in the same action, use \texttt{[both]} to describe their joint activity, but still ensure to detail the left and right arm roles (e.g., \texttt{[both] lift the object, with [left] supporting the base and [right] holding the top}).
        \item \textbf{Frame window}: Specify the start and end frame for each step (from \texttt{0} to \texttt{\{cnt-1\}}, since the video has \texttt{\{cnt\}} frames).
        \item \textbf{Handling Camera Obstruction}: If at any point the view is obstructed by one or both of the robotic arms, add a special marker \texttt{[block]} at the beginning of the step description. Then, based on the context before and after the obstruction, infer the likely actions taking place and provide the most accurate analysis possible. The \texttt{[block]} token should be used only when the camera is blocked, and after the token, describe the inferred task as usual (e.g., \texttt{[block] [left] holds the object steady, [right] tightens the bolt}).
    \end{itemize}

    \item \textbf{Frame Range Constraints}:
    \begin{itemize}
        \item \textbf{No Overlapping Frames}: Ensure that the frame ranges of each step do not overlap with each other. Each frame should be assigned to exactly one step.
        \item \textbf{Full Frame Coverage}: Ensure that all \texttt{\{cnt\}} frames (from 0 to \texttt{\{cnt-1\}}) are included in the steps. No frames should be missed or duplicated.
    \end{itemize}

    \item \textbf{Notes}:
    \begin{itemize}
        \item Please annotate as finely as possible and try not to have more than ten frames of the same thing being done (unless it is difficult to distinguish).
        \item A step has only one verb (unless two or more actions are strictly performed simultaneously).
        \item Please ensure the accuracy of labeling and image matching.
        \item For example, if the gripper changes from an open state to a closed state in a sequence of frames, the action is called \texttt{pick}; On the contrary, if it changes from a closed state to an open state, the action is called \texttt{release}.
    \end{itemize}
    
    \item \textbf{Output Format}: 
    Provide the task description and steps in two parts, formatted as JSON:
    \begin{itemize}
        \item \textbf{Task Summary}: A string summarizing the primary task in the video without mentioning the subjects - the robotic arms.
        \item \textbf{Steps}: An array where each element represents a step, containing:
        \begin{itemize}
            \item \texttt{step\_description}: A concise description of the step, specifying the actions of \texttt{[left]}, \texttt{[right]}, or \texttt{[both]} arms (e.g., \texttt{[left] holds the frame, [right] tightens screws}, \texttt{[both] lift the object}). Always describe \texttt{[left]} first and \texttt{[right]} second, when applicable.
            \item \texttt{start\_frame}: The start frame of the step (from \texttt{0} to \texttt{\{cnt-1\}}).
            \item \texttt{end\_frame}: The end frame of the step (from \texttt{0} to \texttt{\{cnt-1\}}).
        \end{itemize}
    \end{itemize}
\end{enumerate}

\textbf{Task Description}: \texttt{\{desc\}}
\newline
\newline
**Example Output Format 1: (This is an example with a total frame length of 30, and the specific situation depends on the actual frame length.)**
\begin{verbatim}
```json
{
  "task_summary": "Assembling an office desk.",
  "steps": [
    {
      "step_description": "[left] removes all components from the package
                while [right] holds the package steady.",
      "start_frame": 0,
      "end_frame": 4
    },
    ...
    {
      "step_description": "[both] ensure all screws are tight and the desk is stable.",
      "start_frame": 29,
      "end_frame": 29
    }
  ]
}
```
\end{verbatim}

**Example Output Format 2: ...

\end{tcolorbox}

    \caption{Prompt for converting dual-arm robot videos into task and step descriptions in natural language.}
    \label{prompt:video2nl_dual}
\end{figure*}

\begin{figure*}[ht]
    \begin{tcolorbox}[colback=gray!5!white, colframe=gray!75!black,
    title=Prompt: Video $\rightarrow$ NL Task \& Steps (Mobile Manipulator), boxrule=0.3mm, width=\textwidth, arc=3mm, auto outer arc=true]
    \tiny
    You will analyze a \textbf{first-person perspective (ego-centric) video} (represented by image frames) of a robot performing a specific task, where the task is described as: ``\{desc\}``. Note that the referenced task summary might not be accurate or complete.

Your task is to identify the primary task during the video with the help of the referenced description, summarize the task and rewrite the description, extract the necessary steps to complete it, and specify the frame range for each step. Each step will include a \textbf{state} and a corresponding \textbf{action description}. Please ensure that the division of atomic tasks is as detailed as possible, the task description is as clear as possible, and there are no vague descriptions.
\newline
\newline
1. \textbf{Task Identification}: First, identify the main task the robot is performing. This task could be a clear goal or a series of related activities (e.g., assembling furniture, repairing equipment, preparing food, etc.). Briefly describe the primary task in one sentence.
\newline
\newline
2. \textbf{Step Extraction}: Once the task is identified, extract the key steps required to complete it, ensuring that each step is clearly described and logically ordered. Each step consists of two parts:
    \begin{itemize}
        \item \textbf{State}: The state of the robot during the step---whether it is moving (\texttt{[mobile]}), manipulating (\texttt{[manipulation]}), or observing (\texttt{[observation]}):
        \begin{itemize}
            \item \texttt{[mobile]}: The robot changes position, but the camera view does not rotate. Describe the movement of the robot (e.g., "Move towards the table").
            \item \texttt{[manipulation]}: The robot's position and camera view remain static, but the hands are performing actions. Use \texttt{[left]}, \texttt{[right]}, or \texttt{[both]} to describe the actions of the hands (e.g., "\texttt{[left]} holds the frame, \texttt{[right]} tightens the screws").
            \item \texttt{[observation]}: This state \textbf{must} be triggered when the ego-centric camera's view or orientation changes, indicating the robot is observing or searching for an object. This step is required as a transition whenever the camera view changes, regardless of the robot's other actions. For each \texttt{[observation]} state, use the following format:
            \begin{itemize}
                \item \textbf{[goal]}: Clearly specify the object or scene the robot is searching for, using \texttt{[find]} to indicate what the robot is looking for or detecting (e.g., "\texttt{[find]} the legs of the desk").
                \item \textbf{[current object or scene]}: Describe what the robot currently sees (e.g., "A table and a chair are visible").
                \item \textbf{[search result]}: Indicate whether the target object has been found (\texttt{yes}, \texttt{no}, \texttt{part}).
            \end{itemize}
        \end{itemize}
        \item \textbf{Action Description}: A concise description of what happens during the step, based on the state.
    \end{itemize}

3. \textbf{Output Format}: Provide the task description and steps in two parts, formatted as JSON:
    \begin{itemize}
        \item \textbf{Task Summary}: A string summarizing the primary task in the video without mentioning the subjects - the robot's hands or the mobile base.
        \item \textbf{Steps}: An array where each element represents a step, containing:
        \begin{itemize}
            \item \texttt{state}: The current state of the robot (\texttt{[mobile]}, \texttt{[manipulation]}, or \texttt{[observation]}).
            \item \texttt{action\_description}: A detailed description of the action in the current state (e.g., "\texttt{[mobile]} Move towards the table", "\texttt{[manipulation]} \texttt{[left]} holds the frame, \texttt{[right]} tightens the screws", "\texttt{[observation]} \texttt{[goal]}: \texttt{[find]} the legs of the desk. \texttt{[current object or scene]}: A table is visible. \texttt{[search result]}: no"). Always describe \texttt{[left]} first and \texttt{[right]} second if the two hands have different functions, when applicable.
            \item \texttt{start\_frame}: The start frame of the step (from \texttt{0} to \texttt{\{cnt-1\}}).
            \item \texttt{end\_frame}: The end frame of the step (from \texttt{0} to \texttt{\{cnt-1\}}).
        \end{itemize}
    \end{itemize}

4. \textbf{Frame Range Constraints}:
    \begin{itemize}
        \item \textbf{No Overlapping Frames}: Ensure that the frame ranges of each step do not overlap with each other. Each frame should be assigned to exactly one step.
        \item \textbf{Full Frame Coverage}: Ensure that all \{cnt\} frames (from 0 to \{cnt-1\}) are included in the steps. No frames should be missed or duplicated.
    \end{itemize}

5. \textbf{Important Notes}:
    \begin{itemize}
        \item Please annotate as finely as possible and try not to have one step with more than ten frames doing the same thing (unless it is surely difficult to distinguish).
        \item A step has only one verb (unless two or more actions are strictly performed simultaneously).
        \item Please ensure the accuracy of labeling and image matching.
        \item The video consists of exactly \{cnt\} frames. Therefore, ensure that the \textbf{end frame} for the last step is always \{cnt-1\}, and no frame should exceed this value.
        \item Ensure that the \textbf{[observation]} state appears \textbf{at the beginning} of the task to search for the initial target object, and as needed thereafter.
        \item For \textbf{[observation]} states, use the special tokens \textbf{[goal]}, \textbf{[current object or scene]}, and \textbf{[search result]}.
        \item Always ensure \textbf{[observation]} concludes with \texttt{search result: yes} before transitioning to \texttt{[mobile]} or \texttt{[manipulation]}.
        \item \textbf{Left and Right Hand Determination}: Use global context, spatial relationships, and object positions to distinguish between \texttt{[left]} and \texttt{[right]} hands.
    \end{itemize}

\textbf{Task Description}: \{desc\}
\newline
\newline
\textbf{Example Output Format (for a video with 60 frames):}
\begin{verbatim}
```json
{
    "task_summary": "Assembling an office desk.",
    "steps": [
        {
            "state": "[observation]",
            "action_description": "[goal]: [find] the legs of the desk. [current object or scene]:
                    A table is visible. [search result]: no.",
            "start_frame": 0,
            "end_frame": 1
        },
        ...
    ]
}
```
\end{verbatim}
\end{tcolorbox}
    \caption{Prompt for converting mobile manipulator robot videos into task and step descriptions in natural language.}
    \label{prompt:video2nl_mobile}
\end{figure*}

\paragraph{2. NL Steps $\rightarrow$ Predefined Function Sequence} 
Natural language step descriptions are mapped to a sequence of predefined functions. Two function lists are used: Manipulation (Fig.~\ref{prompt:func_list_manipulation}) and Navigation (Fig.~\ref{prompt:func_list_navigation}), followed by a conversion prompt (Fig.~\ref{prompt:nl2func_conversion}) referencing these lists.

\begin{figure*}[ht]
\begin{tcolorbox}[colback=gray!5!white, colframe=gray!75!black,
title=Manipulation Function List, boxrule=0.3mm, width=\textwidth, arc=3mm, auto outer arc=true]
\tiny

\textbf{Functions for the actions of a gripper:}
    \begin{itemize}
        \item \texttt{move\_to(object, target\_object)}: Move the gripper. First parameter is the object currently held (\texttt{none} if empty), second parameter is the target object. Example: \texttt{move\_to(none, towel)}, \texttt{move\_to(panda\_toy, bowl)}.
        \item \texttt{hold(object)}: Keep an object in static hold. Not applicable if gripper is empty. Example: \texttt{hold(cup)}.
    \end{itemize}

\textbf{Functions for grabbing and releasing:}
    \begin{itemize}
        \item \texttt{pick\_up(object)}: Pick up a graspable object. Example: \texttt{pick\_up(apple)}.
        \item \texttt{grasp(object)}: Lightly hold an object (pick-upable or not). Example: \texttt{grasp(door\_handle)}.
        \item \texttt{place(object, target\_object)}: Place an object at a location or relative position. Example: \texttt{place(apple, table)}, \texttt{place(apple, right\_of\_banana)}.
    \end{itemize}

\textbf{Functions for using a tool:}
    \begin{itemize}
        \item \texttt{scoop(tool, contents, container)}: Use a tool to scoop contents. Use \texttt{unknown} if contents uncertain. Example: \texttt{scoop(spoon, water, bowl)}.
        \item \texttt{pour(container, contents, target\_container)}: Pour contents into target. Example: \texttt{pour(bowl, water, pot)}.
        \item \texttt{wipe(tool, object, target\_object)}: Wipe object using tool on target. Example: \texttt{wipe(towel, water, table)}.
        \item \texttt{stir(tool, contents, target\_container)}: Stir contents with tool in container. Example: \texttt{stir(spoon, soup, pot)}.
        \item \texttt{draw(tool, character, target\_object)}: Draw a character using a tool on target. Example: \texttt{draw(marker, 'A', whiteboard)}.
        \item \texttt{cut(tool, object, target\_object)}: Cut an object with a tool on target. Example: \texttt{cut(knife, tomato, chopping\_board)}.
    \end{itemize}

\textbf{Functions for interacting directly:}
    \begin{itemize}
        \item \texttt{fold(object, target\_position)}, \texttt{unfold(object, target\_position)}: Fold or unfold object to target position.
        \item \texttt{turn(object, direction, state\_of\_target\_object)}: Rotate object to target state. Directions: \{clockwise, anticlockwise, up, down, forward, backward, left, right\}. Example: \texttt{turn(faucet, clockwise, middle\_of\_sink)}.
        \item \texttt{press(tool, object)}: Press object using tool (\texttt{none} if no tool). Example: \texttt{press(none, red\_button)}.
        \item \texttt{push(object, target\_location)}, \texttt{pull(object, target\_location)}: Push or pull object to target. Example: \texttt{push(chair, under\_of\_table)}, \texttt{pull(towel, right\_side\_of\_table)}.
        \item \texttt{insert(object, target\_object)}, \texttt{pullout(object, target\_object)}: Insert or remove object from target. Example: \texttt{insert(plug, socket)}, \texttt{pullout(plug, socket)}.
    \end{itemize}

\textbf{Dual-arm specific:}
    \begin{itemize}
        \item \texttt{transfer(left/right, right/left, object)}: Transfer object between hands. Example: \texttt{transfer(left, right, bottle)}.
    \end{itemize}

\textbf{Mobile-manipulation specific:}
    \begin{itemize}
        \item \texttt{observation(object)}: Object not visible, needs to be found. Example: \texttt{observation(chair)}.
        \item \texttt{mobile(target\_object)}: Object visible but distant, move robot to approach. Example: \texttt{mobile(table)}.
    \end{itemize}

\textbf{No Operation:}
    \begin{itemize}
        \item \texttt{no\_ops}: Stay still or maintain current state.
    \end{itemize}

\end{tcolorbox}
\caption{Predefined Manipulation function list used in NL-to-function conversion.}
\label{prompt:func_list_manipulation}
\end{figure*}

\begin{figure*}[ht]
    \begin{tcolorbox}[colback=gray!5!white, colframe=gray!75!black,
    title=Navigation Function List, boxrule=0.3mm, width=\textwidth, arc=3mm, auto outer arc=true]
    \tiny

    \textbf{Navigation:}
    \begin{itemize}
        \item \texttt{turn\_left()}: Rotate the robot $90^\circ$ to the left in place.
        \item \texttt{turn\_right()}: Rotate the robot $90^\circ$ to the right in place.
        \item \texttt{turn\_back()}: Rotate the robot $180^\circ$ in place (turn around).
        \item \texttt{go\_forward(target\_location)}: Move forward in the current facing direction until reaching the specified \texttt{target\_location}. The parameter corresponds to the semantic name of the target location, e.g., 
        \texttt{go\_forward(dining\_table)} or \texttt{go\_forward(kitchen\_door)}.
    \end{itemize}
    \end{tcolorbox}
    \caption{Predefined navigation function list used in natural-language-to-function conversion. The design ensures a compact yet expressive action space that supports systematic evaluation of compositional navigation instructions.}
    \label{prompt:func_list_navigation}
\end{figure*}

\begin{figure*}[ht]
    \begin{tcolorbox}[colback=gray!5!white, colframe=gray!75!black,
    title=Prompt: NL Steps $\rightarrow$ Function Sequence, boxrule=0.3mm, width=\textwidth, arc=3mm, auto outer arc=true]
    \tiny
\textbf{Task Description} \\
You need to process a set of \textbf{natural language step descriptions} for a robotic manipulation task and extract \textbf{generalized functions}. All generated functions must be selected from a predefined list of functions, ensuring that the function names and parameters match the definitions in the list. If the input action cannot be matched to any function in the list, use \texttt{special\_action()} and generate a suitable action.

\textbf{Rules \& Requirements}
\begin{enumerate}
    \item \textbf{Select from Predefined Functions}:
    \begin{itemize}
        \item All generated functions must be selected from the following categories:
        \begin{itemize}
            \item \textbf{Object Manipulation}
            \item \textbf{Movement and Navigation}
            \item \textbf{Opening and Closing}
            \item \textbf{Special Operations}
            \item \textbf{no\_ops}
            \item \texttt{special\_action()}
        \end{itemize}
        \item If the input action cannot be matched to any function in the list, use \texttt{special\_action()}.
        \item If the input natural language action description indicates a left-handed action and a right-handed action respectively, you need to match a function for each of the left-handed and right-handed actions. Note that the description with \texttt{[both]} tag means both hands operate simultaneously and it is also suitable to match function to each of them.
    \end{itemize}
    
    \item \textbf{Function Selection Rules}:
    \begin{itemize}
        \item Select the most appropriate function based on the semantics of the action.
        \item Refer to the explanations provided for each function to ensure correct selection.
    \end{itemize}
    
    \item \textbf{Output Format}:
    \begin{itemize}
        \item The output must be in a single-line JSON array format.
        \item Each function should be enclosed in quotes and separated by commas.
        \item The order of functions should match the input.
        \item The number of output functions must be strictly equal to the number of input \textbf{natural language step descriptions}.
        \item If \texttt{special\_action()} is used, include the generated description in the JSON.
    \end{itemize}
\end{enumerate}

\textbf{Predefined Functions List with Explanations}

\textbf{[Predefined Functions List with Explanations]...}

\textbf{Example List}

\textbf{Input:} 

pick up cup, move to shelf with block, push ball, place book on desk, hover over table, [left] approach glasses case, [right] approach lid, [left] hold case steady, [right] move lid, [both] close laptop lid

\textbf{Output:} 
\begin{verbatim}
["pick_up(object)", "move_to(object, target_object)", "push(object, target_object)", 
 "place(object, target_object)", "no_ops", 
 "left:move_to(object, target_object), right:move_to(object, target_object)", 
 "left:no_ops, right:move_to(object, target_object)", 
 "left:close(object), right:close(object)"]
\end{verbatim}

\textbf{Task} \\
Convert the following natural language step descriptions into structured generalized functions:
\end{tcolorbox}
    \caption{Prompt for converting natural language steps into a sequence of predefined functions.}
    \label{prompt:nl2func_conversion}
\end{figure*}

\paragraph{3. Function Instantiation} 
Instantiate function arguments with objects extracted from step descriptions. The prompt is presented in Fig.~\ref{prompt:func_instantiation}.

\begin{figure*}[ht]
    \begin{tcolorbox}[colback=gray!5!white, colframe=gray!75!black,
    title=Prompt: Function Instantiation, boxrule=0.3mm, width=\textwidth, arc=3mm, auto outer arc=true]
    \tiny
\textbf{Task Description:} Your task is to process a set of natural language action descriptions and generate function calls based on provided function templates. The function calls must accurately reflect the semantics of the input actions and use the correct objects or targets from the descriptions.

\textbf{Rules \& Requirements}
\begin{itemize}
    \item \textbf{Input:}
    \begin{itemize}
        \item A list of natural language action descriptions (e.g., "reach for the juice bottle") and a list of function templates (e.g., "move\_to(object)").
        \item The number of function templates matches the number of action descriptions, and they correspond in order.
    \end{itemize}
    \item \textbf{Output:}
    \begin{itemize}
        \item A list of function calls that match the input actions and templates (e.g., "move\_to(juice\_bottle)").
        \item The number of function calls must match the number of input descriptions.
        \item If an object or target is composed of multiple words, separate them with underscores (\_).
    \end{itemize}
    \item \textbf{Function Call Rules:}
    \begin{itemize}
        \item Replace the placeholder (e.g., object, target\_object, goal, current\_scene\_or\_object) in the function template with the appropriate object or target from the natural language description.
        \item Ensure the function call accurately reflects the action described in the input.
    \end{itemize}
    \item \textbf{Output Format:} The output must be a list of function calls in the correct format.
\end{itemize}

\textbf{Function Templates contain three types of "observation", "mobile" and "manipulation". The specific templates and explanations are as follows:}

\textbf{[Predefined Functions List with Explanations]}

\textbf{Examples}

\textbf{Input:}
\begin{itemize}
    \item \textbf{Natural Language Descriptions:} "[goal]: [find] a cloth [current object or scene]: cabinets, countertop, and floor are visible [search result]: no", 
    "move towards the cloth", "[right] places the oil bottle back on the counter", "[right] reaches for the bowl of shrimp", 
    "[goal]: [find] spatula [current object or scene]: pan with shrimp and spatula are visible [search result]: yes", 
    "reach for the juice bottle", "grasp the juice bottle", "lift the juice bottle", "move the juice bottle towards the blue sticky note", 
    "place the juice bottle on the blue sticky note", "release the juice bottle".
    
    \item \textbf{Function Templates:} "observation(goal, current\_scene\_or\_object, result)", "mobile(target\_object)", 
    "left:no\_ops, right:place(object, target\_object)", "left:no\_ops, right:move\_to(object, target\_object)", 
    "move\_to(object)", "pick\_up(object)", "move\_to(object, target\_object)", "place(object, target\_object)".
\end{itemize}

\textbf{Output:}
\begin{verbatim}
["observation(a_cloth, ['cabinets','countertop','floor'],no)", "mobile(cloth)", 
"left:no_ops, right:place(oil_bottle, counter)", 
"left:no_ops, right:move_to(none, bowl_with_shrimp)", 
"observation(spatula, ['pan_with_shrimp_and_spatula'], yes)", 
"move_to(juice_bottle)", "pick_up(juice_bottle)", 
"pick_up(juice_bottle)", "move_to(juice_bottle, blue_sticky_note)", 
"place(juice_bottle, blue_sticky_note)", 
"place(juice_bottle, None)"]
\end{verbatim}

\textbf{Task}
Convert the following natural language step descriptions and function templates to the special function calls.
Language Descriptions: {}
Function Templates: {}    \end{tcolorbox}
    \caption{Prompt for instantiating predefined functions with specific object references.}
    \label{prompt:func_instantiation}
\end{figure*}

\paragraph{4. Explicit $\rightarrow$ Implicit Instruction Conversion}
To facilitate natural language task formulation for embodied robots, we convert explicit task instructions into implicit forms that imply the required action without directly naming the target object or task (Fig.~\ref{prompt:explicit2implicit}).
To quantify the quality of the generated implicit instructions, we audit 50 sampled instances along two axes—\emph{naturalness} (whether the phrasing reads like an everyday human request) and \emph{inferability} (whether the intended target/task can be recovered from the instruction and scene), each rated on a 1--5 scale. A text-only judge scores naturalness $4.76{\pm}0.65$ and inferability $4.92{\pm}0.27$, while a vision-grounded judge scores naturalness $4.80{\pm}0.49$ and inferability $5.00{\pm}0.00$. The consistently high scores under both judges indicate that the converted instructions remain natural and reliably inferable.
\begin{figure*}[ht]
    \begin{tcolorbox}[colback=gray!5!white, colframe=gray!75!black,
    title=Prompt: Explicit $\rightarrow$ Implicit, boxrule=0.3mm, width=\textwidth, arc=3mm, auto outer arc=true]
    \scriptsize
    \textbf{Task:}  
Generate implicit task instructions for embodied robots based on a provided explicit task and (if available) an image. The goal is to produce natural, everyday expressions that imply the need for the target task without explicitly mentioning the target object or directly stating the task.  

\textbf{Key Requirements:}  
\begin{enumerate}  
    \item \textbf{Target Entity Focus:}  
    \begin{itemize}  
        \item Identify the unique characteristic of the target entity from the explicit task and ensure that the implicit instruction reflects this characteristic.  
        \item The implicit instruction must have a direct association with the target entity and task, not abstract references.  
    \end{itemize}  

    \item \textbf{Everyday Scenarios and Language:}  
    \begin{itemize}  
        \item Use casual, real-life scenarios to imply the need for the task.  
        \item Avoid technical terms or abstract expressions. The instruction should feel like a natural request from a human in daily life.  
    \end{itemize}  

    \item \textbf{Image Integration (if provided):}  
    \begin{itemize}  
        \item Analyze the provided image directly to extract relevant information about the scene.  
        \item The image may contain multiple objects, including distracting household items unrelated to the task. Focus on the target object from the explicit task while ensuring the implicit instruction remains relevant and natural.  
        \item Do not rely on textual descriptions of the image; all visual information must come from analyzing the provided image itself.  
    \end{itemize}  

    \item \textbf{No Direct Mentions:}  
    \begin{itemize}  
        \item Do not directly mention the target object or the explicit task.  
        \item The need for the task should be implied through observations, feelings, or everyday needs (e.g., "I feel parched," instead of "pour water into the cup").  
    \end{itemize}  

    \item \textbf{Output Format:}  
    \begin{itemize}  
        \item Provide 5 implicit task instruction suggestions for each input.  
        \item The output must be in list format, with each instruction as a separate list item to ensure consistency and ease of post-processing.  
    \end{itemize}  
\end{enumerate}  

\textbf{Examples:}  

\textbf{Target object:} pouring liquid from a bottle into a cup  
\textbf{Provided Image:} A table with a bottle, an empty cup, and other unrelated items such as soap, a photo frame, and a gift bag.  

\textbf{Instruction Output:}  
\begin{itemize}  
    \item "Looking at the empty cup on the table makes me realize how thirsty I am. Could you help me with that?"  
    \item \dots  
\end{itemize}  

\textbf{Target object:} watering a plant  
\textbf{Provided Image:} A windowsill with several potted plants, one with drooping leaves.  

\textbf{Instruction Output:}  
\begin{itemize}  
    \item "This plant’s leaves look a bit droopy today. Could you help bring it back to life?"  
    \item \dots  
\end{itemize}  

\textbf{Target object:} selecting a Batman toy  
\textbf{Provided Image:} A shelf filled with various superhero toys, including a Batman figure.  

\textbf{Instruction Output:}  
\begin{itemize}  
    \item "I’m a big fan of DC series, please help me choose a suitable toy."  
    \item \dots  
\end{itemize}  

\textbf{Input Format:}  
\begin{itemize}  
    \item \textbf{Target object:} \{task\_summary\}  
    \item \textbf{Provided Image:} \{image\_path\}  
\end{itemize}  

\textbf{Output Format:}  
\begin{itemize}  
    \item \textbf{Instruction Output:}  
    \begin{itemize}  
        \item "Instruction 1"  
        \item "Instruction 2"  
        \item "Instruction 3"  
        \item "Instruction 4"  
        \item "Instruction 5"  
    \end{itemize}  
\end{itemize}  

    \end{tcolorbox}
    \caption{Prompt for converting explicit task instructions into implicit form.}
    \label{prompt:explicit2implicit}
\end{figure*}

\paragraph{5. Fine-grained Attribute Extraction} 
We design prompts to extract fine-grained, step-level information from video frames, including \textbf{objects} (Fig.~\ref{prompt:step_attribute_objects}), \textbf{actions} (Fig.~\ref{prompt:step_attribute_actions}), and \textbf{scene labels} (Fig.~\ref{prompt:scene_labels}). These structured annotations provide the foundation for evaluating compositional reasoning and downstream task performance.

\begin{figure*}[ht]
    \begin{tcolorbox}[colback=gray!5!white, colframe=gray!75!black,
    title=Prompt: Step-level Object Extraction, boxrule=0.3mm, width=\textwidth, arc=3mm, auto outer arc=true]
    \tiny
    The following is a long-term robot task instruction text. Please analyze the objects being manipulated and other objects mentioned contained in it and generate the output part in a strict format. All words are in the singular form. Object properties (if there are properties other than quantity) are placed in parentheses. You also need to analyze, what kind of object is this manipulation for, choosing from rigid (like bowl), articulated (like microwaves), deformable (like cloth), special (like liquid). 

For example, 
\begin{itemize}
    \item task instruction: picking grapes and placing them on a plate; output: [manipulated: grape=rigid][other: plate]
    \item task instruction: picking up a bowl; output: [manipulated: bowl=rigid][other: none]
    \item task instruction: pick up a small coca-cola can and place it on a blue paper; output: [manipulated: can(small, coca-cola)=rigid][other: paper(blue)]
    \item task instruction: pouring dice from a white cup into a pink cup; output: [manipulated: dice=special, cup(white)=rigid][other: cup(pink)]
    \item task instruction: pouring water from one cup to another; output: [manipulated: water=special, cup=rigid][other: cup]
    \item task instruction: transferring a usb flash drive between two devices; output: [manipulated: usb flash drive=rigid][other: device]
    \item task instruction: folding a cloth; output: [manipulated: cloth=deformable][other: none]
    \item task instruction: opening a laptop; output: [manipulated: laptop=articulated][other: none]
    \item task instruction: closing a cabinet door; output: [manipulated: door(cabinet)=articulated][other: none]
    \item task instruction: wiping a table; output: [manipulated: cloth=deformable][other: table]
    \item task instruction: writing the number 14 on a whiteboard; output: [manipulated: mark pen=rigid][other: whiteboard]
\end{itemize}
\end{tcolorbox}
    \caption{Prompt used to extract \textbf{objects} at the step level from video frames.}
    \label{prompt:step_attribute_objects}
\end{figure*}

\begin{figure*}[ht]
    \begin{tcolorbox}[colback=gray!5!white, colframe=gray!75!black,
    title=Prompt: Step-level Action Extraction, boxrule=0.3mm, width=\textwidth, arc=3mm, auto outer arc=true]
    \tiny
    Below are task instructions for long-term robotic tasks. Your goal is to extract the primary action from each task instruction. 

For instance:
\begin{itemize}
    \item Task instruction: dragging a strainer backwards, should return: \texttt{drag}
    \item Task instruction: Spelling "THU" with blocks, should return: \texttt{spell}
    \item Task instruction: inserting a three-pronged object into its matching slot, should return: \texttt{insert}
    \item Task instruction: filling a bottle with water from a dispenser, should return: \texttt{fill}
    \item Task instruction: pick apple and place in the drawer, should return: \texttt{pick, place in}
\end{itemize}

These examples illustrate the type of concise action extraction required. The task instruction provides a high-level overview of the task, and your job is to distill it into its core action(s). This prompt is designed for the Gemini model, and it is crucial to focus on identifying the main action verbs that define the task.
\end{tcolorbox}
    \caption{Prompt used to extract \textbf{actions} at the step level from video frames.}
    \label{prompt:step_attribute_actions}
\end{figure*}

\begin{figure*}[ht]
    \begin{tcolorbox}[colback=gray!5!white, colframe=gray!75!black,
    title=Prompt: Step-level Scene Label Extraction, boxrule=0.3mm, width=\textwidth, arc=3mm, auto outer arc=true]
    \tiny
    Please analyze the following image scene and provide a set of scene tags. The specific example of the tags is as follows:

\begin{itemize}
    \item \{'Primary Tag': 'Lab Scene', 'Secondary Tag': 'Robotics Testing Area', 'Tertiary Tag': 'Shape-Sorter Toy', 'Perspective Tag': 'Third-person View'\}
    \item \{'Primary Tag': 'Lab Scene', 'Secondary Tag': 'Robot control room', 'Tertiary Tag': 'Workbench', 'Perspective Tag': 'Third-person View'\}
    \item \{'Primary Tag': 'Family Scene', 'Secondary Tag': 'Kitchen', 'Tertiary Tag': 'Chopping Board', 'Perspective Tag': 'First-person View'\}
\end{itemize}

Among them, the tertiary tag represents the specific background of these images (for example, all operations in the experiment are carried out on the chopping board); the secondary tag is a collection of tertiary tags, such as stoves, chopping boards, these tertiary scene tags should be clustered into the kitchen, this secondary scene tag, basket frames, and basketball hoops, these tertiary scene tags should be clustered into the basketball court, this secondary scene tag; the primary tag is the clustering of secondary tags, such as kitchens, living rooms, and bedrooms, these secondary scene tags should be clustered into the family scene, this primary tag; basketball courts, volleyball courts, and ping-pong tables, these secondary scene tags should be clustered into the gym, this primary tag; the perspective tag indicates whether the image is shot from a first-person or third-person perspective. The first-person perspective should have the entire scene moving with the operation of the robotic arm, while in the third-person perspective, the overall scene should remain stationary, with only the robotic hand moving.

\textbf{Note:}
\begin{enumerate}
    \item If you don't know what scene tags to fill in, you can fill in <unknown> as the answer.
    \item The Primary Tag, Secondary Tag, and Tertiary Tag must fill in the background information of the several frames of images I gave you. Please do not fill in other information!!!
\end{enumerate}

\textbf{Again, your result must use the following format:}
\begin{itemize}
    \item Before providing your tag answers, please explain the reasoning behind the labels you have given.
\end{itemize}
\begin{verbatim}
[
    { 
        "Primary Tag": "the primary scene tag", 
        "Secondary Tag": "the secondary scene tag", 
        'Tertiary Tag': "the tertiary scene tag", 
        "Perspective Tag': "the perspective type of images"
    }
]
\end{verbatim}

\end{tcolorbox}
    \caption{Prompt used to extract \textbf{scene labels} at the step level from video frames.}
    \label{prompt:scene_labels}
\end{figure*}

\subsubsection{Perception Tasks: Functional Attribute Subtasks}
To evaluate an AI system’s understanding of object functionality in images, we generate multi-choice question-answer pairs based on objects highlighted with bounding boxes, focusing on detailed attribute analysis and functional reasoning (Fig.~\ref{prompt:functional_qa_generation}).

\begin{figure*}[ht]
    \begin{tcolorbox}[colback=gray!5!white, colframe=gray!75!black,
    title=Prompt: Functional Attribute QA Generation, boxrule=0.3mm, width=\textwidth, arc=3mm, auto outer arc=true]
    \tiny
    As an AI visual assistant, you are tasked with analyzing an image that includes a single marked bounding box (colored green) and generate a multi-choice question-answer pair. The bounding box format is [x, y, width, height], x and y typically represent the coordinates of the top-left corner of the bounding box, while width and height represent the width and height of the box. The image bounding box: \{bounding\_box\}. Initially, you must acquire a comprehensive and intricate understanding of the image, ideally down to each pixel area. Your primary focus should be on deciphering the specific objects or contexts associated with the marks I have placed within the image, as this is of paramount importance. Your analysis will encompass two principal functions:
\newline
\newline
\textbf{Bounding Box Description}: For each marked bounding box in the image, provide a thorough description using natural language. Detail attributes such as the category, type, color, functionality, and other characteristics of the object, including its location, condition, and any additional pertinent attributes. Envision yourself observing directly and convey your observations as thoroughly and promptly as possible.
\newline
\newline
\textit{-- Template for Role 1}:
\newline
\textit{`Bounding Box 1': [Comprehensive Description]}
\newline
\textit{Continue in this manner for marked bounding box.}
\newline
\newline
\textbf{Multi-choice Question-Answer Pair Generation}: Using the functionality that you describe in Role 1, generate a question about the functionality or use of the object inside the Bounding Box. Provide multiple-choice options and the correct answer. Ensure the question and options are relevant to the object's functionality. The choice in the question must be a little bit difficult to answer, make sure is a good question to distinguish where the model are smart or not.
\newline
\newline
\textit{-- Template for Role 2}:
\newline
\textit{``Question": What functionality does the object inside the Green Bounding Box have? (A) ... (B) ... (C) ... (D) ...}
\newline
``Answer": (...) 
\newline
\textit{--Example}: 
\newline
\textit{``Question": What functionality does the object inside the Green Bounding Box have? (A) Used to store water (B) Used to store food (C) Used for drinking hot beverages (D) Used to plant flowers
\newline
``Answer": (C)}
\end{tcolorbox}
    \caption{Prompt for generating question-answer pairs describing object functionalities.}
    \label{prompt:functional_qa_generation}
\end{figure*}

\subsection{Evaluation Prompts}

\subsubsection{Evaluator Robustness and Implicit-CoT Variants}
\label{subsec:eval_variants}

For the evaluator-robustness study, we use three Q1 scoring prompt variants. Variant v1 is a basic judge prompt without an explicit rollout trace. Variant v2 adds structural scoring rules for matching skill--object--parameter nodes and checking task completion. Variant v3 is the production prompt used in RoboBench: it requires deterministic ground-truth node counting and a four-stage DAG-grounded rollout consisting of visual constraint analysis, critical state extraction, state-order validation, and step-by-step simulation. The same 200 stratified Q1 samples are evaluated by four judges under all variants.

For the implicit-instruction ablation, the CoT condition adds a short reasoning scaffold before final action generation: infer the user's latent intent, identify the scene objects relevant to that intent, infer the desired goal state, and then output its final answer in the same format as the standard condition. This scaffold is used only for diagnostic analysis; the benchmark's default instruction-comprehension setting evaluates direct model behavior without this additional hint.

\subsubsection{Planning Tasks: Q1 Evaluation}
For Q1-type planning tasks, we provide a structured two-step evaluation: first, extracting the action sequence from model responses (Fig.~\ref{prompt:q1_action_extraction}), and second, scoring the extracted plans against ground truth data using embodied task reasoning criteria (Fig.~\ref{prompt:q1_scoring}).
\begin{figure*}[ht]
    \begin{tcolorbox}[colback=gray!5!white, colframe=gray!75!black,
    title=Prompt: Q1 Evaluation - Action List Extraction, boxrule=0.3mm, width=\textwidth, arc=3mm, auto outer arc=true]
    \tiny
    \textbf{Task:}  
    You are given an input dataset containing a robotic manipulation task goal, a previously executed step, 
    and a response describing the remaining steps. Your task is to extract structured action plans 
    in a specific function format.  

    \medskip
    \textbf{Instructions:}
    \begin{enumerate}
        \item \textbf{Extract Key Information:}  
        Identify the task goal from the \texttt{prompt} field and assign it to the \texttt{"task\_summary"} field.  
        Extract action functions from the \texttt{previous\_step} and \texttt{response} fields to construct 
        the sequence of necessary steps in \texttt{"plan\_step"}.

        \item \textbf{Strict Action Function Format:}  
        Use only the predefined action functions listed below. Do not modify function names or introduce new ones.  
        Ensure all function names match exactly. Arguments (\texttt{object}, \texttt{target\_object}, 
        \texttt{carry\_object}, \texttt{direction}) should be generalized but faithful to the task.

        \item \textbf{Maintain Execution Order:}  
        The \texttt{"plan\_step"} list should follow the correct execution order.

        \item \textbf{Reasoning Explanation:}  
        Provide a \texttt{"reason"} field explaining how the \texttt{"task\_summary"} 
        and \texttt{"plan\_step"} were derived, including how you determined the format (single-arm vs dual-arm).
    \end{enumerate}

    \medskip
    \textbf{Predefined Action Functions:}  
    \texttt{""" + PREDEFINED\_ACTIONS + """}  

    \medskip
    \textbf{Output Format (JSON):}  

    \textit{For Single-Arm Tasks:}
    \begin{verbatim}
    {
      "task_summary": "<task goal>",
      "plan_step": ["<action_function_1>", "<action_function_2>", ...],
      "reason": "<your reasoning>"
    }
    \end{verbatim}

    \textit{For Dual-Arm Tasks:}
    \begin{verbatim}
    {
      "task_summary": "<task goal>",
      "plan_step": [
        "<action_function_1>",
        "left:<action_function_2>, right:<action_function_3>",
        ...
      ],
      "reason": "<your reasoning>"
    }
    \end{verbatim}

    The data provided is as follows: \texttt{\{data\}}.  
    Please output your results as required.
    \end{tcolorbox}
    \caption{Prompt for extracting structured action lists from model outputs for Q1-type questions.}
    \label{prompt:q1_action_extraction}
\end{figure*}

\begin{figure*}[ht]
    \begin{tcolorbox}[colback=gray!5!white, colframe=gray!75!black,
    title=Prompt: Q1 Evaluation - World-Simulator Scoring, boxrule=0.3mm, width=\textwidth, arc=3mm, auto outer arc=true]
    \tiny
    You are a world-simulator judge for robot task plans. Given a scene image, a ground-truth (GT) action sequence, a GT DAG of dependencies, and a model's predicted plan, output two integer scores in $[0,10]$.

    \smallskip
    \textbf{Inputs.}
    \begin{itemize}\itemsep0pt
        \item \textbf{Scene image} $I_0$: first frame of the task---use it to identify the objects, their initial spatial relations, and physical constraints.
        \item \textbf{GT action list} $A^*$: e.g.\ \texttt{1-grasp(cup), 2-pick\_up(cup), 3-place(cup, table)} (single-arm) or \texttt{1-left:move\_to(none, towel), right:move\_to(none, towel), 2-left:grasp(edge), right:grasp(edge), ...} (dual-arm).
        \item \textbf{GT DAG} $G$: dependency graph over GT actions; defines which actions enable which milestones.
        \item \textbf{Model plan}: predicted action list in the same syntax. Each action is a node \texttt{(skill, object, parameter?)}.
    \end{itemize}

    \smallskip
    \textbf{Score 1: Node Correctness.}
    \begin{enumerate}\itemsep0pt
        \item \textbf{Count GT nodes} $N_{GT}$ exactly: a line \texttt{i-action(args)} contributes 1 node; a dual-arm line \texttt{i-left:actionL, right:actionR} contributes 2 nodes (one per arm). \texttt{no\_ops} is a node and only matches another \texttt{no\_ops}. Do not collapse, merge, or drop nodes.
        \item \textbf{Greedy one-to-one match}: walk the model plan left-to-right; for each predicted node find at most one unmatched GT node where all three hold: (i) \textbf{Skill} token identical (\texttt{grasp}$\neq$\texttt{pick\_up}, \texttt{push}$\neq$\texttt{pull}, \texttt{place}$\neq$\texttt{insert}); (ii) \textbf{Object} same physical referent, alias allowed only for the same entity (\texttt{faucet}$\approx$\texttt{tap}, \texttt{cup}$\approx$\texttt{mug}), but different colors/sides/IDs never match (\texttt{red\_button}$\neq$\texttt{green\_button}, \texttt{object\_3}$\neq$\texttt{object\_8}); for dual-arm the arm side must match; (iii) \textbf{Parameter} directionally/functionally equivalent (\texttt{clockwise}$\approx$\texttt{cw}, \texttt{open}$\approx$\texttt{on}), named locations must match (\texttt{bottom\_left\_corner}$\neq$\texttt{point\_5}). A node is fully correct or 0---no partial credit.
        \item $\text{NodeCorrectness} = \lfloor \text{matched} / N_{GT} \times 10 \rfloor$.
    \end{enumerate}

    \smallskip
    \textbf{Score 2: Task Completion (world-simulation rollout).}
    Handle trivial goals first: if after removing excluded robot motions the critical-state set $S^*$ is \textbf{empty}, output $\text{Completion}=10$ and stop. For dual-arm tasks, evaluate $S^*$ only from the arm that carries a genuine state change (a \texttt{no\_ops}-only arm contributes nothing and is not penalized). Otherwise:
    \begin{enumerate}\itemsep0pt
        \item \textbf{Initial world} $W_0$: from $I_0$, list objects in $A^*$ with their starting locations/states.
        \item \textbf{Critical states} $S^*$: from $A^*$ and $G$, the required \emph{object-level} state changes (object moved to new container/surface; container open/closed; object activated/deactivated; orientation/assembly changed). \textbf{Excluded} robot motions: \texttt{move\_to, grasp, hold, approach, align, release, retract, observation, look, scan, plan, think}.
        \item \textbf{Order} via $G$: a state can be marked achieved only after its prerequisites.
        \item \textbf{Rollout}: simulate the model plan step-by-step against $W_0$; check preconditions, update $W_t\rightarrow W_{t+1}$, and mark any $s\in S^*$ accomplished. Aggregate achieved set $\hat{S}\subseteq S^*$.
    \end{enumerate}
    Counting rules: \emph{implicit succession} (\texttt{pick\_up} implies \texttt{grasp}+\texttt{move\_to}; \texttt{place} implies \texttt{move\_to}---missing intermediate motions do not block a reached goal); \emph{wrong-direction} actions achieve nothing (\texttt{push(drawer, inward)} does not open it); \emph{wrong object/target} yields no progress; \emph{surrogate skill} counts if it physically reaches the same end state (\texttt{pour} from a container instead of \texttt{scoop}+\texttt{pour}), but not if it cannot (\texttt{push} along a table when the goal needs lifting onto a shelf). $\text{TaskCompletion} = \lfloor |\hat{S}| / |S^*| \times 10 \rfloor$.

    \smallskip
    \textbf{Worked examples (abridged).}
    \textbf{(A)} Single-arm, model skips \texttt{pick\_up}: 3/4 nodes $\rightarrow$ Node $=7$; apple still reaches bowl $\rightarrow$ Completion $=10$.
    \textbf{(B)} Dual-arm towel-fold, model omits both \texttt{pick\_up}s: 6/8 $\rightarrow$ Node $=7$; towel unfolded $\rightarrow$ Completion $=10$.
    \textbf{(C)} Drawer wrong direction (\texttt{push,inward} vs \texttt{pull,outward}): 1/2 $\rightarrow$ Node $=5$; drawer not opened $\rightarrow$ Completion $=0$.
    \textbf{(D)} Wrong object ID throughout: 0/3 $\rightarrow$ Node $=0$; state not achieved $\rightarrow$ Completion $=0$.
    \textbf{(E)} Functional surrogate (\texttt{pour(jar,...)} for \texttt{scoop}+\texttt{pour(spoon,...)}): 0 nodes match $\rightarrow$ Node $=0$; sauce still in bowl $\rightarrow$ Completion $=10$.
    \textbf{(F)} Empty $S^*$ (pre-task \texttt{move\_to}+\texttt{no\_ops}): Completion $=10$ regardless of model output.
    \textbf{(G)} Dual-arm one-side \texttt{no\_ops}: evaluate Completion only against the active arm; the \texttt{no\_ops} lines are not penalized.

    \smallskip
    \textbf{Output (JSON only).}
    \begin{verbatim}
{
  "node_correctness": {"result": <int 0-10>,
     "reason": "matched=X / N_GT=Y -> floor(X/Y*10)=Z. <aliases used>"},
  "task_completion":  {"result": <int 0-10>,
     "reason": "achieved=A / |S*|=B -> floor(A/B*10)=C. <name S* and which achieved>"}
}
    \end{verbatim}
    Both scores must be integers in $[0,10]$; each \texttt{reason} must show explicit counts. Output JSON only, no surrounding prose.
    \end{tcolorbox}
    \caption{Prompt for scoring model outputs for Q1-type questions based on ground truth (the DAG-grounded world-simulator rollout prompt, v3, used in RoboBench).}
    \label{prompt:q1_scoring}
\end{figure*}

\subsubsection{Planning Tasks: Q2 Evaluation}
For Q2-type planning tasks, we follow a two-stage procedure: first, extracting the next actionable step from the model response in a structured robotic action format (Fig.~\ref{prompt:q2_action_extraction}), and second, quantitatively scoring the extracted step against the ground truth using strict skill, object, and parameter evaluation criteria (Fig.~\ref{prompt:q2_scoring}).
\begin{figure*}[ht]
    \begin{tcolorbox}[colback=gray!5!white, colframe=gray!75!black,
    title=Prompt: Q2 Evaluation - Action Extraction, boxrule=0.3mm, width=\textwidth, arc=3mm, auto outer arc=true]
    \tiny
    \textbf{Task:}  
    Extract the next step from the given response and represent it in a structured robotic action format.

    \medskip
    \textbf{Action Format:}  
    The extracted step must follow the format:  
    \texttt{skill(element1, element2, ...)}  

    \medskip
    \textbf{Examples:}  
    \begin{itemize}
        \item \texttt{grasp(microwave\_handle)}  
        \item \texttt{push(microwave\_handle, close)}  
        \item \texttt{move\_to(none, drawer)}  
    \end{itemize}

    \medskip
    \textbf{Input:}  
    Response: \{response\}  

    \medskip
    \textbf{Output Instruction:}  
    Return \textbf{only} the extracted step in the correct format.  
    Do not include any explanations, extra text, or additional symbols.
    \end{tcolorbox}
    \caption{Prompt for extracting structured actions from model outputs for Q2-type questions.}
    \label{prompt:q2_action_extraction}
\end{figure*}

\begin{figure*}[ht]
    \begin{tcolorbox}[colback=gray!5!white, colframe=gray!75!black,
    title=Prompt: Q2 Evaluation - Prompt-based Scoring, boxrule=0.3mm, width=\textwidth, arc=3mm, auto outer arc=true]
    \tiny
        \textbf{Task:}  
    You are given two robot action steps (an \textit{extracted} step from a model and a \textit{ground-truth} step). Evaluate their similarity using the prescribed, quantitative criteria below and provide a concise justification for each sub-score.

    \medskip
    \textbf{Input Data Format:}
    \begin{verbatim}
    {
      "extracted_step": "The step extracted from model response",
      "gt_step": "The ground truth step"
    }
    \end{verbatim}

    \medskip
    \textbf{Evaluation Criteria:}

    \begin{enumerate}
        \item \textbf{Skill usage accuracy (0 or 1 point).}  
        Consider only the skill/action token (e.g., \texttt{grasp}, \texttt{push}, \texttt{move\_to}) in both steps. Award \texttt{1} iff the skills are exactly identical (strict match after normalization); otherwise award \texttt{0}. \\
        \textit{Normalization rule:} lowercase, strip extra spaces/underscores for comparison (e.g., \texttt{Pick\_Up} $\rightarrow$ \texttt{pickup} for matching purposes).
        \smallskip

        \item \textbf{Operation object reasonableness (0, 0.5, or 1 point).}  
        Compare the object argument(s) referenced by the operations:
        \begin{itemize}
            \item \texttt{1.0}: Identical or clearly the same referent (alias/orthographic variants accepted, e.g., \texttt{door\_handle} vs \texttt{door handle}).
            \item \texttt{0.5}: Related or part/whole/category-level match (e.g., \texttt{table} vs \texttt{table\_leg}, \texttt{cup} vs \texttt{mug}).
            \item \texttt{0.0}: Unrelated or incompatible objects.
        \end{itemize}
        Evaluate object similarity using semantic equivalence and task context.
        \smallskip

        \item \textbf{Parameter accuracy (0, 0.5, or 1 point).}  
        Evaluate additional parameters (e.g., target positions, directions, contents). \textbf{Important:} if Skill score = 0 or Object score = 0, Parameter score = 0. Otherwise:
        \begin{itemize}
            \item \texttt{1.0}: Parameters fully correct and precise for execution.
            \item \texttt{0.5}: Parameters partially correct or imprecise but salvageable.
            \item \texttt{0.0}: Parameters incorrect or irrelevant.
        \end{itemize}
    \end{enumerate}

    \medskip
    \textbf{Evaluation Guidelines:}
    \begin{itemize}
        \item Enforce \textbf{strict} skill matching; be \textbf{flexible} on objects in Standard mode (use context), but adhere to strict ID matching in CSS/ID mode if applicable.
        \item Provide brief, explicit reasons for each sub-score (one-line justification).
        \item Consider execution precision required by robotic control when judging parameters (e.g., \texttt{push(drawer, close)} vs \texttt{push(drawer, slight\_nudge)}).
        \item Examples:
        \begin{itemize}
            \item Extracted: \texttt{push(drawer, close)} | GT: \texttt{push(drawer\_handle, open)} $\rightarrow$ Skill=1, Object=0.5 (part-whole), Parameter=0.5 (partial).
            \item Extracted: \texttt{move\_to(none, table)} | GT: \texttt{move\_to(none, table)} $\rightarrow$ Skill=1, Object=1, Parameter=1.
        \end{itemize}
    \end{itemize}

    \medskip
    \textbf{Output Format (JSON):}  
    Return a single JSON object strictly matching the structure below. Each `"reason"` entry should be a concise justification for the assigned score.

    \begin{verbatim}
    {
        "skill_usage_accuracy": {"result": x, "reason": "brief explanation of evaluation"},
        "operation_object_reasonableness": {"result": y, "reason": "brief explanation of evaluation"},
        "parameter_accuracy": {"result": z, "reason": "brief explanation of evaluation"}
    }
    \end{verbatim}

    \medskip
    The data provided is as follows:

    Extracted step: \texttt{\{extracted\_step\}}  
    Ground truth step: \texttt{\{gt\_step\}}

    Please output your results exactly in the JSON format above.
    \end{tcolorbox}
    \caption{Prompt for scoring Q2-type questions based on model outputs and prompt instructions.}
    \label{prompt:q2_scoring}
\end{figure*}

\subsubsection{Planning Tasks: Q3 Evaluation}
For Q3-type planning tasks, the evaluation involves converting model responses into a strict binary decision, where only "yes" or "no" is returned without any additional text or formatting, as illustrated in Fig.~\ref{prompt:q3_yesno}.

\begin{figure*}[ht]
    \begin{tcolorbox}[colback=gray!5!white, colframe=gray!75!black,
    title=Prompt: Q3 Evaluation - Yes/No Conversion, boxrule=0.3mm, width=\textwidth, arc=3mm, auto outer arc=true]
    \tiny
    \textbf{Task:}  
    From a given response, extract only the binary decision: \texttt{"yes"} or \texttt{"no"}.  
    No explanations, additional words, or formatting are allowed.

    \medskip
    \textbf{Input Format:}
    \begin{verbatim}
    Response: {response}
    \end{verbatim}

    \medskip
    \textbf{Output Requirement:}
    \begin{itemize}
        \item Return strictly one token: \texttt{"yes"} or \texttt{"no"}.
        \item Output must be lowercase.
        \item No punctuation, whitespace, or extra text is permitted.
    \end{itemize}

    \medskip
    \textbf{Examples:}
    \begin{itemize}
        \item Response: "Yes, that is correct." $\rightarrow$ Output: \texttt{yes}
        \item Response: "No, it does not work." $\rightarrow$ Output: \texttt{no}
    \end{itemize}
    \end{tcolorbox}
    \caption{Prompt for converting model outputs into yes/no answers for Q3-type questions.}
    \label{prompt:q3_yesno}
\end{figure*}

\clearpage
\bibliographystyle{splncs04}
\bibliography{main}
\end{document}